\documentclass[preprint,12pt,3p]{elsarticle}

\usepackage{amssymb}
\usepackage{float}
\usepackage{graphicx}
\usepackage{url}

\usepackage{color,soul}
\definecolor{Gray}{gray}{0.9}
\usepackage{longtable}
\usepackage{pifont}
\usepackage{siunitx}
\usepackage{amssymb}
\usepackage{multirow}
\usepackage{array}
\usepackage{subcaption}
\usepackage[table,xcdraw]{xcolor}
\usepackage[normalem]{ulem}
\useunder{\uline}{\ul}{}
\usepackage{csquotes}
\usepackage{lscape}
\usepackage[utf8x]{inputenc}
\usepackage{array}
\usepackage{amsmath}
\usepackage{cleveref}
\newcolumntype{C}[1]{>{\centering\arraybackslash}p{#1}}
\newcommand{\xmark}{\ding{55}}%

\journal{Computational and Structural Biotechnology Journal}

\begin{document}
\begin{frontmatter}

\title{Secure and Privacy-Preserving Automated Machine Learning Operations into End-to-End Integrated IoT-Edge-Artificial Intelligence-Blockchain Monitoring System for Diabetes Mellitus Prediction}

\author[label1]{Alain Hennebelle}
\address[label1]{School of Computing and Information Systems, The University of Melbourne, Australia}

\author[label1,label2,label3]{Leila Ismail\corref{cor1}}
\address[label2]{Intelligent Distributed Computing and Systems (INDUCE) Research Laboratory, Department of Computer Science and Software Engineering, College of Information Technology, United Arab Emirates University, United Arab Emirates}
\address[label3]{National Water and Energy Center, United Arab Emirates University, United Arab Emirates}

\cortext[cor1]{I am corresponding author}
\ead{leila@uaeu.ac.ae}

\author[label2,label3]{Huned Materwala}

\author[label4,label5]{Juma Al Kaabi}
\address[label4]{College of Health Sciences, Department of Internal Medicine, United Arab Emirates University, United Arab Emirates}
\address[label5]{Mediclinic, Al Ain, Abu Dhabi, United Arab Emirates}

\author[label6]{Priya Ranjan}
\address[label6]{Bhubaneswar Institute of Technology, India}

\author[label7]{Rajiv Janardhanan}
\address[label7]{Faculty of Medical \& Health Sciences, SRM Institute of Science \& Technology, India}

\begin{abstract}
Diabetes Mellitus, one of the leading causes of death worldwide, has no cure to date and can lead to severe health complications, such as retinopathy, limb amputation, cardiovascular diseases, and neuronal disease, if left untreated.  Consequently, it becomes crucial to take precautionary measures to avoid/predict the occurrence of diabetes.  Machine learning approaches have been proposed and evaluated in the literature for diabetes prediction.  This paper proposes an IoT-edge-Artificial Intelligence (AI)-blockchain system for diabetes prediction based on risk factors.  The proposed system is underpinned by the blockchain to obtain a cohesive view of the risk factors data from patients across different hospitals and to ensure security and privacy of the user's data.  Furthermore, we provide a comparative analysis of different medical sensors, devices, and methods to measure and collect the risk factors values in the system.  Numerical experiments and comparative analysis were carried out between our proposed system, using the most accurate random forest (RF) model, and the two most used state-of-the-art machine learning approaches, Logistic Regression (LR) and Support Vector Machine (SVM), using three real-life diabetes datasets.  The results show that the proposed system using RF predicts diabetes with 4.57\% more accuracy on average compared to LR and SVM, with 2.87 times more execution time.  Data balancing without feature selection does not show significant improvement.  The performance is improved by 1.14\% and 0.02\% after feature selection for PIMA Indian and Sylhet datasets respectively, while it reduces by 0.89\% for MIMIC III. 
\end{abstract}


\end{frontmatter}


\section{Introduction}

Diabetes Mellitus, commonly referred to as diabetes, is one of the top 10 leading causes of death globally \cite{Thetop1095:online}.  It is a metabolic disease in which the body does not produce enough insulin or body cells do not appropriately respond to insulin, leading to increased blood sugar levels \cite{ismail2021association}.  There are three main types of diabetes, type 1 and type 2 diabetes mellitus, and gestational diabetes \cite{ismail2022type}.  According to a report by the International Diabetes Federation, 537 million adults (i.e., 1 in every 10 people), between the ages of 20-79 years, worldwide were having diabetes in 2021 \cite{IDFDiabe18:online}.  Furthermore, this number is predicted to reach 643 million by 2030 and 783 million by 2045.  In 2021, diabetes was responsible for 6.7 million deaths and caused at least USD 966 billion in health expenditure \cite{IDFDiabe18:online}.

The etiopathology of type 2 diabetes mellitus has been linked to dynamic interactions between lifestyle, medical conditions, hereditary, psychosocial, and demographic risk factors \cite{ismail2022type}.  Diabetes if not treated at an early stage can lead to severe complications such as retinopathy, limb amputation, cardiovascular diseases, and neuronal disease \cite{itani2017short}.  In 2021, over 240 million adults with diabetes were undiagnosed (i.e., almost 1 in 2 diabetic) \cite{Factsfig78:online}.  Consequently, machine learning-based diabetes prediction has gained increased attention in the literature \cite{ahmad2021investigating, kopitar2020early, deberneh2021prediction, syed2020machine, joshi2021predicting, chang2022pima, zhang2020machine, lu2022patient} for better prognosis/diagnosis support to the medical health professionals and public health organizations \cite{ismail2020requirements}.  Disparate work in literature focuses on evaluating machine learning algorithms for different diabetes datasets under non-unified experimental setups.  However, to the best of our knowledge, no work proposes an end-to-end IoT-edge-Artificial Intelligence (AI)-blockchain integrated computing system for diabetes monitoring and prediction.  This paper aims to address this void.  The proposed system analyzes diabetes risk factors using medical sensors/devices and predicts the incidence of diabetes in an individual using the most accurate machine learning model.  Furthermore, the proposed system employs edge computing to transform the risk factors data collected from IoT devices and send preprocessed data to the blockchain.  Blockchain \cite{ismail2021scoping} stores the medical records of the patients as well the machine learning model parameters and prediction results in a distributed and replicated ledger. This is based on the potential of blockchain in the healthcare industry \cite{ismail2020performance, ismail2020blockhrCS, ismail2020blockhr}. The consensus, replication, traceability, and distributed features of blockchain aid in security, privacy, audit trail, transparency, and trust in the proposed system.

The main contributions of this paper are as follows.
\begin{itemize}
\item We propose an end-to-end automated IoT-edge-AI-blockchain system for diabetes prediction based on risk factors.
\item We present a comparative list of medical sensors, devices, and methods used to measure the values of diabetes risk factors; hypertension, obesity, cholesterol level, depression, serum uric acid, sleep duration, physical activity, and glucose level.
\item We propose an implementation workflow for the proposed system.
\item The performance of the proposed system is evaluated and compared with the most used machine learning approaches for diabetes prediction in terms of accuracy, precision, recall, F-measure, Area Under the Receiver Operating Characteristics (ROC) Curve (AUC), and execution time.
\end{itemize}

The rest of the paper is organized as follows.  Section 2 summarizes the related work on machine learning-based diabetes prediction.  The proposed automated end-to-end IoT-edge-AI-blockchain system for diabetes mellitus prediction is explained in Section 3.  Section 4 discusses the implementation of the proposed system.  Numerical experiments and comparative performance results are provided in Section 5.  Finally, Section 6 concludes the paper with future research directions.

\section{Related Work}

Several works in the literature have used machine and deep learning algorithms for diabetes prediction \cite{ahmad2021investigating, kopitar2020early, deberneh2021prediction, syed2020machine, joshi2021predicting, chang2022pima, zhang2020machine, lu2022patient}.  Table \ref{table:related work} summarizes these works and presents the dataset, preprocessing techniques, feature selection approaches, and machine/deep learning algorithms used in each work.  However, these works only focus on stand-alone diabetes prediction and do not propose an end-to-end diabetes prediction system.  In contrast, we propose a secure and privacy-preserving end-to-end integrated IoT-edge-AI-blockchain monitoring system for diabetes prediction.

\begin{landscape}
\begin{scriptsize}
\begin{longtable}{p{2cm}p{4cm}p{4cm}p{2cm}p{2cm}p{2cm}p{2cm}p{3cm}}
\caption{Summary of Related Work on Diabetes Prediction.}
\label{table:related work}\\

\hline
\rowcolor[HTML]{C0C0C0} 
\multicolumn{1}{|c|}{\cellcolor[HTML]{C0C0C0}\textbf{Work}}        & \multicolumn{1}{p{4cm}|}{\cellcolor[HTML]{C0C0C0}\textbf{Dataset}}                                                                                                                                                                  & \multicolumn{1}{p{4cm}|}{\cellcolor[HTML]{C0C0C0}\textbf{Features$^{\S}$}}                                                                                                                                                                                                                                   & \multicolumn{1}{p{2cm}|}{\cellcolor[HTML]{C0C0C0}\textbf{Observations$^{\S}$}}                                                                                                                                                                                          & \multicolumn{1}{p{2cm}|}{\cellcolor[HTML]{C0C0C0}\textbf{Data balancing}}                 & \multicolumn{1}{p{2cm}|}{\cellcolor[HTML]{C0C0C0}\textbf{Feature selection}}                                      & \multicolumn{1}{p{2cm}|}{\cellcolor[HTML]{C0C0C0}\textbf{Algorithms}}                                                                         & \multicolumn{1}{p{3cm}|}{\cellcolor[HTML]{C0C0C0}\textbf{Evaluation metrics}}                                                                                                        \\ \hline

\endfirsthead
\multicolumn{8}{@{}l}{\ldots continued}\\\hline

\multicolumn{1}{|c|}{\cellcolor[HTML]{C0C0C0}\textbf{Work}}        & \multicolumn{1}{p{4cm}|}{\cellcolor[HTML]{C0C0C0}\textbf{Dataset}}                                                                                                                                                                  & \multicolumn{1}{p{4cm}|}{\cellcolor[HTML]{C0C0C0}\textbf{Features$^{\S}$}}                                                                                                                                                                                                                                   & \multicolumn{1}{p{2cm}|}{\cellcolor[HTML]{C0C0C0}\textbf{Observations$^{\S}$}}                                                                                                                                                                                          & \multicolumn{1}{p{2cm}|}{\cellcolor[HTML]{C0C0C0}\textbf{Data balancing}}                 & \multicolumn{1}{p{2cm}|}{\cellcolor[HTML]{C0C0C0}\textbf{Feature selection}}                                      & \multicolumn{1}{p{2cm}|}{\cellcolor[HTML]{C0C0C0}\textbf{Algorithms}}                                                                         & \multicolumn{1}{p{3cm}|}{\cellcolor[HTML]{C0C0C0}\textbf{Evaluation metrics}}                                                                                                        \\ \hline

\endhead 
\hline
\multicolumn{8}{r@{}}{continued \ldots}\\
\endfoot
\hline
\endlastfoot

\multicolumn{1}{|l|}{\cite{ahmad2021investigating}}                                            & \multicolumn{1}{p{4cm}|}{\cellcolor[HTML]{FFFFFF}Private: EHRs acquired from 5 hospitals in Saudi Arabia between   2016 – 2018}                                                                                                     & \multicolumn{1}{p{4cm}|}{\cellcolor[HTML]{FFFFFF}DOB, gender, height weight, hypertension, fasting plasma   glucose, haemoglobin A1C, HDL, LDL, physical activity, diagnosis start date,   and primary and secondary diagnosis codes and full names}                                                  & \multicolumn{1}{p{2cm}|}{\cellcolor[HTML]{FFFFFF}3000 patients}                                                                                                                                                                                                  & \multicolumn{1}{p{2cm}|}{\cellcolor[HTML]{FFFFFF}Data is already balanced}                & \multicolumn{1}{p{2cm}|}{\cellcolor[HTML]{FFFFFF}Permutation importance and hierarchical clustering}              & \multicolumn{1}{p{2cm}|}{\cellcolor[HTML]{FFFFFF}LR, SVM, DT, RF$^{\blacklozenge}$, EMV$^{*}$}                                                                       & \multicolumn{1}{p{3cm}|}{\cellcolor[HTML]{FFFFFF}Accuracy, precision, recall, and F-measure}                                                                                         \\ \hline

\multicolumn{1}{|l|}{\cite{kopitar2020early}}                                            & \multicolumn{1}{p{4cm}|}{\cellcolor[HTML]{FFFFFF}Private: EHRs data collected at preventive healthcare   examinations of healthy population in 10 Slovenian primary healthcare   institutions}                                      & \multicolumn{1}{p{4cm}|}{\cellcolor[HTML]{FFFFFF}Related to FINDRISC questionnaire and medical history}                                                                                                                                                                                               & \multicolumn{1}{p{2cm}|}{\cellcolor[HTML]{FFFFFF}27050 patients}                                                                                                                                                                                                 & \multicolumn{1}{p{2cm}|}{\cellcolor[HTML]{FFFFFF}\xmark}                                       & \multicolumn{1}{p{2cm}|}{\cellcolor[HTML]{FFFFFF}\xmark}                                                               & \multicolumn{1}{p{2cm}|}{\cellcolor[HTML]{FFFFFF}Linear regression$^{\blacklozenge}$, Glmnet, RF, XGBoost, and lightGBM}                                        & \multicolumn{1}{p{3cm}|}{\cellcolor[HTML]{FFFFFF}AUC and RMSE}                                                                                                                       \\ \hline

\multicolumn{1}{|l|}{\cite{deberneh2021prediction}}                                            & \multicolumn{1}{p{4cm}|}{\cellcolor[HTML]{FFFFFF}Private: EHRs collected between 2013 – 2018 from a private   medical institute, Hanaro Medical Foundation, in Seoul (South Korea)}                                                 & \multicolumn{1}{p{4cm}|}{\cellcolor[HTML]{FFFFFF}Related to blood test, anthropometric measurements,   diagnostics results, and questionnaire answers}                                                                                                                                                & \multicolumn{1}{p{2cm}|}{\cellcolor[HTML]{FFFFFF}253359 subjects (68.1\% normal, 4.3\% diabetics, and 27.6\%   prediabetes)}                                                                                                                                     & \multicolumn{1}{p{2cm}|}{\cellcolor[HTML]{FFFFFF}Majority under-sampling and SMOTE}       & \multicolumn{1}{p{2cm}|}{\cellcolor[HTML]{FFFFFF}ANOVA, chi-squared test and recursive feature elimination}       & \multicolumn{1}{p{2cm}|}{\cellcolor[HTML]{FFFFFF}LR, RF$^{\blacklozenge}$, SVM, XGBoost, stacking$^{\dagger}$, soft voting$^{\dagger}$,   and confusion matrix-based ensemble$^{\dagger}$}       & \multicolumn{1}{p{3cm}|}{\cellcolor[HTML]{FFFFFF}Accuracy, precision, recall, F-measure, MCC, and KC}                                                                                \\ \hline

\multicolumn{1}{|l|}{\cite{syed2020machine}}                                           & \multicolumn{1}{p{4cm}|}{\cellcolor[HTML]{FFFFFF}\parbox{4cm}{D1: Cross-sectional diabetes survey in Saudi Arabia\\   D2: NHANES\\   D3: PIMA Indian}}                                            & \multicolumn{1}{p{4cm}|}{\cellcolor[HTML]{FFFFFF}\parbox{4cm}{D1: region, age, gender, BMI, waist size, physical activity,   diet, blood pressure, and family history of diabetes\\  D2: smoking, diet, blood pressure, BMI, gender, and region\\   D3: $^{\ddagger}$}}        & \multicolumn{1}{p{2cm}|}{\cellcolor[HTML]{FFFFFF}\parbox{2cm}{D1: 4896   (990 diabetics and 3906 non-diabetics)\\   D2: 4918   (1709 prediabetes and 3209 diabetics)\\    D3: 768 (268 diabetics and 500 non-diabetics)}}      & \multicolumn{1}{p{2cm}|}{\cellcolor[HTML]{FFFFFF}SMOTE}                                   & \multicolumn{1}{p{2cm}|}{\cellcolor[HTML]{FFFFFF}Pearson chi-square test}                                         & \multicolumn{1}{p{2cm}|}{\cellcolor[HTML]{FFFFFF}BPM, AP, DF$^{\blacklozenge}$, LD-SVM, DJ, boosted DT, and NN}                                                 & \multicolumn{1}{p{3cm}|}{\cellcolor[HTML]{FFFFFF}Accuracy, precision, recall, F-measure, and AUC}                                                                                    \\ \hline

\multicolumn{1}{|l|}{\cite{joshi2021predicting}}                                           & \multicolumn{1}{p{4cm}|}{\cellcolor[HTML]{FFFFFF}PIMA Indian}                                                                                                                                                                       & \multicolumn{1}{p{4cm}|}{\cellcolor[HTML]{FFFFFF}$^{\ddagger}$}                                                                                                                                                                                                                                                   & \multicolumn{1}{p{2cm}|}{\cellcolor[HTML]{FFFFFF}768 (268 diabetics and 500 non-diabetics)}                                                                                                                                                                      & \multicolumn{1}{p{2cm}|}{\cellcolor[HTML]{FFFFFF}\xmark}                                       & \multicolumn{1}{p{2cm}|}{\cellcolor[HTML]{FFFFFF}Different combinations based on manual inspection}               & \multicolumn{1}{p{2cm}|}{\cellcolor[HTML]{FFFFFF}LR$^{\blacklozenge}$ and DT}                                                                                   & \multicolumn{1}{p{3cm}|}{\cellcolor[HTML]{FFFFFF}Accuracy, error rate, AIC, BIC, R2, and log likelihood}                                                                             \\ \hline

\multicolumn{1}{|l|}{\cite{chang2022pima}}                                           & \multicolumn{1}{p{4cm}|}{\cellcolor[HTML]{FFFFFF}PIMA Indian}                                                                                                                                                                       & \multicolumn{1}{p{4cm}|}{\cellcolor[HTML]{FFFFFF}$^{\ddagger}$}                                                                                                                                                                                                                                                   & \multicolumn{1}{p{2cm}|}{\cellcolor[HTML]{FFFFFF}768 (268 diabetics and 500 non-diabetics)}                                                                                                                                                                      & \multicolumn{1}{p{2cm}|}{\cellcolor[HTML]{FFFFFF}\xmark}                                       & \multicolumn{1}{p{2cm}|}{\cellcolor[HTML]{FFFFFF}PCA, k-means clustering, and importance ranking}                 & \multicolumn{1}{p{2cm}|}{\cellcolor[HTML]{FFFFFF}NB, RF$^{\blacklozenge}$, and DT}                                                                              & \multicolumn{1}{p{3cm}|}{\cellcolor[HTML]{FFFFFF}Accuracy, precision, sensitivity, specificity, F-measure,   and AUC}                                                                \\ \hline

\multicolumn{1}{|l|}{\cite{zhang2020machine}}                                           & \multicolumn{1}{p{4cm}|}{\cellcolor[HTML]{FFFFFF}Henan rural cohort study: participants aged between 18 – 79   years were recruited from five rural areas in Henan province of China between   July 2015 and September 2017}        & \multicolumn{1}{p{4cm}|}{\cellcolor[HTML]{FFFFFF}Related to socio-demographic characteristics, information on   physical examination, and laboratory tests}                                                                                                                                           & \multicolumn{1}{p{2cm}|}{\cellcolor[HTML]{FFFFFF}39259 participants}                                                                                                                                                                                             & \multicolumn{1}{p{2cm}|}{\cellcolor[HTML]{FFFFFF}SMOTE}                                   & \multicolumn{1}{p{2cm}|}{\cellcolor[HTML]{FFFFFF}Iterative approach}                                              & \multicolumn{1}{p{2cm}|}{\cellcolor[HTML]{FFFFFF}LR, CART, ANN, SVM, RF$^{\blacklozenge}$, and GBM}                                                             & \multicolumn{1}{p{3cm}|}{\cellcolor[HTML]{FFFFFF}AUC, sensitivity, specificity, positive prediction value,   negative prediction value, and area under precision-recall curve}       \\ \hline

\multicolumn{1}{|l|}{\cite{lu2022patient}}                                           & \multicolumn{1}{p{4cm}|}{\cellcolor[HTML]{FFFFFF}CBHS health funds company in Australia: hospital admissions   data between 1995 – 2018}                                                                                            & \multicolumn{1}{p{4cm}|}{\cellcolor[HTML]{FFFFFF}Age, gender, and smoking status}                                                                                                                                                                                                                     & \multicolumn{1}{p{2cm}|}{\cellcolor[HTML]{FFFFFF}2056 (1028 diabetics and 1028 non-diabetics)}                                                                                                                                                                   & \multicolumn{1}{p{2cm}|}{\cellcolor[HTML]{FFFFFF}Data is already balanced}                & \multicolumn{1}{p{2cm}|}{\cellcolor[HTML]{FFFFFF}\xmark}                                                               & \multicolumn{1}{p{2cm}|}{\cellcolor[HTML]{FFFFFF}LR, kNN, SVM, NB, DT, RF$^{\blacklozenge}$, XGBoost, and ANN}                                                  & \multicolumn{1}{p{3cm}|}{\cellcolor[HTML]{FFFFFF}Accuracy, precision, recall, F-measure, and AUC}                                                                                    \\ \hline

\multicolumn{8}{p{21cm}}{\tiny{EHRs – Electronic Health Records; $^{\blacklozenge}$ - outperforming model; LR – Logistic Regression; SVM – Support Vector Machine; DT – Decision Tree; RF – Random Forest; EMV – Ensemble Majority Voting; $^{*}$ - EMV consists of LR, SVM, and DT; Glmnet – Regularized Generalized Linear Model; XGBoost – Extreme Gradient Boosting; lightGBM – light Gradient Boosting Machine; $^{\dagger}$ - ensemble algorithms use LR, RF, SVM, and XGBoost; BPM – Bayes Point Machine; AP – Average Perceptron; DF – Decision Forest; LD-SVM – Locally Deep SVM; DJ – Decision Jungle; NN – Neural Network; NB – Naïve Bayes; CART – Classification and Regression Tree; GBM – Gradient Boosting Machine; kNN – k Nearest Neighbor; AUC – Area Under the ROC Curve; RMSE – Root Mean Squared Error; MCC – Mathews Correlation Coefficient; KC – Kappa's Coefficient; AIC – Akaike's Information Criteria; BIC – Bayesian Information Criteria; PCA – Principal Component Analysis; SMOTE – Synthetic Minority Oversampling Technique; NHANES – National Health and Nutrition Examination Survey; HDL – High Density Lipoprotein; LDL – Low Density Lipoprotein; $^{\ddagger}$ - Number of times pregnant, plasma glucose concentration at 2h oral glucose tolerance test, diastolic pressure, triceps skin fold thickness, 2-h serum insulin, BMI, diabetes pedigree function, and age; BMI – Body Mass Index; DOB – Date of Birth; D – Dataset; $^{\S}$ - Before data preprocessing; \xmark - Not performed}}
\end{longtable}
\end{scriptsize}
\end{landscape}

\section{Proposed Automated End-to-End Integrated IoT-Edge-Artificial Intelligence-Blockchain Monitoring System for Diabetes Mellitus Prediction}
\begin{figure}[H]
\centering
  \includegraphics[width=\linewidth]{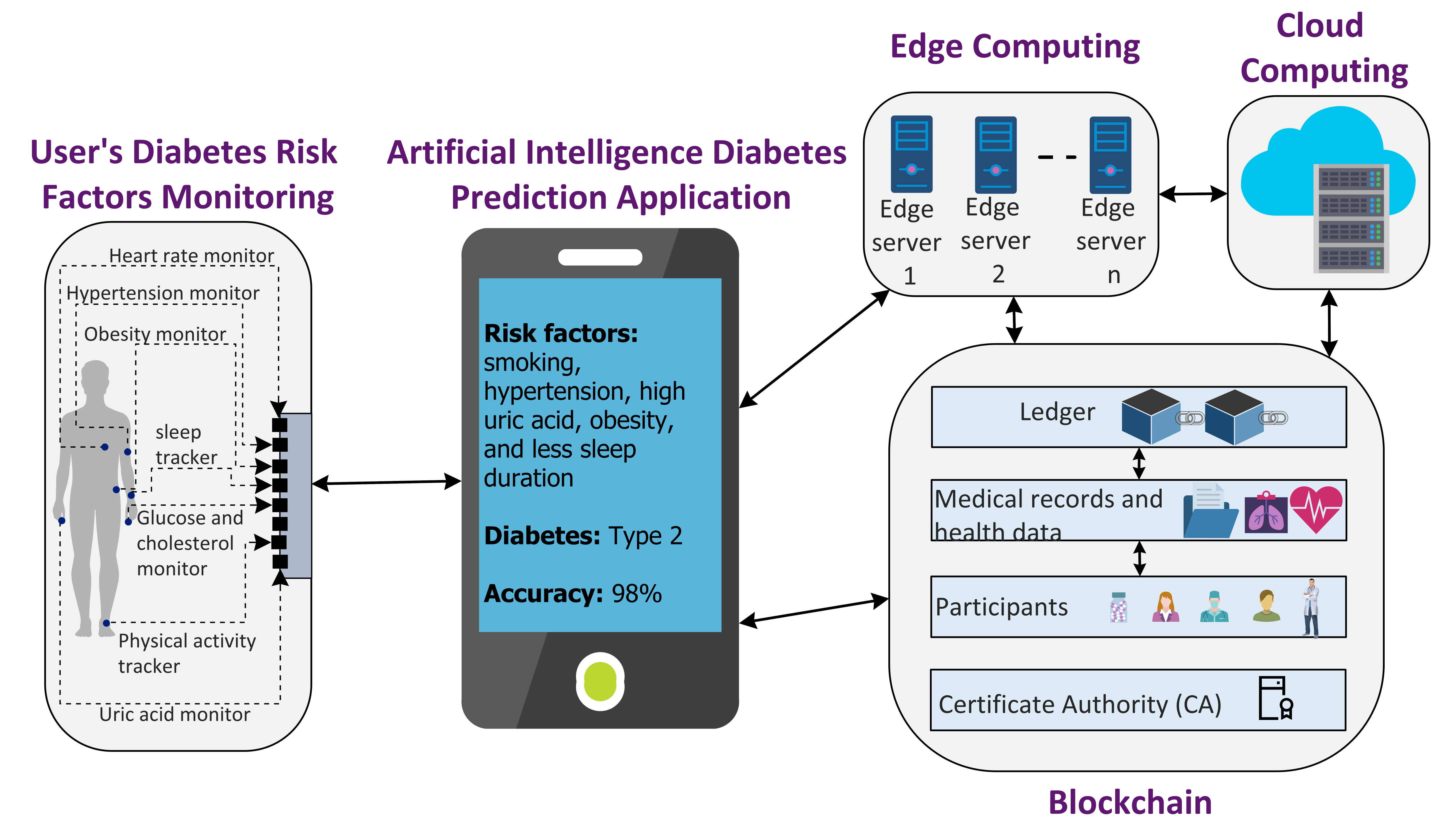}
  \caption{Architecture Overview of Proposed Automated End-to-End Integrated IoT-Edge-Artificial Intelligence-Blockchain Monitoring System for Diabetes Mellitus Prediction.}
  \label{fig:Architecture overview}
\end{figure}

The overall architecture of our proposed end-to-end system for diabetes prediction is presented in Figure \ref{fig:Architecture overview}. The main components of the architecture are explained in the following subsections.

\graphicspath{{./Images/}}

\subsection{User's Diabetes Risk Factors Monitoring}

Diabetes, i.e., increased glucose levels, is associated with different demographic, psychosocial, hereditary, medical conditions, and lifestyle-related risk factors \cite{ismail2021association, ismail2022type}.  The values of these risk factors can be either self-reported by the users (i.e., patients/external participants) or measured using biosensors, wearable devices, or medical tests.  The self-reported risk factors are age, gender, ethnicity, family history of diabetes, smoking, and alcohol consumption. Age, gender, and ethnicity are reported on the first visit to the hospital.  Family history of diabetes is reported on every visit to the hospital. Smoking and alcohol consumption are reported daily. The measurable risk factors are hypertension, obesity, cholesterol level, depression, serum uric acid, sleep duration, physical activity, and glucose levels.  Hypertension, obesity, serum uric acid, sleep duration, physical activity, and glucose levels are acquired daily.  Cholesterol level and depression are collected on every visit to the hospital.  The measured risk factors data are sent to the mobile phone.  A user communicates with the mobile application to identify the risk of incident diabetes.

In the following, we compare different biosensors, devices, and methods used to acquire the values of measurable risk factors.

\begin{itemize}

\item \textit{Hypertension Monitoring:} Hypertension is a medical condition where the blood pressure in the arteries remains elevated, i.e., a systolic blood pressure greater than or equal to 140 mmHg and a diastolic blood pressure greater than or equal to 90 mmHg \cite{kim2015hypertension}.  It increases the risk of developing diabetes.  Table \ref{table:hypertension} lists different hypertension monitors along with their measurement method, accuracy, and approximate cost in US dollars.  As shown in the table, Omron Evolv (HEM-7600T-E) \cite{OMRONHea42:online} has the best performance whereas Omron M3 Comfort (HEM-7134-E) \cite{Medaval26:online} has the least cost.

\begin{scriptsize}
\begin{longtable}{|p{3cm}|p{8.8cm}|p{3.4cm}|}
\caption{Comparison Between Different Hypertension Monitoring Devices.}
\label{table:hypertension}\\
\hline
\rowcolor{Gray}
\textbf{Device}& \textbf{Performance}& \textbf{Approximate cost (in US Dollars)}\\
\hline

Omron Evolv (HEM-7600T-E) \cite{OMRONHea42:online}
& \parbox{9cm}{Mean difference compared to standard mercury sphygmomanometer test \cite{takahashi2019validation}:\\
-0.1 ± 5.0 mmHg (for systolic blood pressure)\\
-0.2 ± 4.1 mmHg (for diastolic blood pressure)}
&\multicolumn{1}{|c|}{136 \cite{BLOODPRE92:online}}\\ \hline

Omron M3 Comfort (HEM-7134-E) \cite{Medaval26:online}
& \parbox{9cm}{Mean difference compared to standard mercury sphygmomanometer test \cite{takahashi2019validation}:\\
-0.9 ± 5.4 mmHg (for systolic blood pressure)\\
-0.6 ± 4.7 mmHg (for diastolic blood pressure)}
&\multicolumn{1}{|c|}{63.16 \cite{OmronM3U36:online}}\\ \hline

Omron (HEM-9210T) \cite{HEM9210T45:online}
& \parbox{9cm}{Mean difference compared to standard mercury sphygmomanometer test \cite{takahashi2019validation}:\\
-2.1 ± 4.7 mmHg (for systolic blood pressure)\\
-1.2 ± 4.1 mmHg (for diastolic blood pressure)}
&\multicolumn{1}{|c|}{Not available}\\ \hline

Mobil-O-Graph \cite{MobilOGr33:online}
& \parbox{9cm}{Mean difference compared to standard mercury sphygmomanometer test \cite{wei2010validation}:\\
-2.2 ± 7.3 mmHg (for systolic blood pressure)\\
-0.4 ± 6.1 mmHg (for diastolic blood pressure)}
&\multicolumn{1}{|c|}{1365.86 \cite{IEMMobil4:online}}\\ \hline

\end{longtable}
\end{scriptsize}

\item \textit{Obesity Monitoring:} Obesity is characterized by an excessive amount of body fat and is often defined in terms of Body Mass Index (BMI), waist circumference, and/or waist-hip ratio \cite{neovius2005bmi}.  It is strongly associated with the prevalence of type 2 diabetes.  Table \ref{table:obesity} shows different methods and devices used to measure obesity with their strengths and weaknesses.

\begin{scriptsize}
\begin{longtable}{|p{3cm}|p{4cm}|p{8cm}|}
\caption{Comparison Between Different Obesity Monitoring Methods and Devices.}
\label{table:obesity}\\
\hline
\rowcolor{Gray}
\textbf{Method/device}& \textbf{Strengths}& \textbf{Weaknesses}\\
\hline

Statistical BMI calculation \cite{bjelica2021trajectories}
& Quick, cost-effective, and easy
& Not accurate for elderly, muscular, and pregnant individuals \\ \hline

Skinfold calipers \cite{martin1985prediction}
& Easy to use, portable, and cost-effective
& Accuracy depends on the skill of the person using the caliper \\ \hline

Smart weighing scales \cite{frija2021accuracy}
& Quick and easy
& Reliability of the result depends on the condition of the individual whose measurement is taken (for instance, hydrated or dehydrated), some accurate scales are costly \\ \hline

Hydrodensitometry \cite{brodie1998body}
& Accurate and reliable
& Costly and not suitable for children and elderly people as it requires the individual to be submerged in water for 5-7 seconds repeatedly 2-3 times \\ \hline

Air displacement plethysmography \cite{ginde2005air}
& Quick, accurate, reliable, and suitable for any age
& Costly \\ \hline

Dual energy x-ray absorptiometry \cite{kaul2012dual}
& Quick, precise, and reliable
& Costly \\ \hline

\end{longtable}
\end{scriptsize}

\item \textit{Cholesterol Level Monitoring:} Abnormal level of cholesterol and triglycerides increases the risk of type 2 diabetes prevalence.  In particular, low level of high-density lipoproteins (HDL) and elevated level of low-density lipoproteins (LDL) leads to the development of diabetes \cite{kawamoto2011relationships}.  The standard method to measure the cholesterol level is the lipid panel test (also known as lipid profile test) \cite{birtcher2004measurement}.  This test determines the levels of triglycerides, total, LDL, and HDL cholesterols in an individual.  Recently, several portable devices have been developed to measure cholesterol levels.  Table \ref{table:cholesterol} provides a summary of the performance and cost of these devices.  As shown in the table, EasyTouch \cite{glucomet5:online} is more economical compared to BeneCheck Plus \cite{httpswww35:online}.

\begin{scriptsize}
\begin{longtable}{|p{4cm}|p{4cm}|p{4cm}|}
\caption{Comparison Between Different Cholesterol Level Monitoring Devices.}
\label{table:cholesterol}\\
\hline
\rowcolor{Gray}
\textbf{Device}& \textbf{Performance (Coefficient of variation)}& \textbf{Approximate cost (in US Dollars)}\\
\hline

EasyTouch \cite{glucomet5:online}
& Not reported
& 60 \cite{Amazonco2:online} \\ \hline

BeneCheck Plus \cite{httpswww35:online}
& Not reported
& 136 \cite{Benechec5:online} \\ \hline

\end{longtable}
\end{scriptsize}

\item \textit{Depression Monitoring:} Depression is a medical condition that negatively affects the feelings, thoughts, and actions of an individual.  It has a strong association with the prevalence of type 2 diabetes \cite{golden2004depressive}.  Depression is generally measured using clinical rating scales such as Beck's Depression Inventory (BDI), Center for Epidemiological Studies – Depression scale (CES-D), and Zung Self-Rating Depression Scale (SDS) \cite{shafer2006meta}.  A BDI score $\geq11$, CES-D score $\geq8$, or SDS score $>39$  increases the risk of developing diabetes.

\item \textit{Serum Uric Acid Monitoring:} Serum uric acid is a waste product generated by the body during the purines breakdown process.  A serum uric acid level $>$370 $\mu$mol/l is associated with a risk of developing type 2 diabetes \cite{dehghan2008high}. A uric acid test is commonly used to measure the amount of uric acid either using blood or urine samples \cite{UricAcid46:online}.  Recently, several test meters have been introduced to measure serum uric acid levels.  Table \ref{table:uric acid} shows a comparison between these meters.  As shown in the table, HumaSens$^{plus}$ \cite{fabre2018accuracy} is the most economical compared to other methods/devices for serum uric acid monitoring.

\begin{scriptsize}
\begin{longtable}{|p{4cm}|p{6cm}|p{4cm}|}
\caption{Comparison Between Different Serum Uric Acid Monitoring Methods and Devices.}
\label{table:uric acid}\\
\hline
\rowcolor{Gray}
\textbf{Method/device}& \textbf{Performance (Coefficient of variation)}& \textbf{Approximate cost (in US Dollars)}\\
\hline

Smartphone as electro-chemical analyzer \cite{guo2016uric}
& \parbox{6cm}{Low concentration: 4.1\%$^{*}$ \\ Mid concentration: 2.47\%$^{*}$ \\ High concentration: 1.87\%$^{*}$ \cite{guo2016uric}}
& Not available \\ \hline

EasyTouch \cite{glucomet5:online}
&27.2\% \cite{paraskos2016analytical} (Not acceptable)$^{\dagger}$
& 60 \cite{Amazonco2:online} \\ \hline

UAsure \cite{kuo2002portable}
&25.9\% \cite{paraskos2016analytical} (Not acceptable)$^{\dagger}$
& 64 \cite{BuyUASur5:online} \\ \hline

BeneCheck Plus \cite{httpswww35:online}
&9.5\% \cite{paraskos2016analytical} (Acceptable)$^{\dagger}$
& 136 \cite{Benechec5:online} \\ \hline

HumaSens$^{plus}$ \cite{fabre2018accuracy}
&11.5\% \cite{paraskos2016analytical} (Acceptable)$^{\dagger}$
& 52 \cite{HumaSens65:online} \\ \hline

Liquid chromatography mass spectrometry \cite{kim2009sensitive}
&0.01 – 3.37\%$^{*}$ \cite{kim2009sensitive}
& Not available \\ \hline
\multicolumn{3}{p{14cm}}{$^{*}$Average; $^{\dagger}$According to College of American Pathologists}
\end{longtable}
\end{scriptsize}

\item \textit{Sleep Duration Monitoring:} The quantity of sleep during night time is highly associated with the prevalence of type 2 diabetes \cite{ismail2021association}.  Compared to 6-8 hours of night time sleep, a shorter sleep duration ($<$6 hours/night) and a longer sleep duration ($>$8 hours/night) are associated with diabetes.  In addition, day-time napping can lead to the prevalence of diabetes. Table \ref{table:sleep} summarizes different tests, devices, and applications used for tracking sleep. It shows the performance of each test/device/application along with its cost in USD.  As shown in the table, Fitbit Charge HR \cite{ChargeHR91:online} costs the least compared to other tests/devices for monitoring sleep duration.

\begin{scriptsize}
\begin{longtable}{|p{2.5cm}|p{3.5cm}|p{6cm}|p{2cm}|}
\caption{Comparison Between Different Sleep Duration Monitoring Tests and Devices.}
\label{table:sleep}\\
\hline
\rowcolor{Gray}
\textbf{Type} & \textbf{Test/device}& \textbf{Performance}& \textbf{Approximate cost (in US Dollars)}\\
\hline

Non-invasive
& Polysomnography test \cite{kang2017validity}
& \parbox{5cm}{Sensitivity: 0.957$^{*}$ \\ Specificity: 0.532$^{*}$ \\ Accuracy: 0.904$^{*}$ \\ Cohen's kappa: 0.495$^{*}$ \cite{kang2017validity}}
& 943 – 2,798 \cite{HowMuchD49:online} \\ \hline

Wearable
& ŌURA ring \cite{OuraRing38:online}
& \parbox{5cm}{Sensitivity (to detect sleep): 96\%$^{*}$ \\ Specificity (to detect wake): 48\%$^{*}$ \cite{de2019sleep}}
& 299-399 \cite{OuraRing38:online} \\ \hline

Wearable
& Fitbit Flex \cite{ShopFitb94:online}
& \parbox{5cm}{97.46\% accuracy \cite{kaewkannate2016comparison}}
& 100 \cite{kaewkannate2016comparison} \\ \hline

Wearable
& Fitbit Charge HR \cite{ChargeHR91:online}
& \parbox{5cm}{Overestimates the sleep duration \cite{lee2017comparison}}
& 65.39 \cite{FitbitCh79:online} \\ \hline

Wearable
& Polar A370 fitness tracker \cite{PolarA3727:online}
& \parbox{5cm}{\textbf{Age group (mean $\pm$ SD): 11 $\pm$ 0.8} \\ Sensitivity$^{*}$: 0.93 \\ Specificity$^{*}$: 0.77 \\ Accuracy$^{*}$: 0.91\\
\textbf{Age group (mean $\pm$ SD): 17.8 $\pm$ 1.8} \\ Sensitivity$^{*}$: 0.91 \\ Specificity$^{*}$: 0.83 \\ Accuracy$^{*}$: 0.90 \cite{pesonen2018validity}}
& 163 \cite{PolarA3727:online} \\ \hline

Wearable
& Actiwatch 2 \cite{Actiwatc21:online}
& \parbox{5cm}{\textbf{Age group (mean $\pm$ SD): 11 $\pm$ 0.8} \\Sensitivity$^{*}$: 0.93 \\ Specificity$^{*}$: 0.68 \\ Accuracy$^{*}$: 0.90\\
\textbf{Age group (mean $\pm$ SD): 17.8 $\pm$ 1.8} \\ Sensitivity$^{*}$: 0.93 \\ Specificity$^{*}$: 0.58 \\ Accuracy$^{*}$: 0.89 \cite{pesonen2018validity}}
& Not available \\ \hline

Wearable
& Fitbit Alta HR \cite{ShopFitb8:online}
& \parbox{5cm}{All sleep \\ Sensitivity: 0.96 $\pm$ 0.02 \\ Specificity: 0.58 $\pm$ 0.16 \\ Accuracy: 0.90 $\pm$ 0.04 \cite{cook2019ability}}
& 270 \cite{Amazonco80:online} \\ \hline

Wearable
& Withings Pulse \cite{Healthfi40:online}
& \parbox{5cm}{98.1\% accuracy \cite{kaewkannate2016comparison}}
& 100 \cite{Healthfi40:online} \\ \hline

Wearable
& Misfit Shine \cite{kaewkannate2016comparison}]
& \parbox{5cm}{96\% accuracy \cite{kaewkannate2016comparison}}
& 100 \cite{kaewkannate2016comparison} \\ \hline

Wearable
& Jawbone Up24 \cite{Riseandf93:online}
& \parbox{5cm}{97.23\% accuracy \cite{kaewkannate2016comparison}}
& 100 \cite{kaewkannate2016comparison} \\ \hline

Non-wearable
& EMFIT Quantified Sleep \cite{CONTACTF36:online}
& \parbox{5cm}{Overestimates total sleep time and underestimates wake after sleep \cite{kholghi2021validation}}
& Not available \\ \hline

Mobile application
& Sleep Cycle \cite{robbins2020four}
& \parbox{5cm}{Not reported}
& Free\\ \hline

\multicolumn{4}{p{14cm}}{$^{*}$Average}

\end{longtable}
\end{scriptsize}

\item \textit{Physical Activity Monitoring:} Physical inactivity can lead to obesity and depression, resulting in the prevalence of type 2 diabetes \cite{ismail2021association}.  An individual performing 30-60 minutes of exercise 3 – 4 times/week can be considered as physically active.  Table \ref{table:physical activity} summarizes different devices to track physical activity.  As shown in the table, Sportline 340 Strider \cite{Sportlin33:online} outperforms other physical activity monitors in terms of performance and cost.

\begin{scriptsize}
\begin{longtable}{|p{3.5cm}|p{3.5cm}|p{5cm}|p{2cm}|}
\caption{Comparison Between Different Physical Activity Monitoring Devices.}
\label{table:physical activity}\\
\hline
\rowcolor{Gray}
\textbf{Type} & \textbf{Device}& \textbf{Performance (Accuracy)}& \textbf{Approximate cost (in US Dollars)}\\
\hline

Waist-based
& Fitbit One \cite{ShopFitb50:online}
& $>$90\% \cite{battenberg2017accuracy}
& 70 \cite{FitbitOn49:online} \\ \hline

Waist-based
& Omron HJ-321 \cite{HJ321IMW88:online}
& $>$90\% \cite{battenberg2017accuracy}
& 67.25 \cite{OmronHJ32:online} \\ \hline

Waist-based
& Sportline 340 Strider \cite{Sportlin33:online}
& $>$90\% \cite{battenberg2017accuracy}
& 22 \cite{Sportlin90:online} \\ \hline

Wrist-based
& Fitbit Force \cite{FitbitFo26:online}
& $<$90\% \cite{battenberg2017accuracy}
& Not available \\ \hline

Ankle-based
& StepWatch activity monitor \cite{CymaHome73:online}
& \parbox{5cm}{Non-running activities: $>$95\%  \\ Running activities: 74.4\% \cite{battenberg2017accuracy}}
& Not available \\ \hline

Mobile phone
& Apple iPhone 5 \cite{iPhone5T1:online}
& $<$90\% \cite{battenberg2017accuracy}
& Obsolete \\ \hline

Mobile phone
& Samsung Galaxy S4 \cite{SamsungG84:online}
& $<$90\% \cite{battenberg2017accuracy}
& 405 \cite{SamsungG50:online} \\ \hline

\end{longtable}
\end{scriptsize}

\item \textit{Glucose Level Monitoring:} Diabetes is characterized by elevated glucose levels.  For instance, an individual having a fasting plasma glucose level less than 100 mg/dl is non-diabetic, whereas one having a level between 100-125 mg/dl is considered pre-diabetic and having fasting plasma glucose level greater than 125 mg/dl is diabetic \cite{yu2010application}.  Table \ref{table:glucose} compares different invasive and non-invasive glucose monitoring devices. 

\begin{scriptsize}
\begin{longtable}{|p{3cm}|p{4cm}|p{5cm}|p{2cm}|}
\caption{Comparison Between Different Glucose Level Monitoring Devices.}
\label{table:glucose}\\
\hline
\rowcolor{Gray}
\textbf{Type} & \textbf{Device}& \textbf{Performance}& \textbf{Approximate cost (in US Dollars)}\\
\hline

Non-invasive
& Wearable-band type visible-near infrared optical \cite{rachim2019wearable}
& Average correlation coefficient between actual and measured glucose: 0.86 \cite{rachim2019wearable}
& Not available \\ \hline

Non-invasive
& Triple-pole complementary split ring resonator-based microwave bio-sensor \cite{omer2020non}
& Sensitivity: 6.2 dB/(mg/ml) \cite{omer2020non}
& Not available \\ \hline

Invasive
& EasyTouch \cite{glucomet5:online}
& Not reported
& 60 \cite{Amazonco2:online} \\ \hline

Invasive
& BeneCheck Plus \cite{httpswww35:online}
& Not reported
& 136 \cite{Benechec5:online} \\ \hline

\end{longtable}
\end{scriptsize}

\end{itemize}

\subsection{Artificial Intelligence Diabetes Prediction Application}
As shown in Figure \ref{fig:Architecture overview}, the mobile application plays the role of a gateway between the sensors connected to the users and the edge computing devices for uploading risk factors data.  A user can also communicate with the mobile application to identify the risk of incidence diabetes based on risk factors data.  In that case, the mobile application will communicate with the edge server to retrieve the prediction results.
 
\subsection{Edge and Cloud Computing}
\begin{figure}
\centering
  \includegraphics[width=\linewidth]{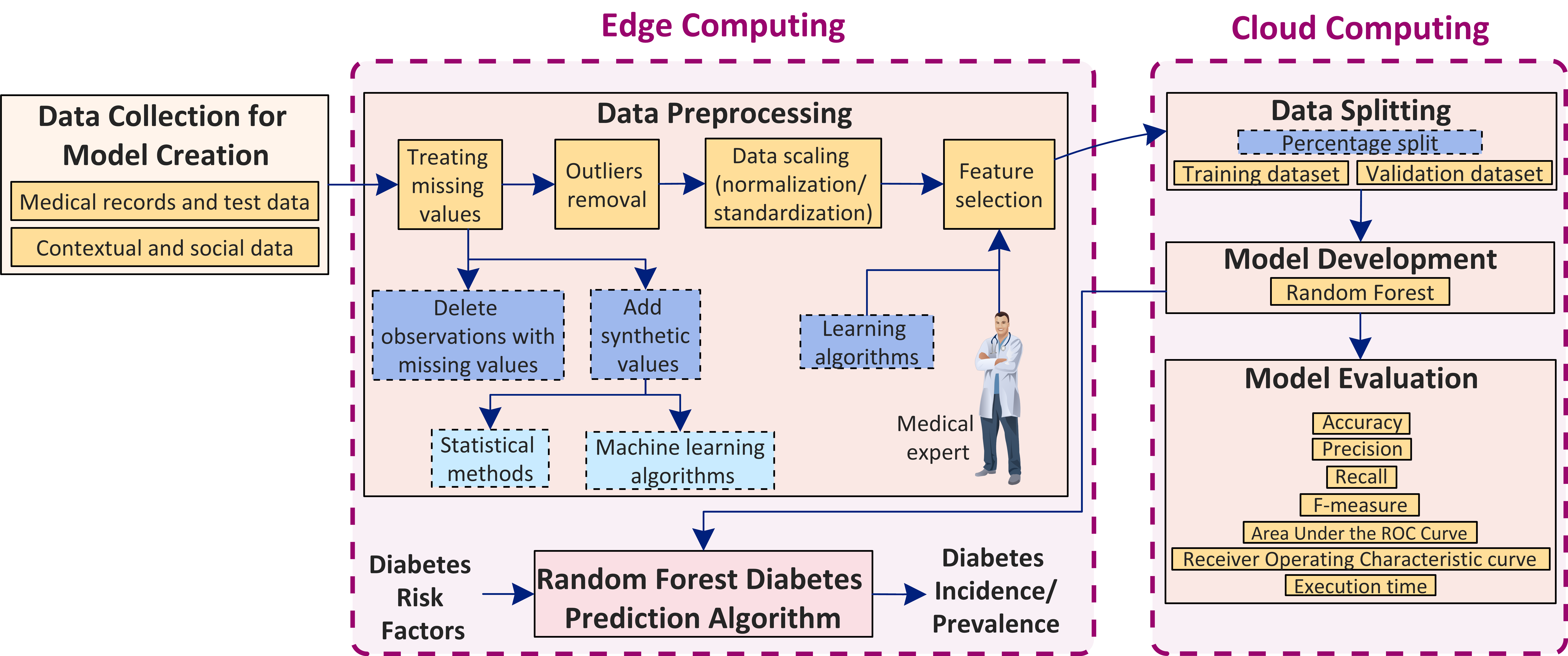}
  \caption{Stages of Artificial Intelligence-based Diabetes Mellitus Prediction System.}
  \label{fig:stages of AI framework}
\end{figure}

The risk factors data, collected using different medical sensors and devices, is sent to edge servers to transform it into a format that can be used by a machine learning algorithm.  Edge servers in close proximity to the users, compared to the cloud, aid in real-time data acquisition.  The collected data is then preprocessed by the edge servers.  However, edge servers are not capable of training compute-intensive machine learning models due to low processing and storage capacity.  Consequently, the preprocessed data is sent to the cloud for storage and machine learning model(s) development and validation.  The developed model is then sent back to the edge for predicting the risk of diabetes incident based on risk factors data.  Figure \ref{fig:stages of AI framework} shows the machine learning operations pipeline used in our framework for diabetes prediction.  The following explains the different operations involved in the machine learning pipeline for AI-based diabetes mellitus prediction.

\begin{itemize}
\item \textit{Data collection for model creation:} in this stage medical records, laboratory results, and contextual and social data are collected.  The inclusion of the risk factors in the dataset should be verified.  The collected data is then required to be aggregated.  The diabetes class labels should be defined.  For instance, all the observations in the dataset having fasting plasma glucose levels less than 100 mg/dl can be labeled as a non-diabetic class, whereas all having levels between 100-125 mg/dl can be labeled as a pre-diabetic class and all having fasting plasma glucose level greater than or equal to 125 mg/dl can be labeled as a diabetic class.  This can be done with the help of an expert's advice.
\item \textit{Data preprocessing:}  which involves handling missing values, removal of outliers, data scaling, and feature selection.  The missing values can be treated by either removing the corresponding observations or adding synthetic values.  Synthetic values can be generated using statistical (mean/mode/median) or machine learning (kNN imputation and rpart) approaches \cite{jerez2010missing}.  Data scaling is achieved through normalization and/or standardization.  The numerical features having varying ranges should be normalized.  This is because the model could be biased towards the feature with a bigger range \cite{larose2006data}.  For example. the range for BMI is 18.2-67.1, whereas that for plasma glucose is 44-199.  In feature selection, the features that do not contribute to diabetes are excluded to avoid overfitting the model at its development stage.  For instance, features, such as data sequence number, hospital ID, time, and date should be re-moved.  All the features (diabetes risk factors) available in the dataset can be used or a subset of features can be selected by applying feature selection algorithms  \cite{chandrashekar2014survey} or taking an expert's advice or using a hybrid approach. In our proposed system we use Recursive Feature Elimination \cite{guyon2002gene} which selects the set of features that are more relevant to the incidence of diabetes.  
\item \textit{Data splitting:} the data is split for training (model development) and testing.  This is done by dividing the dataset into 70\% and 30\% for training and validation respectively.
\item \textit{Model Development:} k-fold cross-validation technique \cite{fushiki2011estimation} is used to develop the model with the preprocessed training data. In the proposed system, we use decision-tree random forest (RF) classification model \cite{cutler2012random} as it is the top-used algorithm in the diabetes literature \cite{ahmad2021investigating, kopitar2020early, deberneh2021prediction, syed2020machine, joshi2021predicting, chang2022pima, zhang2020machine, lu2022patient}.
\item \textit{Model Evaluation:} The developed model is evaluated using validation data in terms of accuracy, precision, recall, F-measure, ROC, AUC, and execution time.  F-measure is an important metric to evaluate the performance of a machine learning model when trained using an imbalance dataset.  This is because F-measure can reveal the ability of the model to detect both majority and minority classes \cite{ismail2021idmpf}.
\item \textit{Diabetes Prediction:} The evaluated machine learning algorithm is used to predict the incidence or diagnose the prevalence of diabetes based on the risk factors data.
\end{itemize}

\subsection{Blockchain}
Security and privacy of healthcare data are the main requirements for a trustworthy and patient-centric system \cite{ismail2020requirements}.  The cloud provides scalable computing and storage facilities for healthcare data.  However, the involvement of a third-party cloud service providers leads to increased security and privacy threats due to a lack of transparency and data integrity.  Blockchain eliminates a centralized authority and ensures trust and transparency among the network participants.  The blockchain component in our proposed framework connects all the network participants in a peer-to-peer manner.  The network participants involve allied health professionals, patients, pharmacies, medical experts, and hospitals.  Each participant is authenticated by a certificate authority.  Table \ref{table:blockchain_security} shows how blockchain addresses different security and privacy issues that prevail in an only cloud-based system.

\begin{scriptsize}
\begin{longtable}{|p{3.5cm}|p{12cm}|}
\caption{Security and Privacy Analysis using Blockchain.}
\label{table:blockchain_security}\\
\hline
\rowcolor{Gray}
\textbf{Issue} & \textbf{Blockchain solution} \\
\hline

Data confidentiality
& The private and sensitive health data records can be only accessed by authorized network participants based on access control rights defined in the blockchain. A transaction for unauthorized access will not be validated by the network participants. \\ \hline

Data integrity
& Health data records are stored in blocks and each block is linked to the previous one using a cryptography mechanism. Modifying existing data in a block is computationally very expensive as the attacker has to change all the subsequent blocks in each copy of the ledger. Furthermore, any modification if performed will be logged in the ledger and can be easily traced.   \\ \hline

Data repudiation
& Data update and query events are recorded in an immutable ledger after validation ensuring fraud denials. \\ \hline

Data audit
& The replicated, time-stamped and immutable ledger ensures efficient, trusted, and integral auditing. \\ \hline

Data access control
& Access control rights for health data records in the blockchain can be defined using smart contracts for secure access by authorized participants. \\ \hline

\end{longtable}
\end{scriptsize}

We use non-encapsulated integrated blockchain-cloud architecture \cite{ismail2021scoping}, in which the diabetes risk factors data are stored in the cloud database and the associated meta-data is recorded in the blockchain, such as the hash of the risk factors data, update and query events, access control policy, and diabetes prediction results.  Storing data in the cloud aids in system scalability, whereas recording meta-data in the blockchain ledger enables security and privacy.  The hash of risk factors data and prediction results in the ledger ensures data integrity. In addition, recording data update and query events in the ledger discourages unauthorized access, leading to enhanced privacy.  Furthermore, we employ multi-ledger-based permissioned blockchain architecture that provides configurable access control rights and facilitates the development of a separate ledger for collaborating allied health professionals \cite{ismail2019review}.  The selection of permissioned blockchain over permissionless \cite{zheng2018blockchain} is due to the following disadvantages of the latter: 1) unauthorized participation in the network leading to impersonate account holders, 2) clear transaction data in the ledger accessible to each network participant revealing sensitive patients' data, 3) slow network throughput hindering real-time patient's treatment, and 4) the need of paying transaction execution fees and mining rewards limiting the usability of the network.

The blockchain component consists of participants, assets, transactions, and events.  Table \ref{table:blockchain network} shows the different types of participants, assets, transactions, and events that will be used in our system along with their descriptions.

\begin{scriptsize}
\begin{longtable}{|p{2.5cm}|p{3.5cm}|p{9cm}|}
\caption{Description of Participants, Assets, Transactions, and Events for the Proposed Blockchain network.}
\label{table:blockchain network}\\
\hline
\rowcolor{Gray}
\cellcolor[HTML]{FFFFFF} &\textbf{Name} & \textbf{Description}\\
\hline

Participants
& Hospitals
& Responsible for uploading medical records to the cloud, validating healthcare transactions, and responding to the data retrieval query.  They store a copy of the ledger. \\ \hline

Participants
& Allied health professionals
& They are the doctors and nurses registered with the hospitals.  They are responsible for updating patients' medical records based on symptoms, diagnoses, treatments, and medications.  They can also update the laboratory and pathological results.  In addition, they can query the medical records from the cloud by performing query transactions. \\ \hline

Participants
& Pharmacists
& They are responsible for updating the information related to medications, bills, and insurance claims to the cloud.  This is by performing update transactions.  In addition, they can query a patient's records. \\ \hline

Participants
& Patients
& They are the diabetic and pre-diabetic patients registered with the hospitals.  They can query their medical data and update contextual and life-style data to the cloud by performing update transactions.  The patients can enter the data into the network using mobile phones. \\ \hline

Participants
& External users
& They are the participants not necessarily registered with the hospitals.  They can insert their lifestyle, medical conditions, hereditary, psychosocial, and demographic data to predict the development of diabetes. \\ \hline

Assets
& Laboratory and pathological data (by hospitals)
& This asset includes laboratory and pathological test data such as blood and urine reports, x-rays, MRIs, ultrasound, endoscopy, fasting plasma glucose, uric acid level, etc.  These data are updated by the hospitals to the cloud with the hash of the data being recorded in the blockchain.  The data is made available to the corresponding patient upon a data retrieval query. \\ \hline

Assets
& Medical condition data (by hospitals)
& This asset includes the medical condition data such as symptoms, diagnosis, medications, treatments, and vitals, i.e., heart rate, blood pressure, oxygen level, cholesterol level, and BMI.  These data are sent to the cloud for storage with meta-data recorded in the blockchain. \\ \hline

Assets
& Social and contextual data (by patients)
& This asset includes the social and contextual data such as age, gender, family history of diabetes, history of heart disease, depression, ethnicity, geographical location, smoking habits, alcohol consumption, diet, sleep duration, physical activity, educational level, and socioeconomic status.  These data are sent as transactions by the patients for ledger updates. \\ \hline

Assets
& Risk factors data (by external users)
& This asset includes the diabetes risk factors data such as lifestyle, medical condition, hereditary, psychosocial, and demographic.  These data are sent by external users as transactions for the prediction of diabetes incidence. \\ \hline

Transactions
& Medical records update (by hospitals)
& This transaction involves the update of the patient's medical records by the hospitals to the cloud.  The meta-data is recorded in the ledger. \\ \hline

Transactions
& Laboratory and pathological results update (by hospitals)
& This transaction involves the update of the patient's laboratory and pathological results by the hospitals to the cloud.  The meta-data is recorded in the ledger. \\ \hline

Transactions
& Social and contextual data update (by patients)
& This transaction involves the update of the social and contextual data by the patients to the cloud.  The meta-data is recorded in the ledger. \\ \hline

Transactions
& Query (from patients to hospitals)
& This transaction involves the data retrieval request by the patient to the registered hospital for his/her medical data. \\ \hline

Transactions
& Response to query (from hospitals to the patients
& This transaction involves the response from the hospital to the data retrieval query made by the patient. \\ \hline

Transactions
& Risk factors data (from external users to AI-based prediction system)
& This transaction involves the risk factors data sent by the external user as transactions for the prediction of diabetes incidence.  The prediction request to the AI-based system will be recorded as a transaction in the ledger. \\ \hline

Transactions
& Risk of diabetes incidence (from AI-based prediction system to the external users)
& This transaction involves the prediction result regarding the development of diabetes.  This data is used by the hospitals to develop a prevention plan. \\ \hline

Events
& Patient's medical records update (to patients)
& The patient is notified about his/her records being added to the cloud by the corresponding hospital.  This notification helps the patient to be up-to-date with his/her records. \\ \hline

Events
& Patient's laboratory and pathological results up-date (to patients)
& The patient is notified about his/her laboratory and pathological results being added to the cloud by the corresponding hospital.  This notification helps the patient to be up-to-date with his/her results. \\ \hline

Events
& Patient's social and con-textual data update (to hospitals)
& The hospital receives the social and contextual data transaction from the patient requesting to be added to the cloud.  Upon validation, the data is added to the ledger. \\ \hline

Events
& External user's risk factors data update (to hospitals)
& The hospital receives the risk factors data transactions by the external to update the ledger. \\ \hline

Events
& External user's prediction update (to hospitals)
& The hospital updates the prediction results of the AI-based prognosis/diagnosis system in the blockchain ledger.  This will aid in the development of a nationwide prevention plan. \\ \hline

\end{longtable}
\end{scriptsize}

\begin{figure}
\centering
  \includegraphics[width=0.8\linewidth]{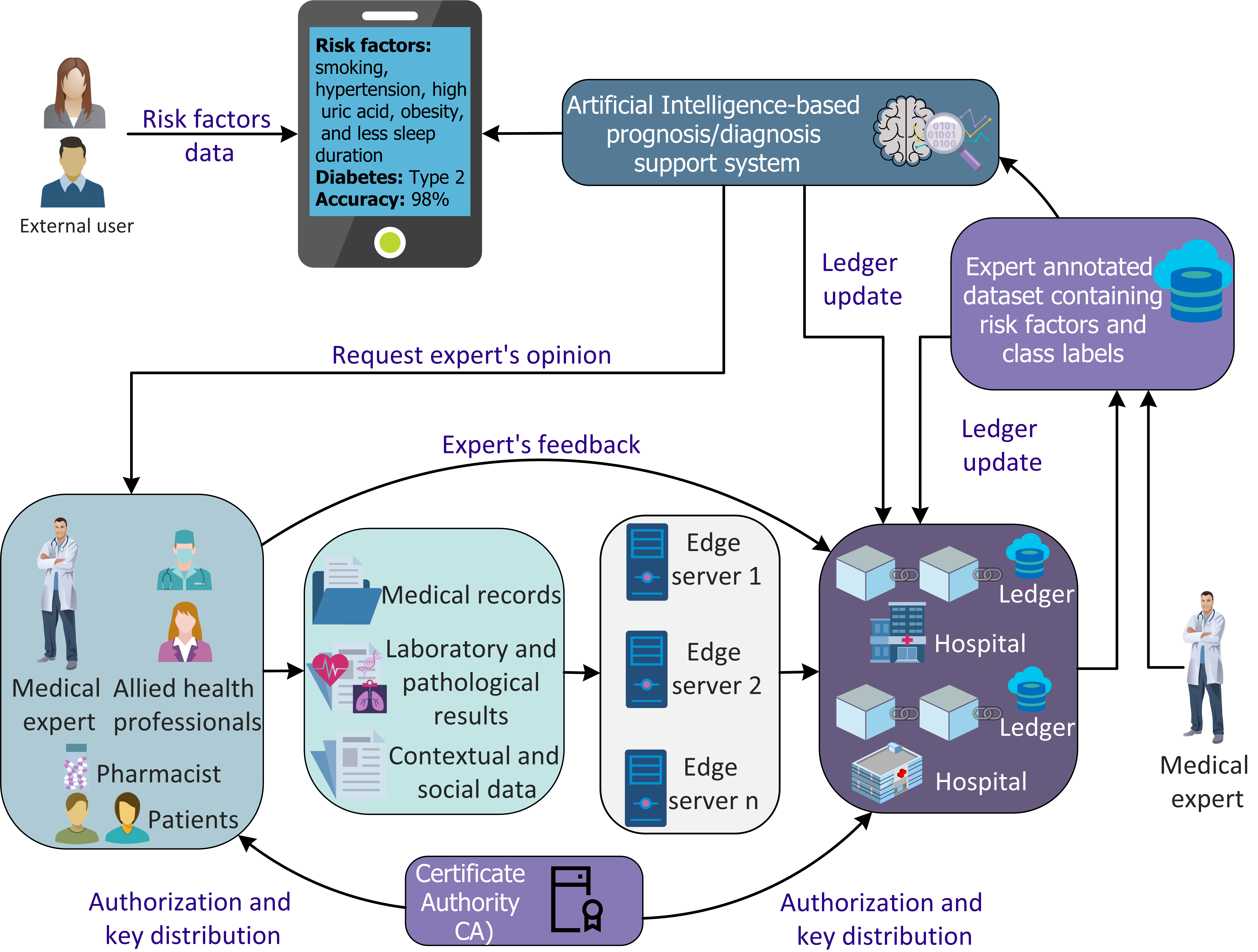}
  \caption{Proposed Blockchain and Artificial Intelligence Integrated Monitoring System for Prediction of Diabetes Mellitus.}
  \label{fig:proposed system}
\end{figure}

Figure \ref{fig:proposed system} shows the blockchain usage in our end-to-end AI-based prognosis/diagnosis support system for healthcare management.  In addition to the network participants described in Table \ref{table:blockchain network}, the system consists of a certificate authority (CA) and a medical expert.  The CA works as both a system administrator by removing malicious nodes from the network and an authority management entity by generating and distributing digital certificates.  A participant's public-private key pair is also generated by the CA.  The public-private key pair for each participant is linked to the participant ID, a secret PIN code set by the participant, and the participant identity proof.  In a situation where the participant loses his/her public-private key pair, a new pair is generated by the CA after authenticating the participant ID, secret PIN code, and identity proof.  Each network participant, i.e., patient, allied health professionals, pharmacists, and external users, is identified using an identity number.  For instance, a patient is identified by the patient ID whereas a doctor is identified by the doctor ID.  A medical expert is responsible for annotating the diabetes risk factors and class labels to the medical records data present in the cloud.  The hash of annotated data is stored in the blockchain ledger.  In addition, the medical expert will give feedback on the performance of the AI system when asked for an opinion.  The expert's feedback is recorded as a transaction in the cloud with its hash in blockchain.  The AI-based prognosis/diagnosis support system consists of classification learning models for diabetes prediction.  The prediction query and results are stored as transactions in the blockchain.  Table \ref{table:data and attributes} shows the data and the corresponding attributes used in the proposed system.

\begin{scriptsize}
\begin{longtable}{|p{3.5cm}|p{12cm}|}
\caption{Health Data and Corresponding Attributes used in the Proposed System.}
\label{table:data and attributes}\\
\hline
\rowcolor{Gray}
\textbf{Data} & \textbf{Attributes}\\
\hline

Laboratory and pathological results
& X-rays, MRIs, CT scans, blood report, and urine report \\ \hline

Medical records
& File number, patient ID, patient name, age, gender, nationality, national identity number, medical insurance number, contact details, patient name, height, weight, waist circumference, body temperature, blood pressure, the reason for attendance, patient medical history, family medical history, allergies, symptoms, diagnosis, point of care testing (random blood sugar, urine dip, pregnancy test), medications \\ \hline

Social and contextual data
& Age, diet, sleeping pattern, heart rate, physical activity, smoking habits, alcohol consumption \\ \hline

Risk factors data
& High-level serum uric acid, sleep quality/quantity, smoking, depression, cardiovascular disease, dyslipidemia, hypertension, aging, ethnicity, family history of diabetes, physical inactivity, and obesity \\ \hline

\end{longtable}
\end{scriptsize}

\section{Implementation of Proposed Automated End-to-End Blockchain Artificial Intelligence-System for Diabetes Mellitus Prediction}

In this section, the implementation of the system is discussed. The system operates through two main functions: 1) \textit{DP(user\_risk)}  which allows end-users to get diabetes prediction from the system through a front-end device (e.g. smart phone), and 2) \textit{DPMT($df_{risk}$)}, the diabetes prediction model trainer, which trains or updates the system's AI model by using new labeled data.

\begin{figure}
\centering
  \includegraphics[width=\linewidth]{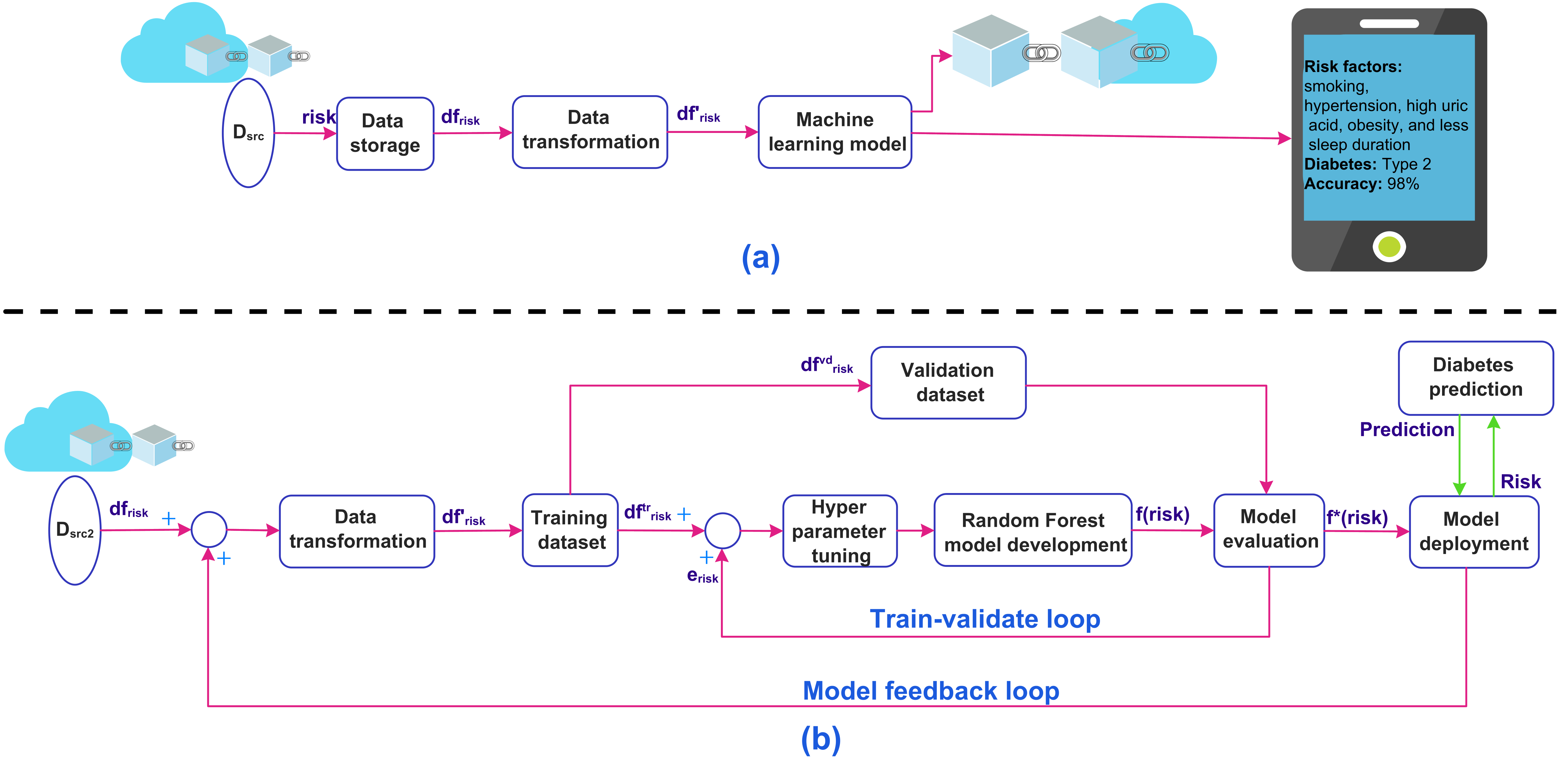}
  \caption{Implementation of the Proposed End-to-End Automated Artificial Intelligence (AI)-Blockchain Systems for Diabetes Monitoring.}
  \label{fig:implementation}
\end{figure}

For the first operative function, \textit{DP(user\_risk)}, the implementation diagram is shown in Figure \ref{fig:implementation}(a).  The risk factor data (which is unlabelled) is collected from a data source \textit{$D_{src}$}. The users' health records including diabetes risk factors data are stored in the cloud with meta-data recorded in the blockchain.  The raw risk factor data \textit{$df_{risk}$} is fed as an input to the data transformation component for preprocessing.  Data transformation is performed by edge servers.  The preprocessed data frame \textit{$df^{'}_{risk}$} is then passed as input to the current Machine Learning model for diabetes prediction. The result of the prediction is sent back to the end-user device and the Blockchain ledger.

For the second operative function, \textit{DPMT($df_{risk}$)}, the system is upgraded using previous modeling data and the new data generated by users and/or health professionals, which is already labeled by the health professionals and stored in the cloud, as shown in Figure \ref{fig:implementation}(b). This new training data, \textit{$D_{src2}$}, is extracted from the cloud data source to be fed as input to the data transformation component for preprocessing.  The data extraction and preprocessing operations along with the meta-data are recorded in the blockchain ledger.  The preprocessed data is then divided into training \textit{$df^{tr}_{risk}$} and validation \textit{$df^{vd}_{risk}$} datasets.  The selected Random Forest model \textit{f(risk)} is trained again using \textit{$df^{tr}_{risk}$}.  The performance of the model is evaluated using \textit{$df^{vd}_{risk}$}.  The model development is a feedback control process where the model is tuned using hyperparameter tuning unless the desired performance is obtained. The diabetes prediction error \textit{$e_{risk}$} obtained from the evaluation of the prediction model is fed back to tune the hyperparameters.  The tuned model \textit{$f^{*}(risk)$} is deployed in the system for predicting accurately the risk of diabetes occurrence in users.  Consequently, the diabetes prediction function, \textit{DP(user\_risk)}, uses the deployed model to predict the risk of diabetes. 

\section{Method}

In this section, we present the methodology used to evaluate our proposed system.  The proposed system consists of four components: 1) data collection and storage, 2) data preprocessing, 3) machine learning model development and validation, and 4) machine learning model deployment.  In this paper, we use three public diabetes datasets: PIMA Indian \cite{smith1988using}, Sylhet \cite{islam2020likelihood}, and MIMIC III \cite{johnson2016mimic}, to evaluate the data preprocessing and machine learning model development and validation components.  The datasets are selected in a way that they include as many diabetes risk factors as possible, and they are not integrated as they have different sets of features for diabetes prediction.  Each dataset contains two class labels, i.e., diabetic or non-diabetic.  The datasets are then explored to identify the correlations between the risk factors and the occurrence of diabetes in patients and users.  Datasets are preprocessed using cleaning and normalization, by removing the observations with missing values, and normalizing values across the observations, respectively.  We evaluate the performance using the most used machine learning models in the literature on diabetes prediction, namely Random Forest (RF), Logistic Regression (LR) \cite{hosmer2013applied}, and Support Vector Machine (SVM) \cite{hearst1998support}, with and without feature selection, with and without balancing.  Recursive Feature Elimination, Cross-Validated (RFECV) feature selection algorithm, and Synthetic Minority Oversampling Technique (SMOTE) balancing are used because they showed efficient performance \cite{chawla2002smote}. RFECV \cite{guyon2002gene}  selects the best subset of features by removing some features and selecting the best subset based on a cross-validation score.  SMOTE \cite{chawla2002smote} adds synthetic data points by selecting random samples of the minority class and choosing a point in between these points and one of their k-nearest neighbors. We evaluated those models using different evaluation metrics such as Accuracy, F-measure, precision, recall, and AUC.

\subsection{Datasets}

Table \ref{table:datasets characteristics} shows the characteristics of the datasets used to evaluate our proposed system respectively:  1) PIMA India \cite{smith1988using} from the National Institute of Diabetes and Digestive and Kidney Diseases,  2) Sylhet \cite{islam2020likelihood} was collected using direct questionnaires from the patients of Sylhet Diabetes Hospital in Sylhet, Bangladesh, and 3) MIMIC III \cite{johnson2016mimic}, a large dataset which contains information of over 40,000 patients who stayed in critical care units of the Berth Israel Deaconess Medical Center between 2001 and 2012.

\begin{scriptsize}
\begin{longtable}{|p{2cm}|p{4.5cm}|p{2.5cm}|p{2.5cm}|p{2.5cm}|}
\caption{Original Datasets Characteristics.}
\label{table:datasets characteristics}\\
\hline
\rowcolor{Gray}
\textbf{Dataset} & \textbf{Features} & \textbf{Positive Classes} & \textbf{Negative Classes} & \textbf{Total Records}\\
\hline

PIMA Indian
& Pregnancies, Glucose, Blood Pressure, Skin Thickness, Insulin, BMI, Diabetes pedigree$^{1}$, and Age	
& 268 (34.9\%)
& 500 (65.1\%)
& 768 \\ \hline

Sylhet
& Age, Gender, Polyuria$^{2}$, Polydipsia$^{3}$, sudden weight loss, weakness, Polyphagia$^{4}$, Genital thrush$^{5}$, visual blurring, Itching, Irritability, delayed healing, partial paresis$^{6}$, muscle stiffness, Alopecia$^{7}$, and Obesity
& 320 (61.5\%)
& 200 (38.5\%)
& 520 \\ \hline

MIMIC III
& Ethnicity, Gender, Age, and Family History of Diabetes
& N/A
& N/A
& 46,520 \\ \hline	 		

\multicolumn{5}{p{14cm}}{$^{1}$Diabetes pedigree provides a synthesis of diabetes history in relatives and the genetic relationship of those relatives to the subject \cite{smith1988using}} \\

\multicolumn{5}{p{14cm}}{$^{2}$Polyuria is a condition where the body urinates more than usual and passes excessive or abnormally large amounts of urine each time you urinate \cite{islam2020likelihood}} \\

\multicolumn{5}{p{14cm}}{$^{3}$Polydipsia is the feeling of extreme thirstiness \cite{islam2020likelihood}} \\

\multicolumn{5}{p{14cm}}{$^{4}$Polyphagia, also known as hyperphagia, is the medical term for excessive or extreme hunger \cite{islam2020likelihood}} \\

\multicolumn{5}{p{14cm}}{$^{5}$Genital thrush is a common infection caused by an overgrowth of yeast \cite{islam2020likelihood}} \\

\multicolumn{5}{p{14cm}}{$^{6}$Paresis involves the weakening of a muscle or group of muscles. It may also be referred to as partial or mild paralysis. Unlike paralysis, people with paresis can still move their muscles. These movements are just weaker than normal \cite{islam2020likelihood}.} \\

\multicolumn{5}{p{14cm}}{$^{7}$Alopecia areata is an autoimmune disorder that causes your hair to come out, often in clumps the size and shape of a quarter \cite{islam2020likelihood}}

\end{longtable}
\end{scriptsize}

\subsection{Data Exploration}

\begin{figure}
\centering
  \includegraphics[width=\linewidth]{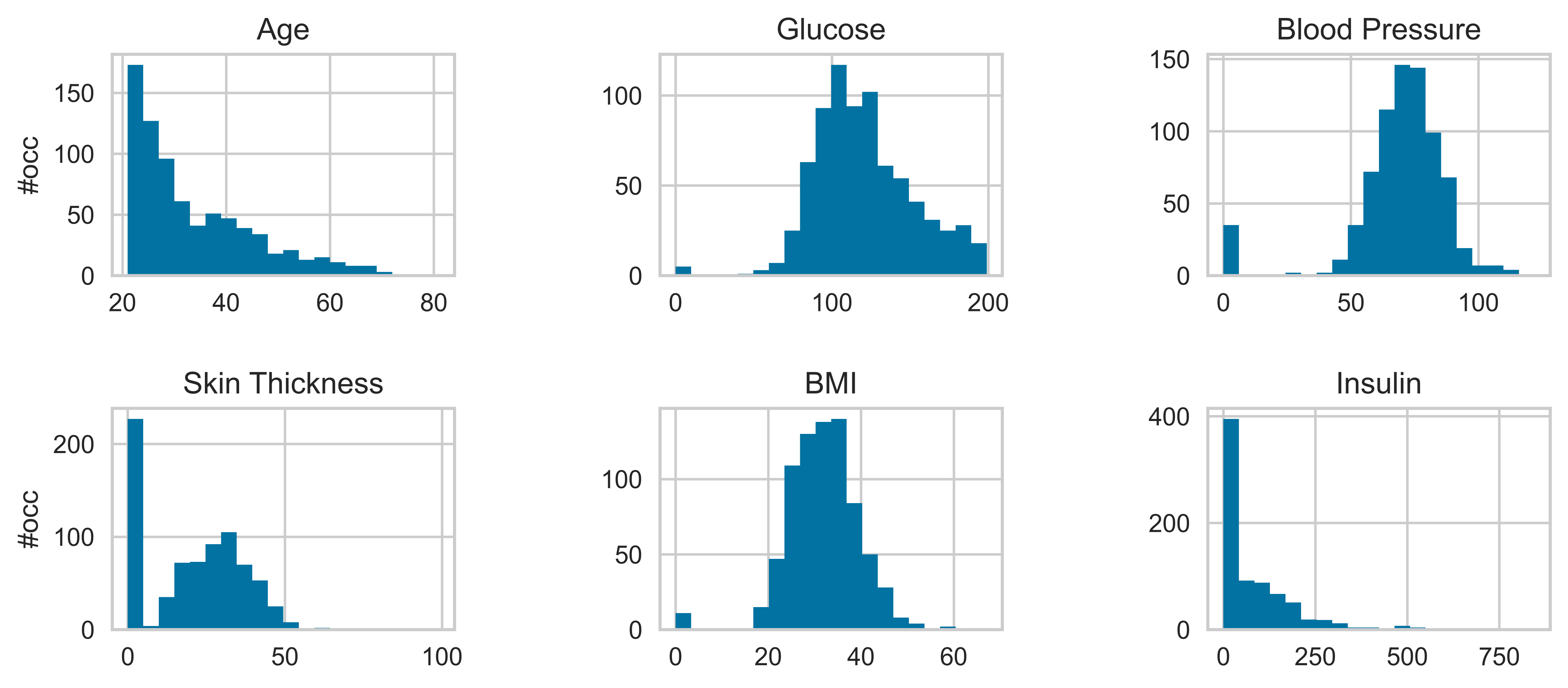}
  \caption{Data Exploration Histograms for PIMA Indian Dataset Numerical Features.}
  \label{fig:data exploration PIMA indian}
\end{figure}

PIMA Indian dataset shows a number of missing values in some numerical features. In particular, Blood Pressure, Skin Thickness, and BMI are characterized by a heavy weight for the \textit{'zero'} value, on, shown in Figure \ref{fig:data exploration PIMA indian}.  This implies that the corresponding observations should be removed at the preprocessing stage.

\begin{figure}
\centering
\begin{subfigure}{0.5\textwidth}
    \includegraphics[width=\textwidth]{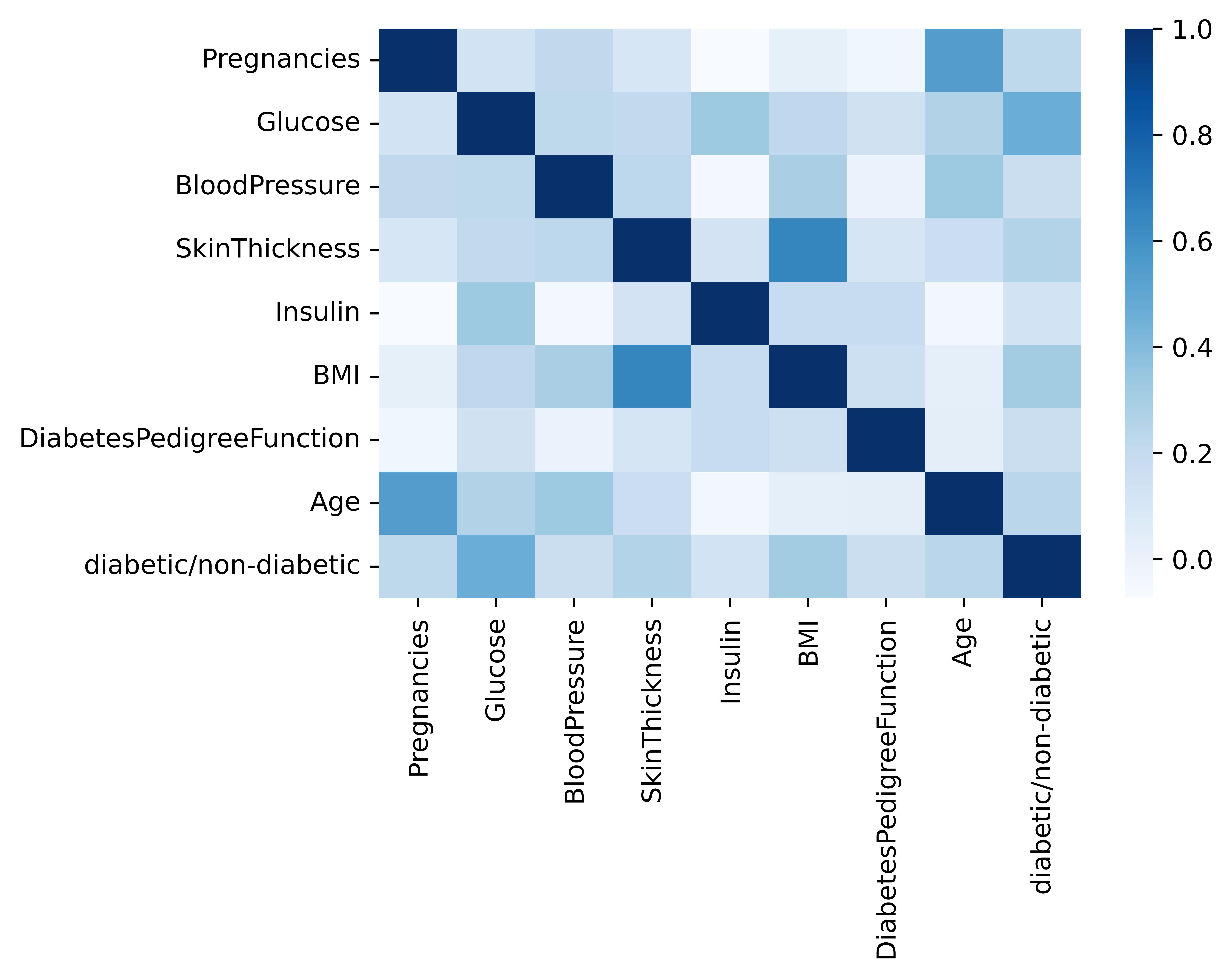}
    \caption{PIMA Indian}
    \label{fig:correlation PIMA indian}
\end{subfigure}
\hfill
\begin{subfigure}{0.5\textwidth}
    \includegraphics[width=\textwidth]{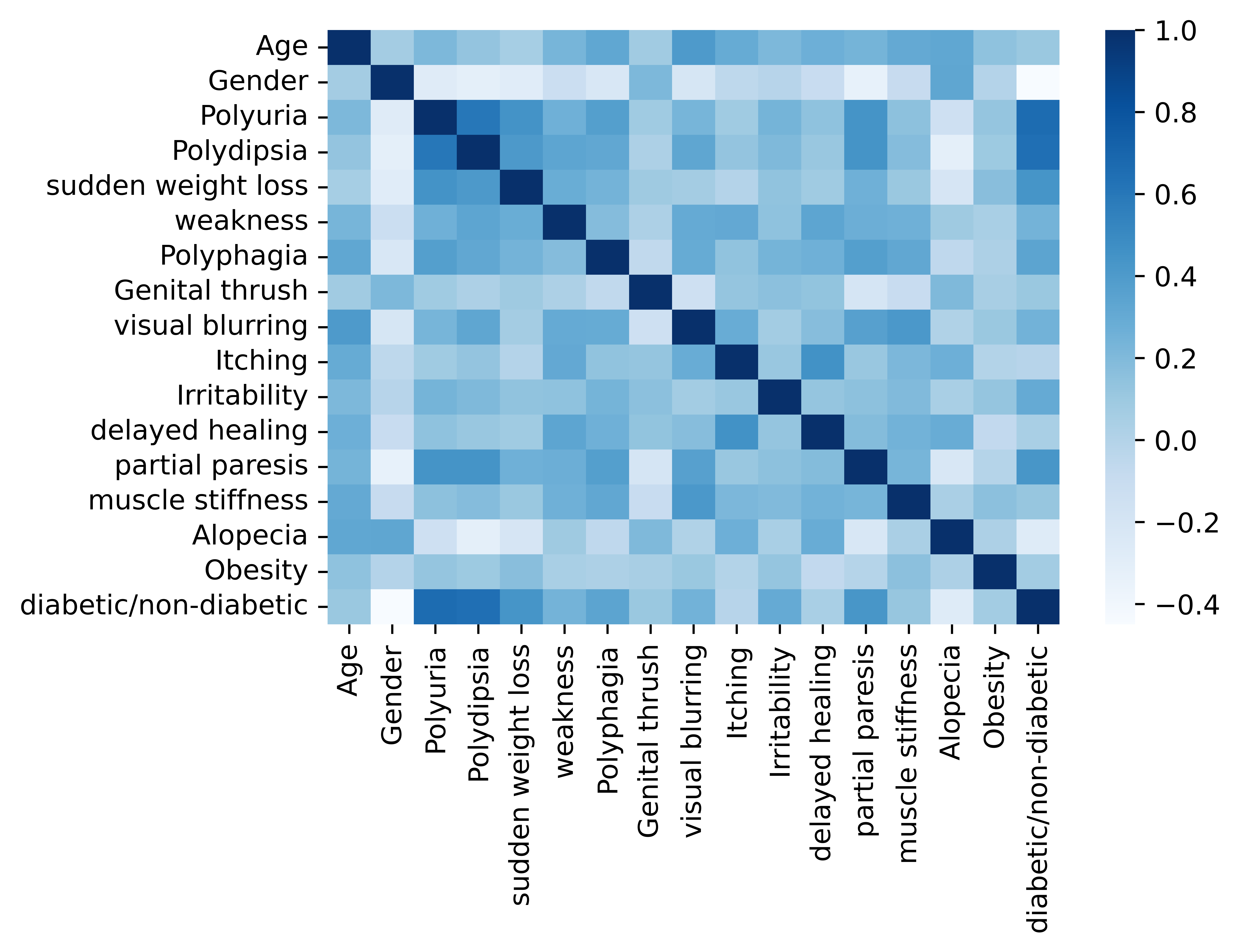}
    \caption{Sylhet}
    \label{fig:correlation SYLHET}
\end{subfigure}
\hfill
\begin{subfigure}{0.5\textwidth}
    \includegraphics[width=\textwidth]{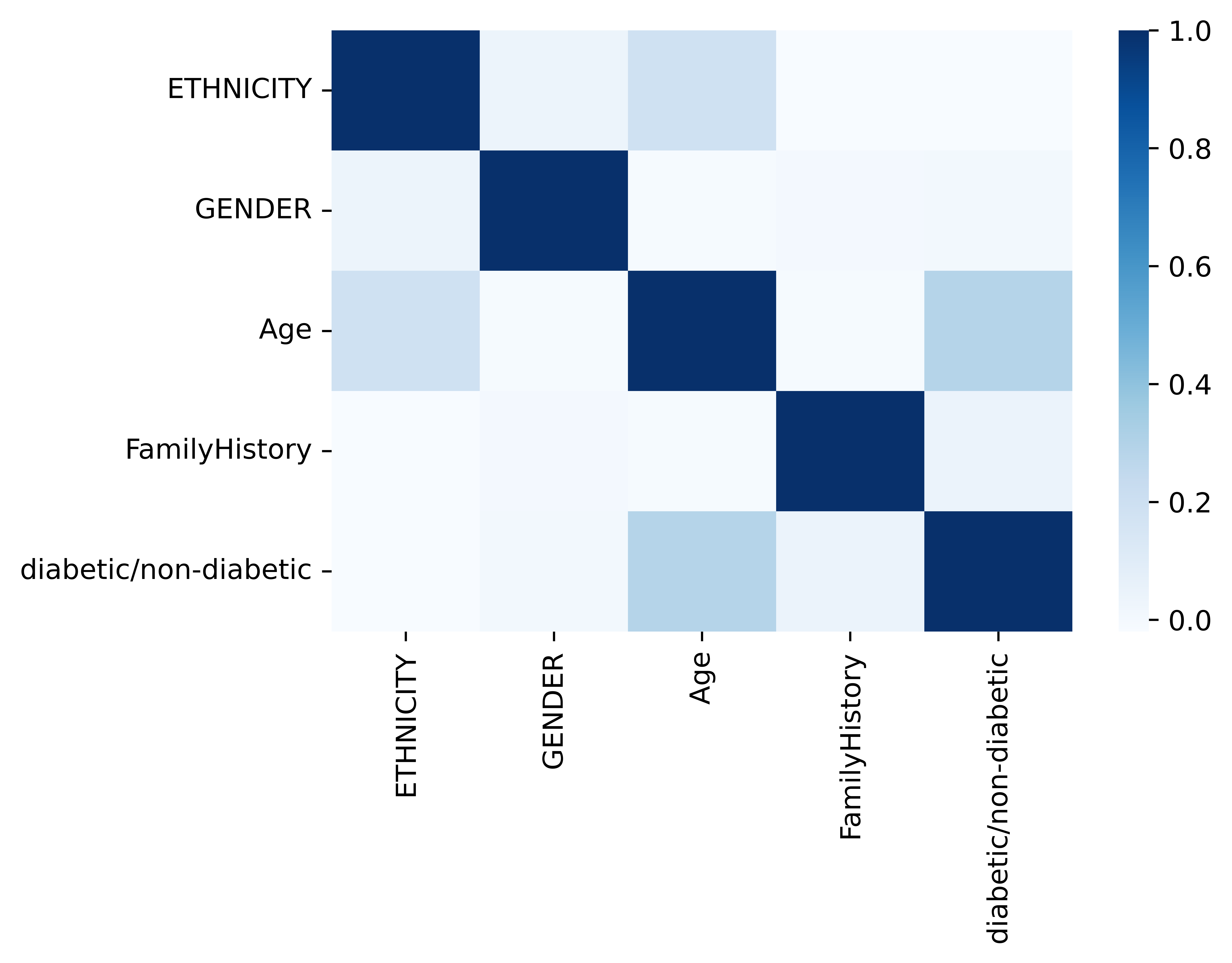}
    \caption{MIMIC III}
    \label{fig:correlation MIMIC III}
\end{subfigure}
\caption{Correlation between Features and Diabetic/Non-diabetic Class for PIMA Indian, Sylhet, and MIMIC III Datasets}
\end{figure}

For each dataset, we study the correlations among features and the diabetic/non-diabetic class. We choose the Phik ($\Phi$k) correlation coefficient because that works consistently between categorical, ordinal, and interval variables. It captures non-linear dependency and reverts to the Pearson correlation coefficient in the case of bi-variate normal input distribution \cite{baak2020new}. So, it encompasses multiple types of correlations.  As shown in Figure \ref{fig:correlation PIMA indian}, PIMA India presents a logical correlation between Age and Number of Pregnancies.  Regarding diabetes detection, the features that are correlated with the diabetic/non-diabetic outcome of the patient are Glucose, Age, BMI, Insulin, and Skin Thickness.  In addition, BMI is correlated with Blood Pressure. Correlations for Sylhet are displayed in Figure \ref{fig:correlation SYLHET}.  In this dataset, the diabetic/non-diabetic outcome is highly correlated with Polydipsia and Polyuria and in a lower manner with partial paresis, Gender, and sudden weight loss. Furthermore, Polydipsia and Polyuria are highly correlated with each other.  Similar to PIMA India, MIMIC III dataset shows high correlations between the class (diabetic/non-diabetic) outcome and the Age feature (Figure \ref{fig:correlation MIMIC III}). However, Figure \ref{fig:correlation MIMIC III} shows a correlation between Age and Ethnicity that may indicate the randomness in the MIMIC III dataset under study.

\subsection{Data Preprocessing}

In the PIMA Indian dataset, we remove observations with missing data for Skin Thickness, BMI, and Blood Pressure. The Sylhet dataset does not have any missing values. Regarding the MIMIC III dataset, there is a need to extract the available risk factor feature from the raw data. MIMIC III raw data are split into different tables. The data of interest in MIMIC III for bringing out risk factors and diabetic/non-diabetic class is shown in Table \ref{table:MIMIC}.

\begin{scriptsize}
\begin{longtable}{|p{4cm}|p{8cm}|}
\caption{Data Tables used from MIMIC III Dataset.}
\label{table:MIMIC}\\
\hline
\rowcolor{Gray}
\textbf{Table Name} & \textbf{Available data and purpose} \\
\hline

PATIENTS
& Subject ID, Gender, Date of Birth \\ \hline

ADMISSIONS
& Subject ID, visits of a patient, start and end of the patient visit, other demographic data (Ethnicity) \\ \hline

DIAGNOSES\_ICD
& Subject ID, Association of ICD9 diagnostics with patients \\ \hline

D\_ICD\_DIAGNOSES
& Dictionary of ICD9 codes associated with their description \\ \hline

\end{longtable}
\end{scriptsize}

We build the MIMIC III machine learning dataset by joining information of the patients from the different data tables (Table \ref{table:MIMIC}).  For each patient, we have Age, and Ethnicity from PATIENTS and ADMISSIONS tables.  The information about diabetic/non-diabetic outcomes is retrieved from the ICD9 diagnostics associated with the patient in table DIAGNOSES\_ICD.  If one of the diagnostics is for diabetes mellitus, then the patient is set to have diabetes.  In the same manner, we create a feature 'Family History of Diabetes' by querying if a patient has ICD9 diagnostic code V180 (Family history of diabetes mellitus). The categorical values of 'UNKNOWN/NOT SPECIFIED', 'PATIENT DECLINED TO ANSWER', and 'UNABLE TO OBTAIN' for Ethnicity in the MIMIC III dataset are interpreted as missing values.  Consequently, patients with such values for Ethnicity are removed from the dataset. Table \ref{table:Dataset characteristics after preprocessing} shows the characteristics of the resulting datasets after preprocessing.

\begin{scriptsize}
\begin{longtable}{|p{2.5cm}|p{2.5cm}|p{2.5cm}|p{2.5cm}|p{2.5cm}|}
\caption{Dataset Characteristics after Preprocessing.}
\label{table:Dataset characteristics after preprocessing}\\
\hline
\rowcolor{Gray}
\textbf{Dataset} & \textbf{\# of Features} & \textbf{Positive Classes} & \textbf{Negative Classes} & \textbf{Total Records} \\
\hline

PIMA Indian & 8 & 177 (33.3\%) & 355 (66.7\%) & 532 \\ \hline

Sylhet & 16 & 320 (61.5\%) & 200 (38.5\%) & 520 \\ \hline

MIMIC III & 4 & 8,820 (22.5\%) & 30,469 (77.5\%) & 39,289 \\ \hline

\end{longtable}
\end{scriptsize}

\subsection{Feature Selection}

We use RFECV \cite{guyon2002gene} with Random Forest \cite{cutler2012random} as a cross-validation evaluator. The Random Forest model is used to detect feature importance in learning. Random Forest is a kind of Bagging Algorithm that aggregates a specified number of decision trees. The tree-based random forest ranks the features according to how well the purity of the feature is improved, that is, a decrease in the impurity (Gini impurity) over all the trees. Features with the greatest decrease in impurity happen at the start of the trees, while features with the least decrease in impurity occur at the end of trees. Therefore, by pruning trees below a particular feature, one can create a subset of the most important features. Recursive Feature Elimination works by searching for a subset of features by starting with all features in the training dataset and successfully removing features until the desired number remains. This is achieved by fitting random forest, ranking features by importance, discarding the least important features, and re-fitting the model. This process is repeated until a specified number of features remains.

\subsection{Balancing Data Augmentation}

The three datasets that we are using are slightly imbalanced towards the negative class, MIMIC III is more imbalanced. The number of diabetes class observations is roughly 30\% for PIMA Indian and Sylhet datasets and 22\% for MIMIC. To reduce the biases in the created models, the synthetic minority oversampling technique (SMOTE) \cite{chawla2002smote} is used as a data balancing technique. SMOTE is an oversampling technique that increases the number of minority class samples in the dataset, by generating new samples from existing minority class samples. The application of SMOTE to clinical datasets can improve model performance by reducing the negative effects of imbalanced data as observed in recent literature. SMOTE is only applied on the training/validation split (70\%) of the data samples so that the model sees equal numbers of both class types, and the test split (30\%) is not modified.

\subsection{Machine Learning Models for Diabetes Prediction}

\begin{figure}
\centering
  \includegraphics[width=\linewidth]{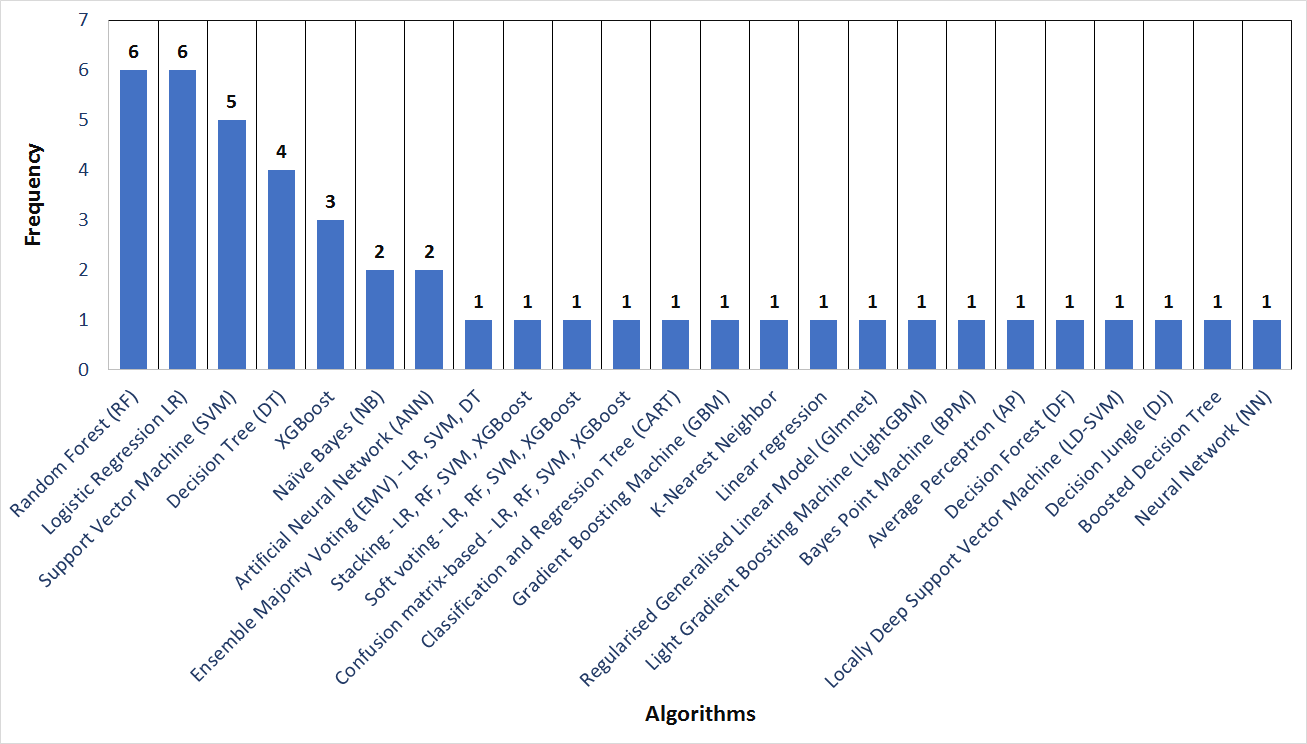}
  \caption{Frequency of Classification Algorithms used in Literature for Diabetes Prediction.}
  \label{fig:frequency}
\end{figure}

We implement our proposed automated end-to-end blockchain AI-system for diabetes prediction using the most popular and accurate Random Forest (RF) model. Figure \ref{fig:frequency} shows the relative usage frequency of different Machine Learning algorithms in current research papers on Diabetes prediction \cite{ahmad2021investigating, kopitar2020early, deberneh2021prediction, syed2020machine, joshi2021predicting, chang2022pima, zhang2020machine, lu2022patient}.  To evaluate our proposed system, we implemented Logistic Regression (LR) and Support Vector Machine (SVM) algorithms for diabetes prediction and compare their performances with RF.  The selection of LR and SVM is based on their popularity as shown in Figure \ref{fig:frequency}.

\subsubsection{Random Forest (RF)}

This algorithm is based on Decision Tree (DT), which constructs a tree structure to define the sequences of decisions and outcomes, and to use it for prediction. At each node of the tree, the algorithm selects the branch having the maximum information gain.

Random Forest is a set of decision trees constructed using randomly selected samples of the dataset \cite{cutler2012random}. It performs voting on the output of each decision tree and classifies an observation into diabetes or non-diabetes depending on the majority of the decision trees' output.

\subsubsection{Logistic Regression (LR)}

This algorithm predicts the probability that a given observation belongs to the diabetes or non-diabetes class using a sigmoid function \cite{hosmer2013applied} as stated in Equation \ref{eq:LR1}.

\begin{equation}
\label{eq:LR1}
P(diabetes) = \frac{e^{\beta_0}+\sum_{i=1}^{n}\beta_{i}R_{i}}{1+e^{\beta_0+\sum_{i=1}^{n}\beta_{i}R_{i}}}
\end{equation}

where p(diabetes) represents the probability of having diabetes, R is the set of risk factors, and $\beta_0$ and $\beta_i$ are the regression coefficients representing the intercept and the slope respectively. The values of regression coefficients are calculated using maximum likelihood estimation such that the value of Equation \ref{eq:LR2} is the maximum.

\begin{equation}
\label{eq:LR2}
l(\beta_0, ..., \beta_1) = \prod_{i,y_i=1}P(diabetes)\prod_{i,y_i=0}(1-P(diabetes))
\end{equation}

\subsubsection{Support Vector Machine (SVM)}

This algorithm aims to create a decision boundary known as a hyperplane that can separate n-dimensional instance space into diabetes and non-diabetes classes. The hyperplane is created using the extreme points (support vectors) of the dataset. The generation of a hyperplane is an iterative process to find the maximum possible margin between the support vectors of the opposite classes.  Let $r^{(i)}$ and $y^{(i)}$ represent the risk factors and classes in the dataset and there exists a hyperplane that separates diabetes and non-diabetes classes as stated in Equation \ref{eq:SVM1}.

\begin{equation}
\begin{aligned}
\label{eq:SVM1}
w^Tr + b = 0 \\
w^Tr^{(i)} + b > 0, if y^{(i)} = +1 \: and \: w^Tr^{(i)} + b < 0, if y^{(i)} = -1
\end{aligned}
\end{equation}

where w is the normal of the hyperplane and b is the bias. The minimization problem to obtain the optimal hyperplane that maximizes the margin can be formulated using.

\begin{equation}
\label{eq:SVM2}
Minimize \Phi(W) = \frac{1}{2}||W||^2, such \: that \: y_i(W.r_i + b) \geq 1
\end{equation}

\section{Performance Evaluation}

For experiments, we use the three models we have selected as being mostly used in the context of Diabetes prediction, that is Random Forest (RF), Logistic Regression (LR), and Support Vector Machine (SVM). We will do the experiments with and without Feature Selection. and with and without balancing.

We evaluate the models under study with and without features selection, before and after balancing, using the tenfold cross-validation method where the dataset is divided into k (k=10) partitions. One partition is for testing data and k-1 partitions are for training with replacement. This is repeated until each partition is used for training and testing. The resultant model is then obtained by averaging the result of each iteration.  For SVM, we use the polynomial kernels. Each model is executed 10 times on each dataset and the average for accuracy, F-measure, precision, recall, AUC, and execution time is calculated. The use of Accuracy as a comparative metric between the models is justified because the datasets are not heavily imbalanced. The accuracy, F-measure, recall, and precision are calculated using Equations \ref{eq:accuracy} and \ref{eq:fmeasure} respectively.  Recall and precision for the positive (negative) class are calculated using Equations \ref{eq:recall} and \ref{eq:precision} respectively.

\begin{equation}
\label{eq:accuracy}
Accuracy = \frac{TP + TN}{TP + FP + TN + FN}
\end{equation}

\begin{equation}
\label{eq:fmeasure}
F-measure = \frac{2(Recall \times Precision)}{Recall + Precision}
\end{equation}

\begin{equation}
\label{eq:recall}
Recall = \frac{TP(TN)}{TP(TN) + FN(FP)}
\end{equation}

\begin{equation}
\label{eq:precision}
Precision = \frac{TP(TN)}{TP(TN) + FP(FN)}
\end{equation}

where TP is True Positive, TN is True Negative, FP is False Positive, and FN is False Negative. TP (TN) represents the number of observations in the positive (negative) class that are classified as positive (negative), and FP (FN) represents the number of observations in the negative (positive) class that are classified as positive (negative).

We also calculate AUC. The Area Under the Curve (AUC) is the measure of the ability of a classifier to distinguish between classes and is used as a summary of the ROC curve. The higher the AUC, the better the performance of the model at distinguishing between the positive and negative classes.

\subsection{Hyperparameter Tuning}

To achieve the best performance possible with the end-to-end system for diabetes prediction, we also perform fine parameter tuning with the best algorithm we have selected.  We perform Hyperparameter tuning on the three AI models for the three datasets.  Hyperparameter tuning relies on experimental results and thus the best method to determine the optimal settings is to try many different combinations and evaluate the performance of each model. However, evaluating each model only on the training set can lead to overfitting. To reduce the effect of overfitting we perform again stratified k-fold Cross Validation with k =10.  The parameters we study for each algorithm, their ranges, and their optimal values are described in Table \ref{table:optimal values}.  The ranges are selected in a way that they include the values considered in literature.  To perform the search for the best parameters, we use GridSearchCV from python library sklearn.model\_selection module.

\begin{landscape}
\begin{table}[]
\begin{scriptsize}
\caption{Value(s) of Hyperparameters used and Optimal Values for Hyperparameters Obtained in our Experiments.}
\label{table:optimal values}
\begin{tabular}{cclcccccccccccc}
\hline
\rowcolor[HTML]{C0C0C0} 
\multicolumn{1}{|c|}{\cellcolor[HTML]{C0C0C0}}                                                        & \multicolumn{1}{c|}{\cellcolor[HTML]{C0C0C0}}                                              & \multicolumn{1}{c|}{\cellcolor[HTML]{C0C0C0}}                                                                                                                                         & \multicolumn{12}{c|}{\cellcolor[HTML]{C0C0C0}\textbf{Optimal values}}                                                                                                                                                                                                                                                                                                                                                                                                                                                                                                                                                                                                                                                    \\ \cline{4-15} 
\rowcolor[HTML]{C0C0C0} 
\multicolumn{1}{|c|}{\multirow{-2}{*}{\cellcolor[HTML]{C0C0C0}\textbf{Algorithm}}}                    & \multicolumn{1}{c|}{\multirow{-2}{*}{\cellcolor[HTML]{C0C0C0}\textbf{Hyperparameter}}}     & \multicolumn{1}{c|}{\multirow{-2}{*}{\cellcolor[HTML]{C0C0C0}\textbf{\begin{tabular}[c]{@{}c@{}}Values used in\\ our experiments\end{tabular}}}}                                      & \multicolumn{1}{c|}{\cellcolor[HTML]{C0C0C0}\textbf{1}} & \multicolumn{1}{c|}{\cellcolor[HTML]{C0C0C0}\textbf{2}} & \multicolumn{1}{c|}{\cellcolor[HTML]{C0C0C0}\textbf{3}} & \multicolumn{1}{c|}{\cellcolor[HTML]{C0C0C0}\textbf{4}} & \multicolumn{1}{c|}{\cellcolor[HTML]{C0C0C0}\textbf{5}} & \multicolumn{1}{c|}{\cellcolor[HTML]{C0C0C0}\textbf{6}} & \multicolumn{1}{c|}{\cellcolor[HTML]{C0C0C0}\textbf{7}} & \multicolumn{1}{c|}{\cellcolor[HTML]{C0C0C0}\textbf{8}} & \multicolumn{1}{c|}{\cellcolor[HTML]{C0C0C0}\textbf{9}} & \multicolumn{1}{c|}{\cellcolor[HTML]{C0C0C0}\textbf{10}} & \multicolumn{1}{c|}{\cellcolor[HTML]{C0C0C0}\textbf{11}} & \multicolumn{1}{c|}{\cellcolor[HTML]{C0C0C0}\textbf{12}} \\ \hline
\multicolumn{1}{|c|}{}                                                                                & \multicolumn{1}{c|}{\begin{tabular}[c]{@{}c@{}}Number of \\ estimators/trees\end{tabular}} & \multicolumn{1}{l|}{\begin{tabular}[c]{@{}l@{}}100 {[}7,13{]}, (300, 500, \\ 1000) {[}13{]}, 20, 40, 60, \\ 80, 100, 200, 300, 400, \\ 500, 600, 700, 800, \\ 900, 1000\end{tabular}} & \multicolumn{1}{c|}{50}                                 & \multicolumn{1}{c|}{40}                                 & \multicolumn{1}{c|}{50}                                 & \multicolumn{1}{c|}{50}                                 & \multicolumn{1}{c|}{20}                                 & \multicolumn{1}{c|}{100}                                & \multicolumn{1}{c|}{50}                                 & \multicolumn{1}{c|}{50}                                 & \multicolumn{1}{c|}{20}                                 & \multicolumn{1}{c|}{100}                                 & \multicolumn{1}{c|}{50}                                  & \multicolumn{1}{c|}{50}                                  \\ \cline{2-15} 
\multicolumn{1}{|c|}{}                                                                                & \multicolumn{1}{c|}{\begin{tabular}[c]{@{}c@{}}Splitting \\ criteria\end{tabular}}         & \multicolumn{1}{l|}{entropy and Gini}                                                                                                                                                 & \multicolumn{4}{c|}{entropy}                                                                                                                                                                                                          & \multicolumn{3}{c|}{Gini}                                                                                                                                                   & \multicolumn{1}{c|}{entropy}                            & \multicolumn{3}{c|}{Gini}                                                                                                                                                     & \multicolumn{1}{c|}{entropy}                             \\ \cline{2-15} 
\multicolumn{1}{|c|}{}                                                                                & \multicolumn{1}{c|}{\begin{tabular}[c]{@{}c@{}}Maximum\\ features\end{tabular}}            & \multicolumn{1}{l|}{\begin{tabular}[c]{@{}l@{}}Nmax$^{*}$, sqrt, and\\ log2\end{tabular}}                                                                                                  & \multicolumn{1}{c|}{None}                               & \multicolumn{2}{c|}{sqrt}                                                                                         & \multicolumn{1}{c|}{log2}                               & \multicolumn{8}{c|}{sqrt}                                                                                                                                                                                                                                                                                                                                                                                                                                                        \\ \cline{2-15} 
\multicolumn{1}{|c|}{\multirow{-4}{*}{\begin{tabular}[c]{@{}c@{}}Random\\ Forest\end{tabular}}}       & \multicolumn{1}{c|}{Max depth}                                                             & \multicolumn{1}{l|}{None, 2, 5, 8}                                                                                                                                                    & \multicolumn{2}{c|}{5}                                                                                            & \multicolumn{1}{c|}{None}                               & \multicolumn{1}{c|}{8}                                  & \multicolumn{8}{c|}{None}                                                                                                                                                                                                                                                                                                                                                                                                                                                        \\ \hline
\multicolumn{1}{|c|}{\begin{tabular}[c]{@{}c@{}}Support\\ Vector\\ Machine\end{tabular}}              & \multicolumn{1}{c|}{\begin{tabular}[c]{@{}c@{}}Regularization\\ parameter\end{tabular}}    & \multicolumn{1}{l|}{\begin{tabular}[c]{@{}l@{}}(0.001, 0.01, 0.1, \\ 1, 2, 3, 5, 7, 10) {[}13{]}, \\ 4, 6, 8, 9, 10\end{tabular}}                                                     & \multicolumn{3}{c|}{1}                                                                                                                                                      & \multicolumn{1}{c|}{7}                                  & \multicolumn{8}{c|}{1}                                                                                                                                                                                                                                                                                                                                                                                                                                                           \\ \hline
\multicolumn{1}{|c|}{}                                                                                & \multicolumn{1}{c|}{\begin{tabular}[c]{@{}c@{}}Regularization\\ parameter\end{tabular}}    & \multicolumn{1}{l|}{\begin{tabular}[c]{@{}l@{}}2$^{-6}$, 2$^{-4}$, 2$^{-2}$, 2$^{0}$,\\ 2$^{2}$, 2$^{4}$, 2$^{6}$\end{tabular}}                                                                                          & \multicolumn{2}{c|}{16}                                                                                           & \multicolumn{1}{c|}{0.25}                               & \multicolumn{1}{c|}{4}                                  & \multicolumn{1}{c|}{16}                                 & \multicolumn{3}{c|}{4}                                                                                                                                                      & \multicolumn{1}{c|}{16}                                 & \multicolumn{3}{c|}{4}                                                                                                                                                         \\ \cline{2-15} 
\multicolumn{1}{|c|}{\multirow{-2}{*}{\begin{tabular}[c]{@{}c@{}}Logistic\\ Regression\end{tabular}}} & \multicolumn{1}{c|}{Solver}                                                                & \multicolumn{1}{l|}{\begin{tabular}[c]{@{}l@{}}Newton-cg, lbfgs,\\ liblinear, sag, and\\ saga\end{tabular}}                                                                           & \multicolumn{1}{c|}{lbfgs}                              & \multicolumn{2}{c|}{liblinear}                                                                                    & \multicolumn{9}{c|}{lbfgs}                              \\ \hline
\multicolumn{15}{l}{\begin{tabular}[c]{@{}l@{}}Nmax$^{*}$: Number of features in the dataset, Newton-cg: Newton Conjugate Gradient,\\ 1 - PIMA Indian: no feature selection and no balancing, 2 - PIMA Indian: feature selection and no balancing\\ 3 - PIMA Indian: feature selection and balancing, 4 - PIMA India: no feature selection and balancing,\\ 5 - Sylhet: no feature selection and no balancing, 6 - Sylhet: feature selection and no balancing,\\ 7 - Sylhet: feature selection and balancing, 8 - Sylhet: no feature selection and balancing,\\ 9 - MIMIC IIII: no feature selection and no balancing, 10 - MIMIC III: feature selection and no balancing,\\ 11 - MIMIC III: feature selection and balancing, 12: MIMIC III - no feature selection and balancing\end{tabular}}                                                                                                                                                                                                                                                                                                                            
\end{tabular}
\end{scriptsize}
\end{table}
\end{landscape}

\subsection{Feature Selection}

Following the feature selection method described in 5.4, we give the results for the three datasets.

\begin{figure}
\centering
\begin{subfigure}{0.45\textwidth}
    \includegraphics[width=\textwidth]{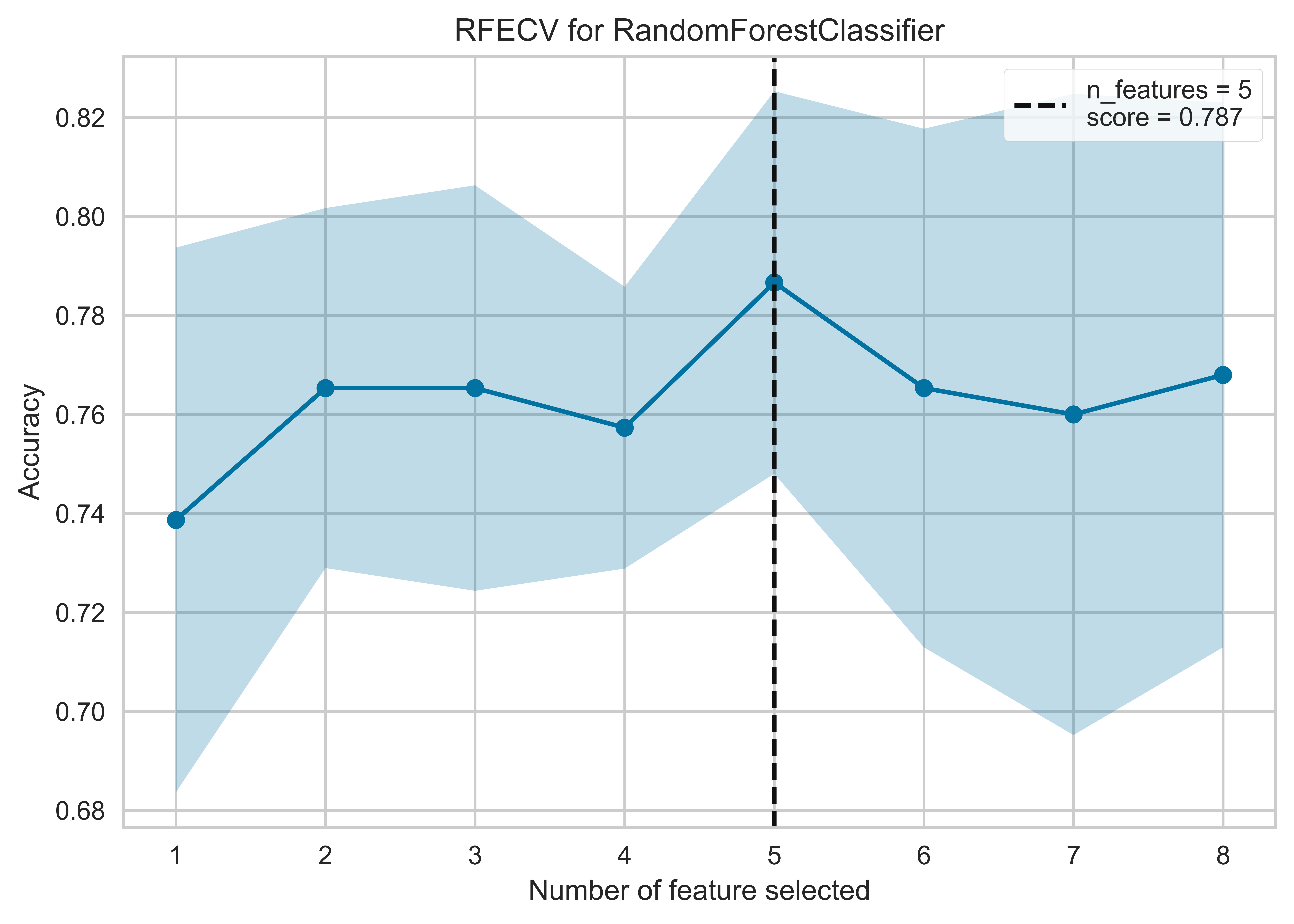}
    \caption{RFECV Performance}
    \label{fig:feature selection PIMA indian (a)}
\end{subfigure}
\hfill
\begin{subfigure}{0.45\textwidth}
    \includegraphics[width=\textwidth]{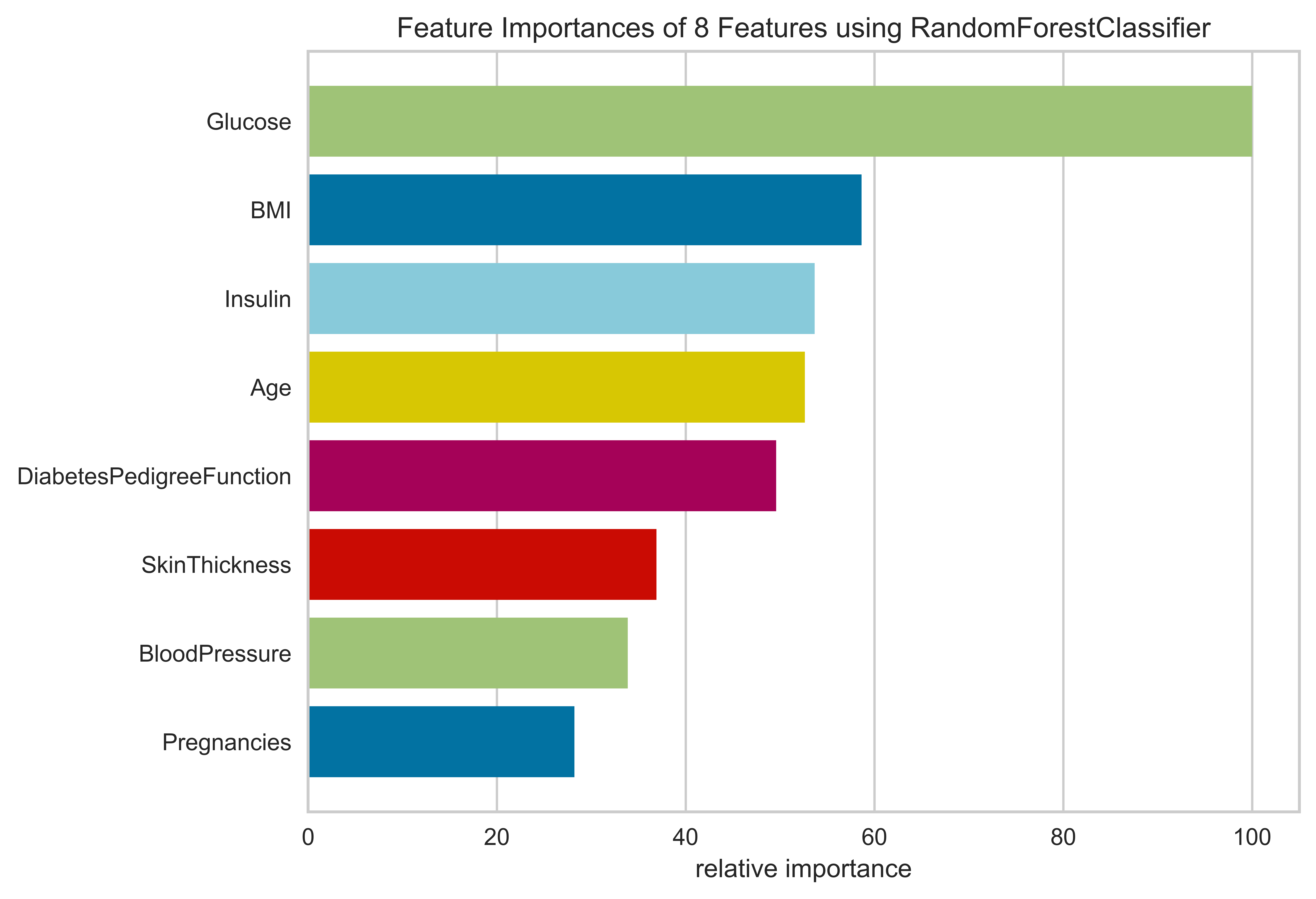}
    \caption{Importance of Features}
    \label{fig:feature selection PIMA indian (b)}
\end{subfigure}
        
\caption{Performance of Feature Selection Algorithms for PIMA Indian Dataset.}
\end{figure}

For the PIMA Indian dataset, there are 5 selected features: glucose, BMI, insulin, age, and diabetes pedigree function (Figure \ref{fig:feature selection PIMA indian (a)}).  Figure \ref{fig:feature selection PIMA indian (b)} shows the importance of each feature to prediction.  It shows that glucose is the most important feature for the prevalence/incidence of diabetes in users, followed by BMI, insulin, age, and diabetes pedigree function.  This is confirmed by studies in literature \cite{boden2004lipids, action2008effects} and type 2 diabetes risk assessment form by the Finnish Diabetes Association \cite{eRiskite84:online}.

\begin{figure}
\centering
\begin{subfigure}{0.45\textwidth}
    \includegraphics[width=\textwidth]{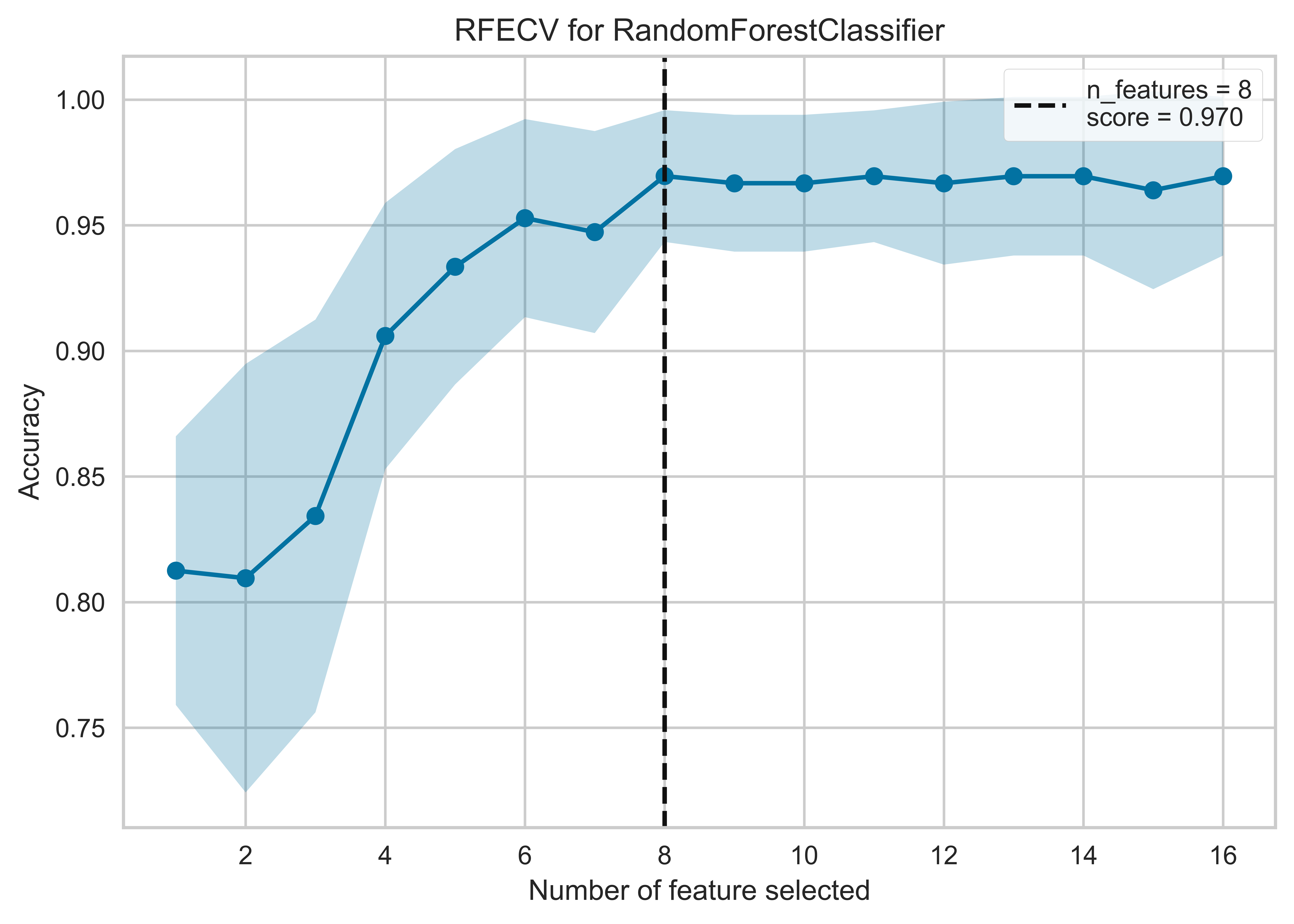}
    \caption{RFECV Performance}
    \label{fig:feature selection SYLHET (a)}
\end{subfigure}
\hfill
\begin{subfigure}{0.45\textwidth}
    \includegraphics[width=\textwidth]{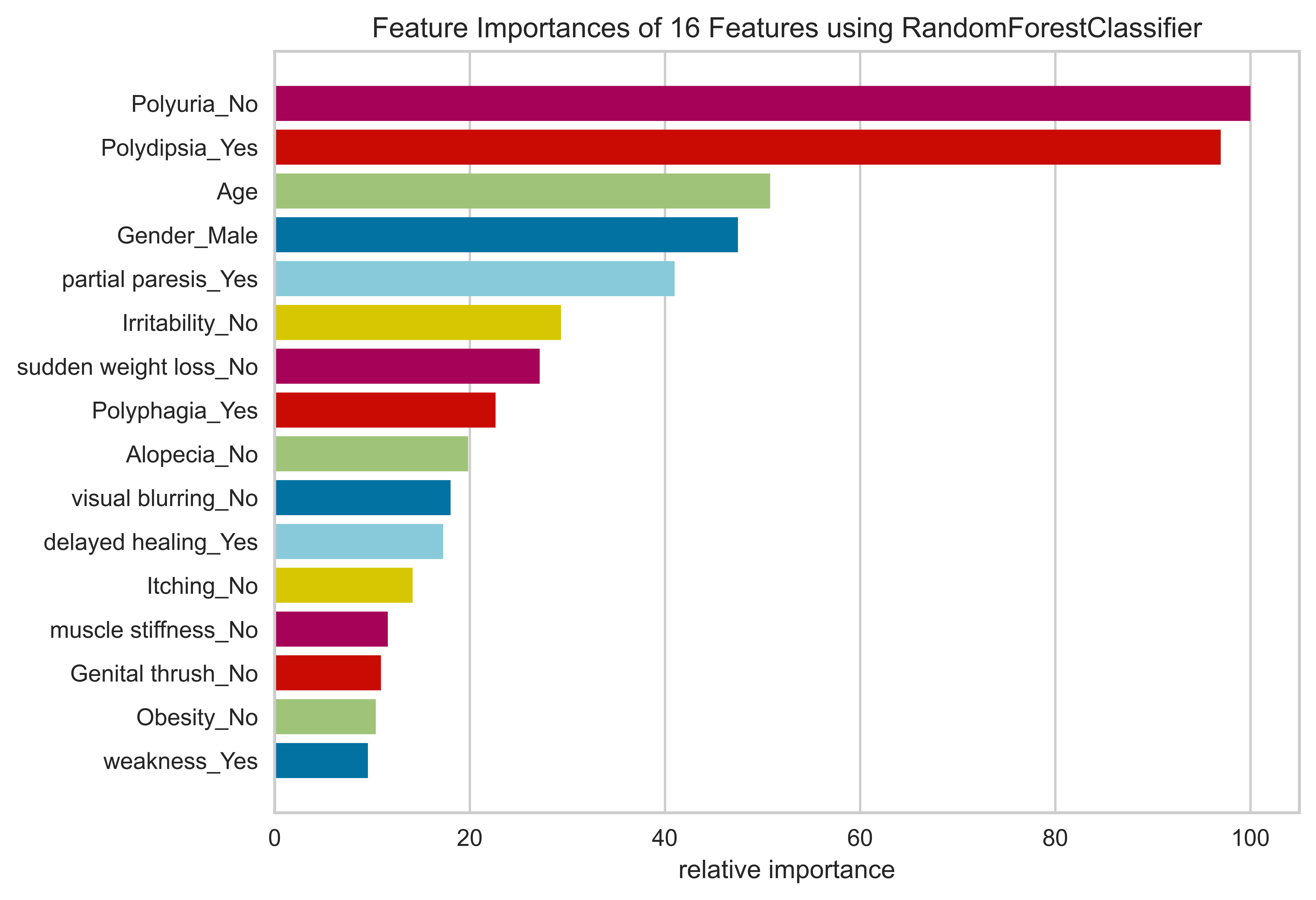}
    \caption{Importance of Features}
    \label{fig:feature selection SYLHET (b)}
\end{subfigure}
        
\caption{Performance of Feature Selection Algorithms for Sylhet Dataset.}
\end{figure}

For the Sylhet dataset, there are 8 selected features: polyuria, polydipsia, age, gender, partial paresis, irritability, sudden weight loss, and polyphagia (Figure \ref{fig:feature selection SYLHET (a)}). Figure \ref{fig:feature selection SYLHET (b)} shows the importance of each feature to prediction.  It shows that polyuria and polydipsia are the most important features in the prevalence/incidence of diabetes in users.  This is in alignment with the result obtained in the literature \cite{kumar2014type}.  In the context of gender, figure reveals that men are more correlated with the prevalence/incidence of diabetes.  This is confirmed by the American Diabetes Association's type 2 diabetes risk test \cite{DIABETES14:online}.

\begin{figure}
\centering
\begin{subfigure}{0.45\textwidth}
    \includegraphics[width=\textwidth]{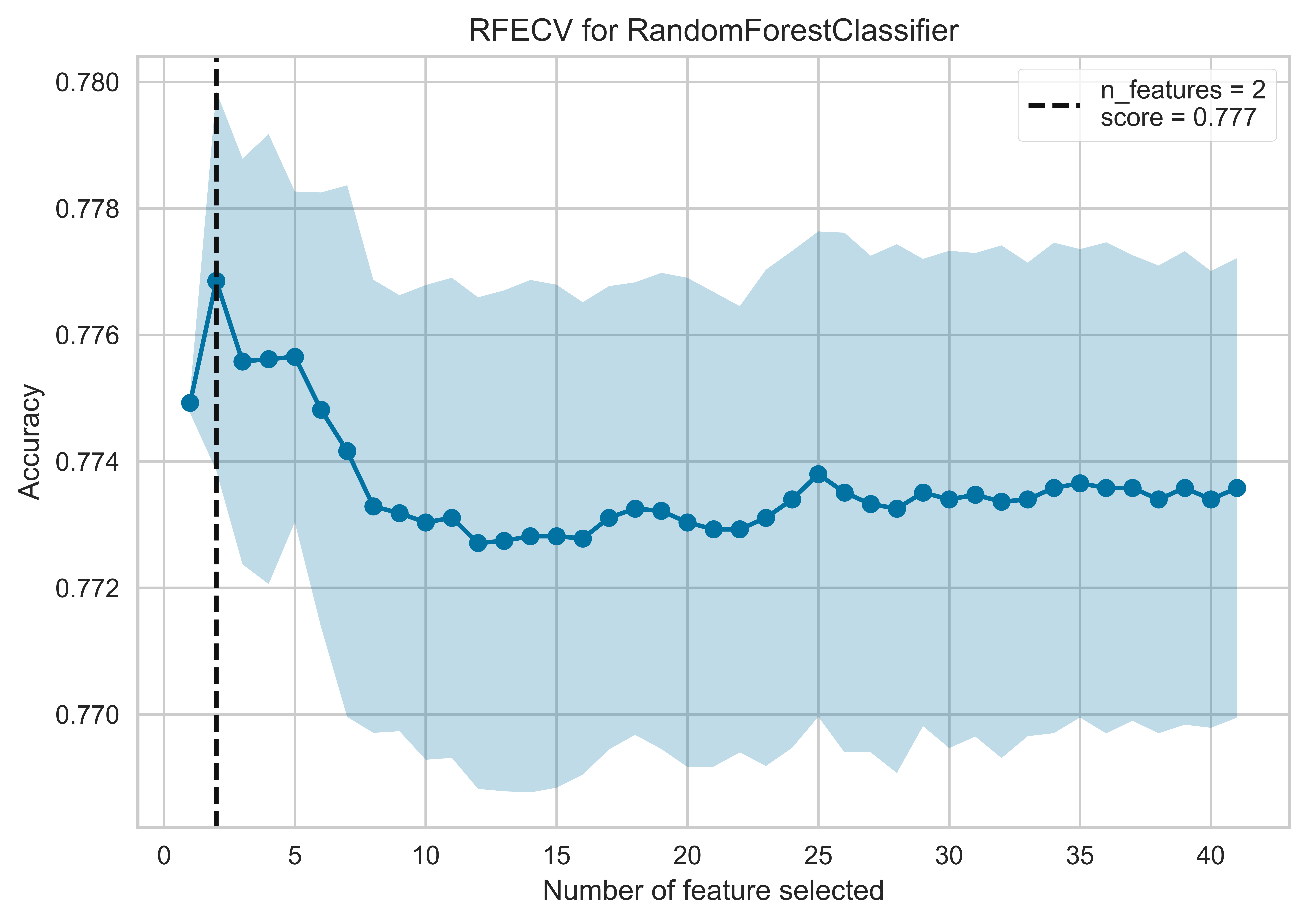}
    \caption{RFECV Performance}
    \label{fig:feature selection MIMIC-III (a)}
\end{subfigure}
\hfill
\begin{subfigure}{0.45\textwidth}
    \includegraphics[width=\textwidth]{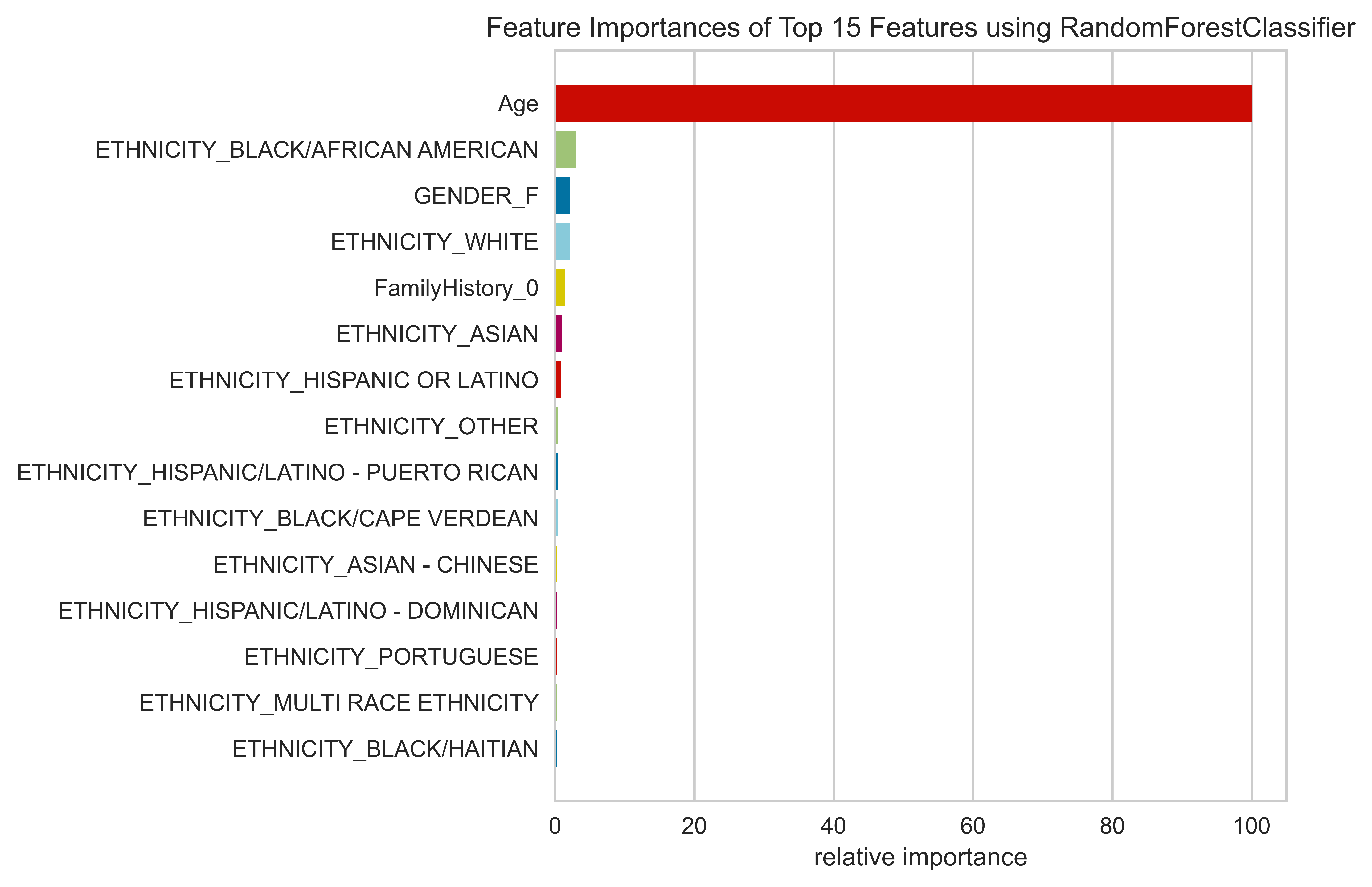}
    \caption{Importance of Features}
    \label{fig:feature selection MIMIC-III (b)}
\end{subfigure}
        
\caption{Performance of Feature Selection Algorithms for MIMIC III Dataset.}
\end{figure}

In the MIMIC III dataset, there are 4 attributes which all related to diabetes risk factors: gender, age, ethnicity, and family history of diabetes. The data preparation stage transforms the categorical feature Ethnicity into several binary features which explain the 'ETHNICITY\_xx' feature names.  Figure \ref{fig:feature selection MIMIC-III (a)} shows that two features are selected as significant.  The feature importance Graph Figure \ref{fig:feature selection MIMIC-III (b)} shows that age has the highest importance for the prevalence/incidence of type 2 diabetes in the population.  This is in alignment with the American Diabetes Association's type 2 diabetes risk test \cite{DIABETES14:online}.  The second important feature is the Black/African American ethnicity.  This is also confirmed by studies in literature \cite{shai2006ethnicity,jackson2013association,zizi2012race}.  Furthermore, the gender feature has low importance and consequently, it was not selected.

\subsection{Hardware and Execution Time}

\begin{figure}
\centering
\begin{subfigure}{0.3\textwidth}
    \includegraphics[width=\textwidth]{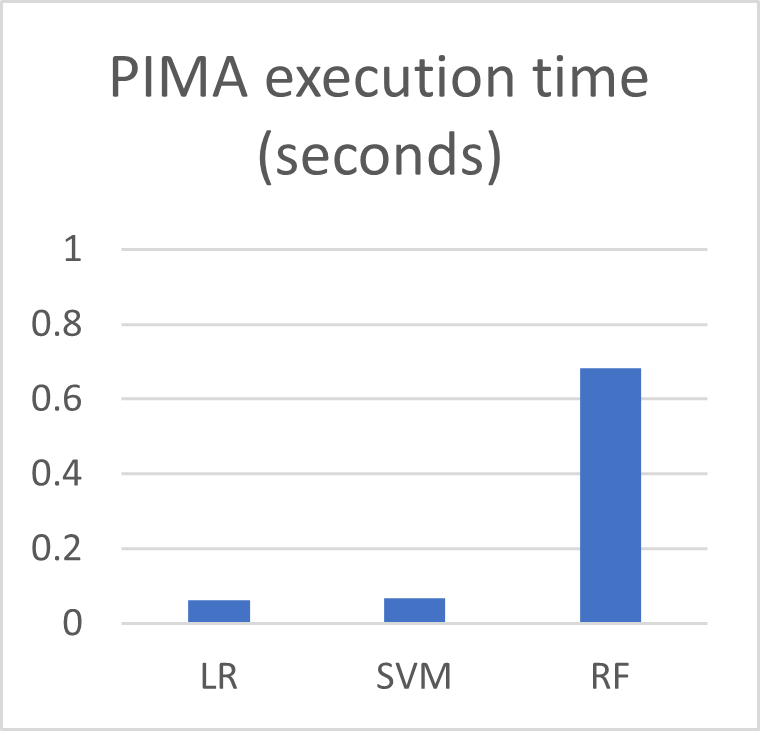}
    \caption{PIMA Indian}
    \label{fig:time PIMA indian}
\end{subfigure}
\hfill
\begin{subfigure}{0.3\textwidth}
    \includegraphics[width=\textwidth]{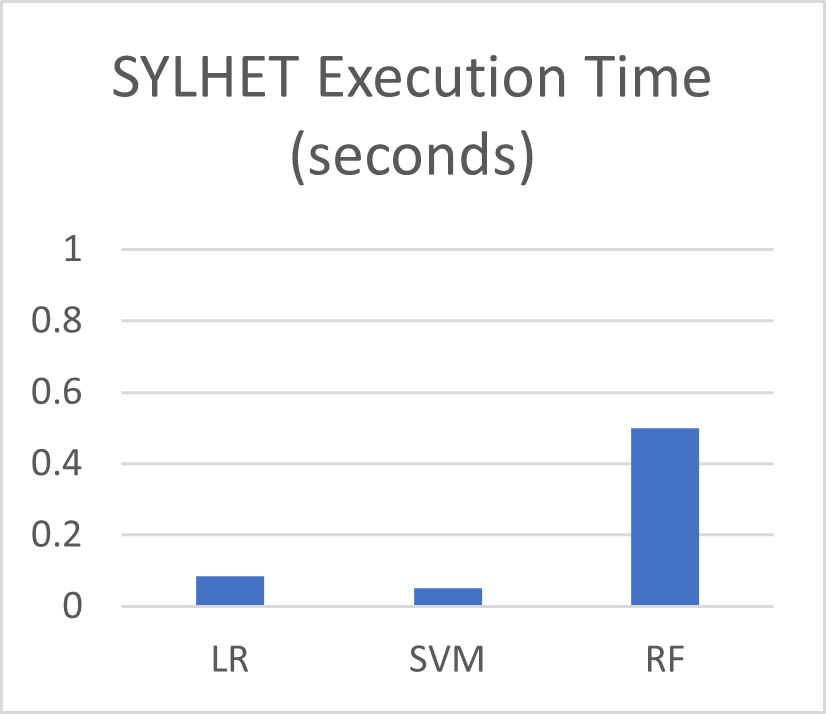}
    \caption{Sylhet}
    \label{fig:time SYLHET}
\end{subfigure}
\hfill
\begin{subfigure}{0.3\textwidth}
    \includegraphics[width=\textwidth]{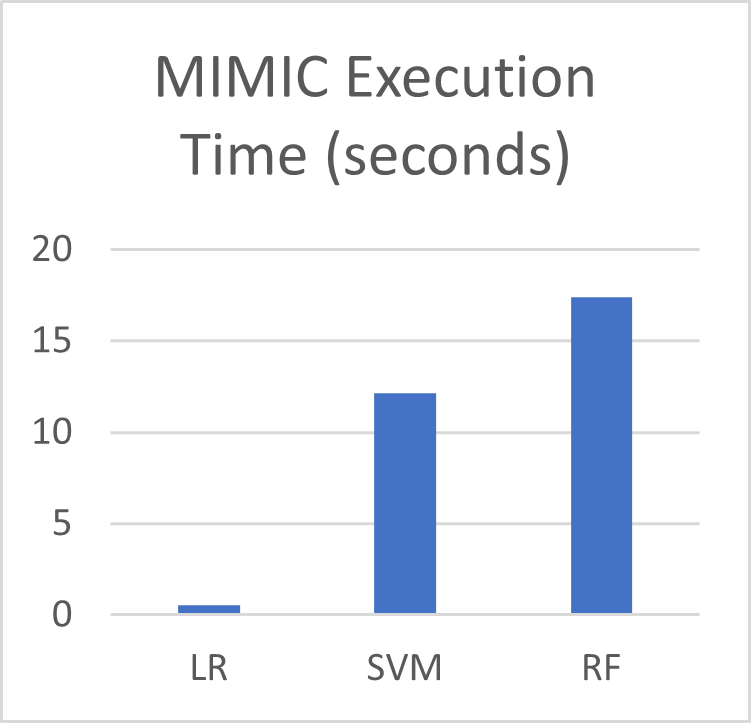}
    \caption{MIMIC III}
    \label{fig:time MIMIC III}
\end{subfigure}
        
\caption{Execution Time for Logistic Regression, Support Vector Machine, and Random Forest Algorithms for the Datasets under study.}
\end{figure}

The Hardware used for the performance analysis is Intel(R) Core (TM) i7-9700,  with 32 Kilobytes of L1 Data-cache, 32 Kilobytes of L1 Instruction-cache, 256 Kilobytes of L2 Cache, 12 Megabytes of L3 Cache.  The total execution times of each machine learning model under study for PIMA Indian, Sylhet, and MIMIC III datasets are shown in Figures \ref{fig:time PIMA indian}, \ref{fig:time SYLHET}, and \ref{fig:time MIMIC III} respectively.  The measurements have been done using the tuned parameters for each model. It consists of the total time for training and validating the model. We can observe that Random Forest (RF) uses more CPU. The main reason is that number of estimators (n\_estimator) is the principal parameter driving computational usage.

\subsection{Experimental Results Analysis}

\begin{figure}
\centering
\begin{subfigure}{0.5\textwidth}
    \includegraphics[width=0.8\textwidth]{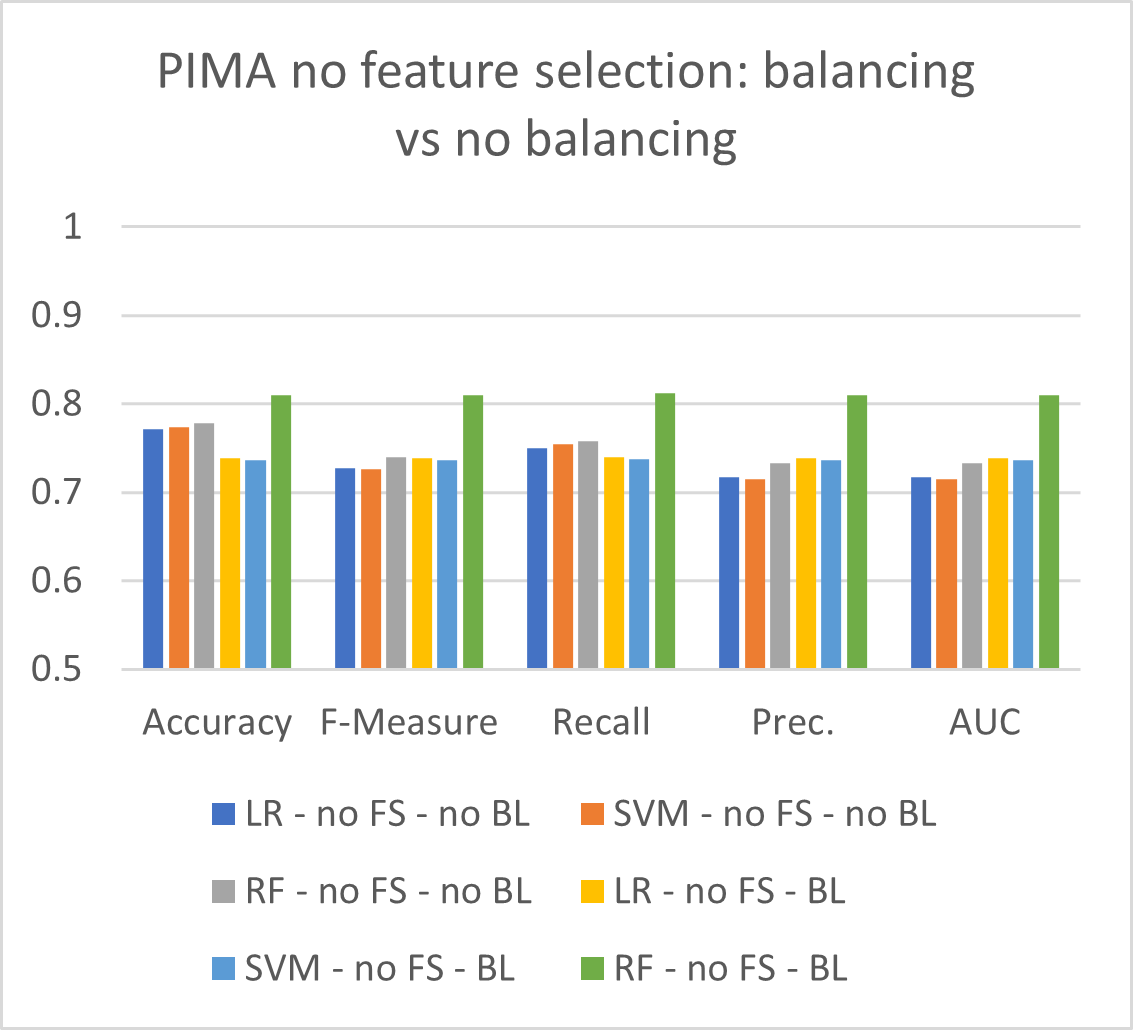}
    \caption{No Feature Selection: Balancing versus No Balancing}
    \label{fig:PIMA No FS}
\end{subfigure}
\hfill
\begin{subfigure}{0.5\textwidth}
    \includegraphics[width=0.8\textwidth]{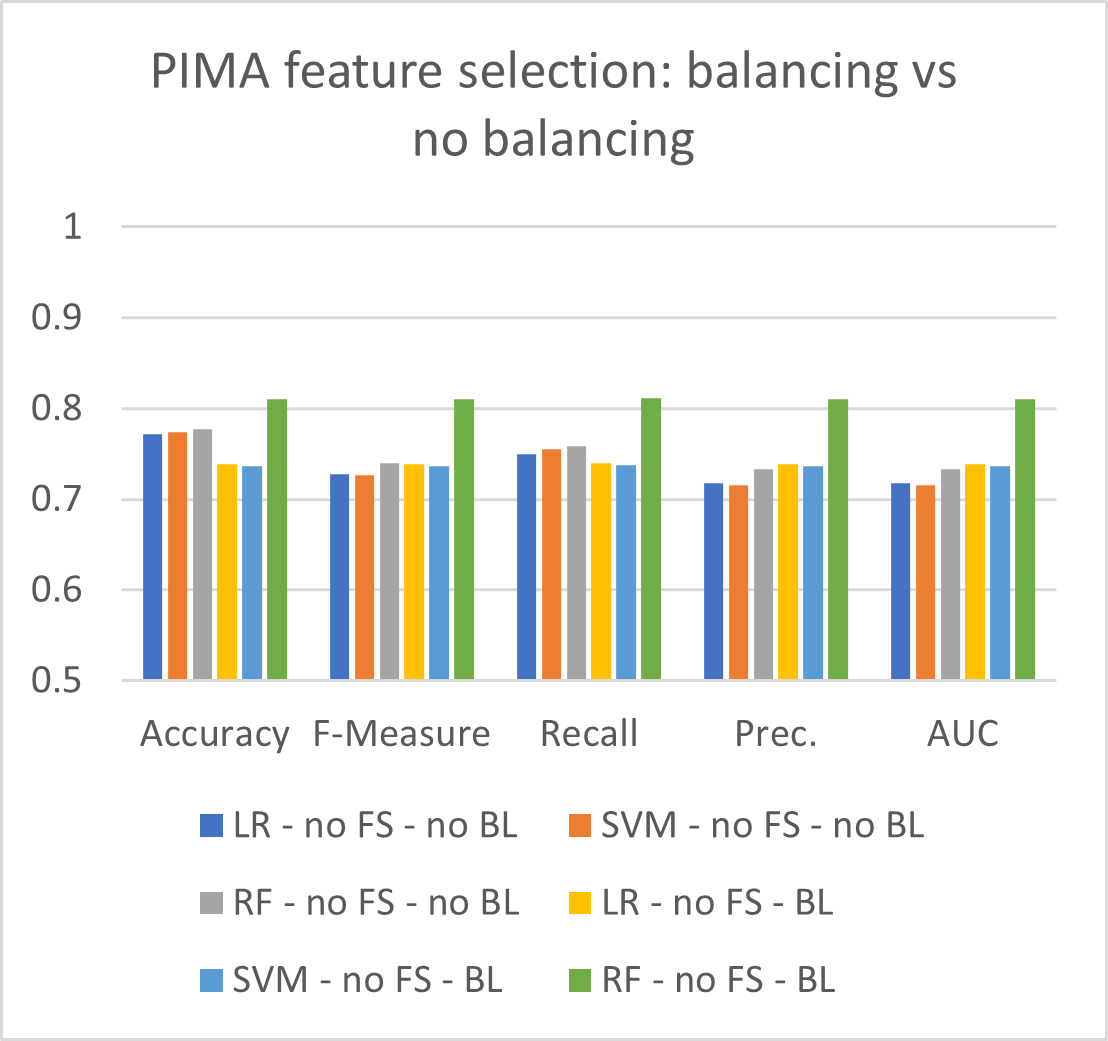}
    \caption{Feature Selection: Balancing versus No Balancing}
    \label{fig:PIMA FS}
\end{subfigure}
\hfill
\begin{subfigure}{0.5\textwidth}
    \includegraphics[width=0.8\textwidth]{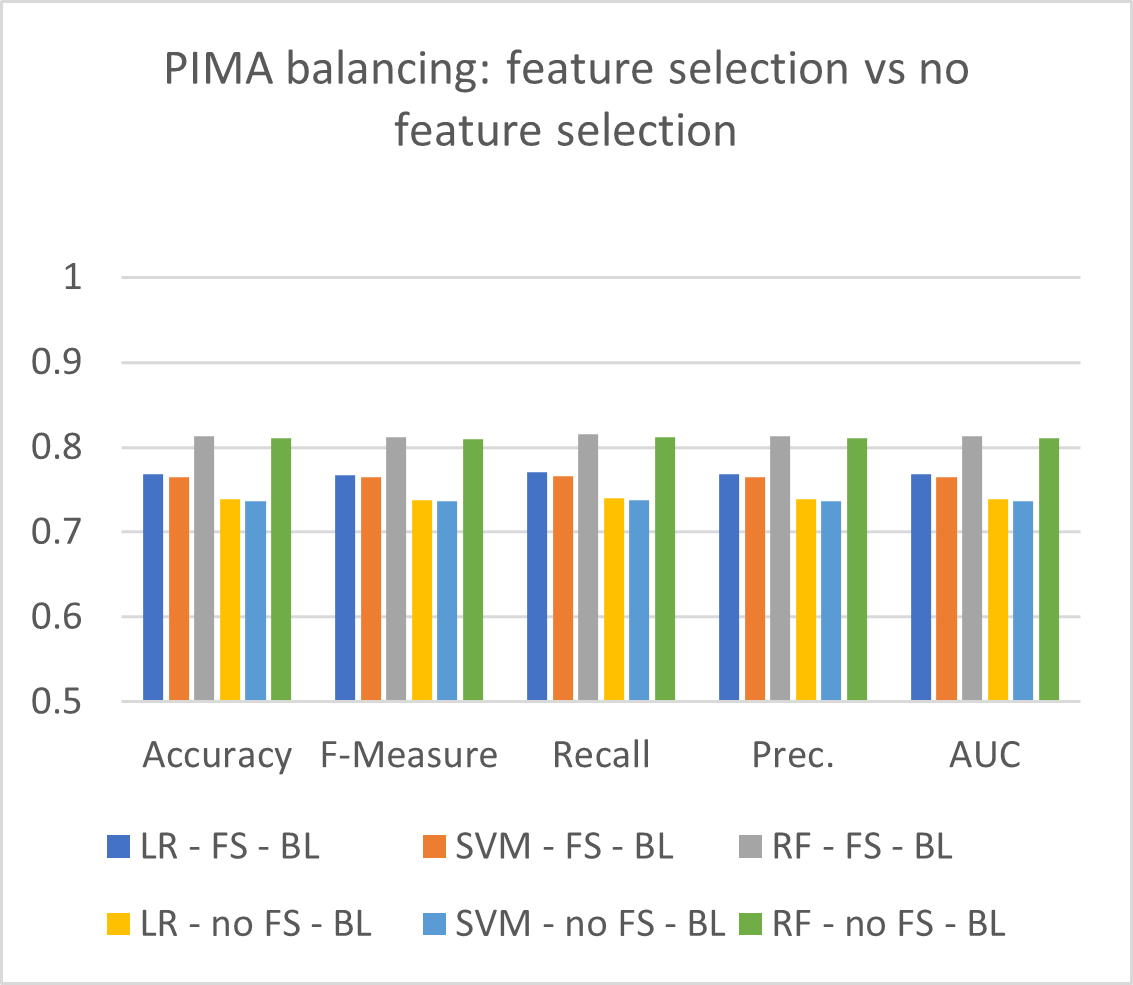}
    \caption{Balancing: Feature selection versus No Feature Selection}
    \label{fig:PIMA BAL}
\end{subfigure}

\caption{Comparison of Accuracy, F-measure, Recall, Precision, and AUC for the algorithms under study on PIMA Indian dataset (FS: Feature Selection, BL: Data Balancing).}
\label{fig:performance PIMA}
\end{figure}

\begin{figure}
\centering
\begin{subfigure}{0.5\textwidth}
    \includegraphics[width=0.8\textwidth]{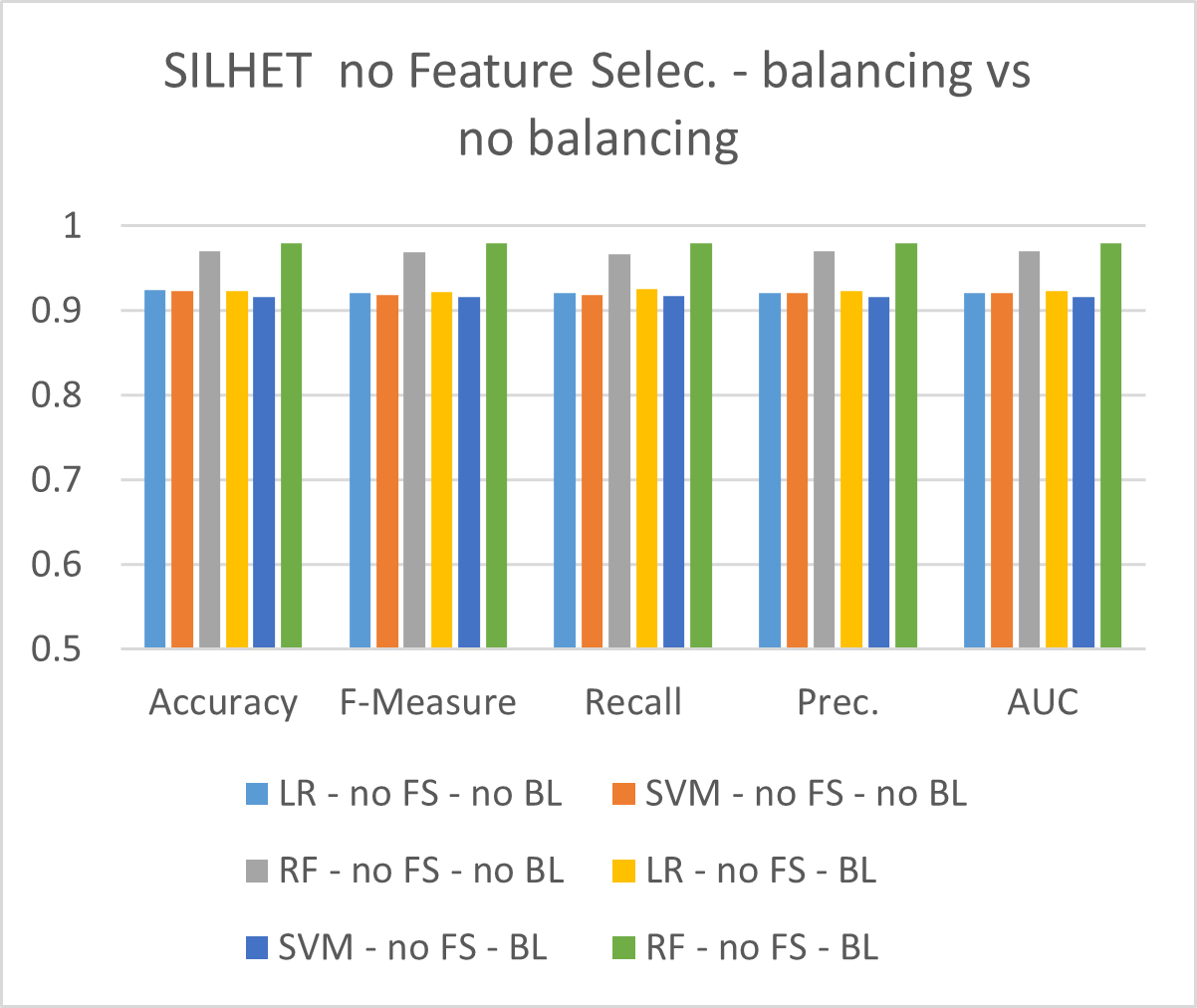}
    \caption{No Feature Selection: Balancing versus No Balancing}
    \label{fig:SYLHET No FS}
\end{subfigure}
\hfill
\begin{subfigure}{0.5\textwidth}
    \includegraphics[width=0.8\textwidth]{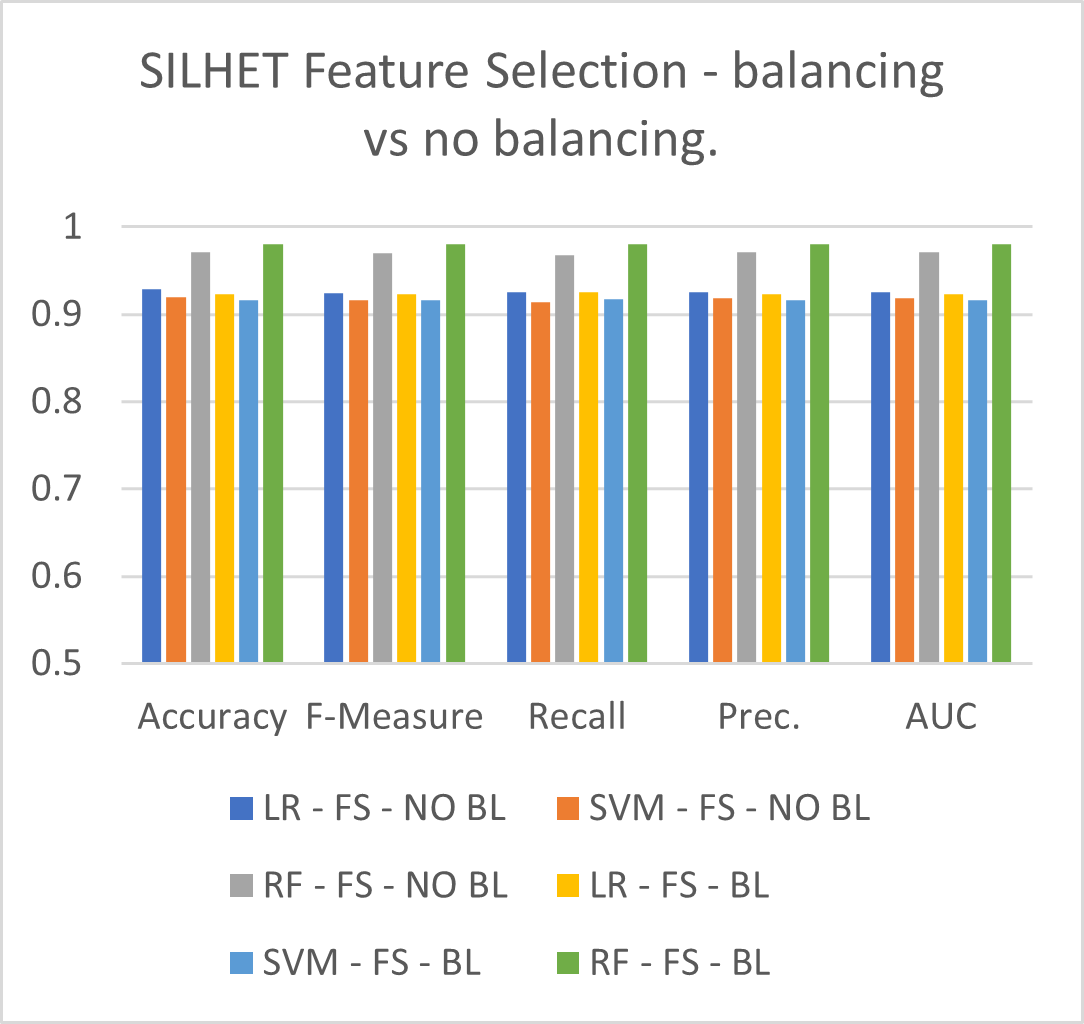}
    \caption{Feature Selection: Balancing versus No Balancing}
    \label{fig:SYLHET FS}
\end{subfigure}
\hfill
\begin{subfigure}{0.5\textwidth}
    \includegraphics[width=0.8\textwidth]{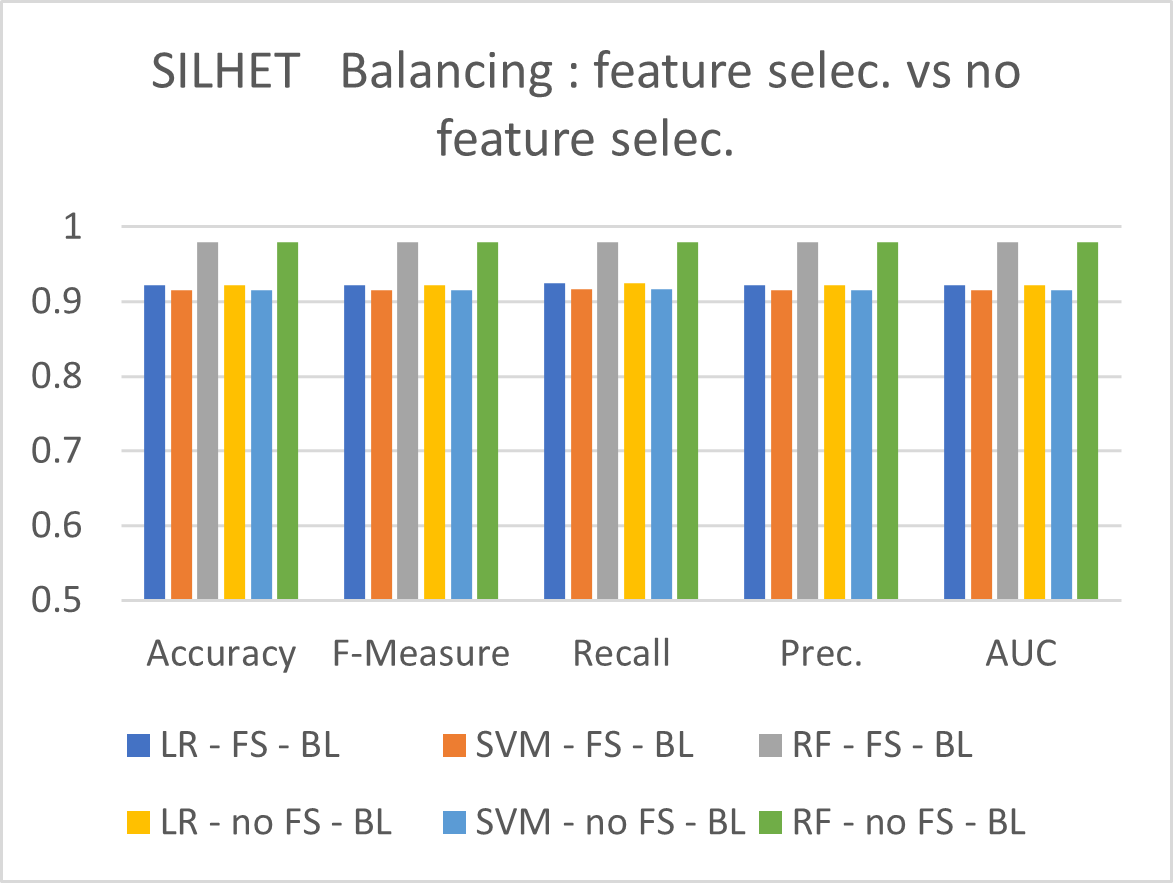}
    \caption{Balancing: Feature Selection versus No Feature Selection}
    \label{fig:SYLHET BAL}
\end{subfigure}
   
\caption{Comparison of Accuracy, F-measure, Recall, Precision, and AUC for the algorithms under study on Sylhet dataset (FS: Feature Selection, BL: Data Balancing).}
\label{fig:performance Sylhet}
\end{figure}

\begin{figure}
\centering
\begin{subfigure}{0.5\textwidth}
    \includegraphics[width=0.8\textwidth]{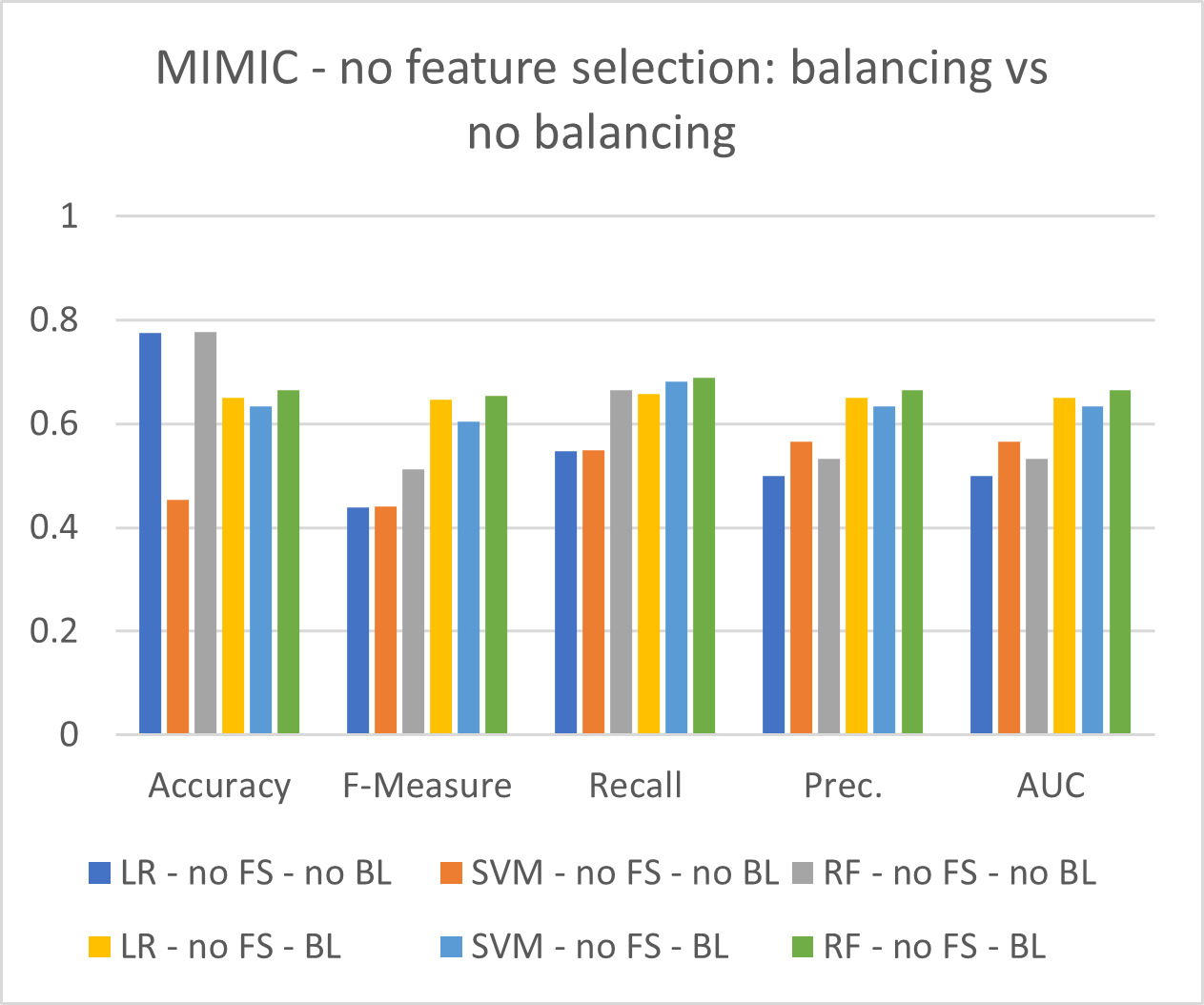}
    \caption{No Feature Selection: Balancing versus No Balancing}
    \label{fig:MIMIC No FS}
\end{subfigure}
\hfill
\begin{subfigure}{0.5\textwidth}
    \includegraphics[width=0.8\textwidth]{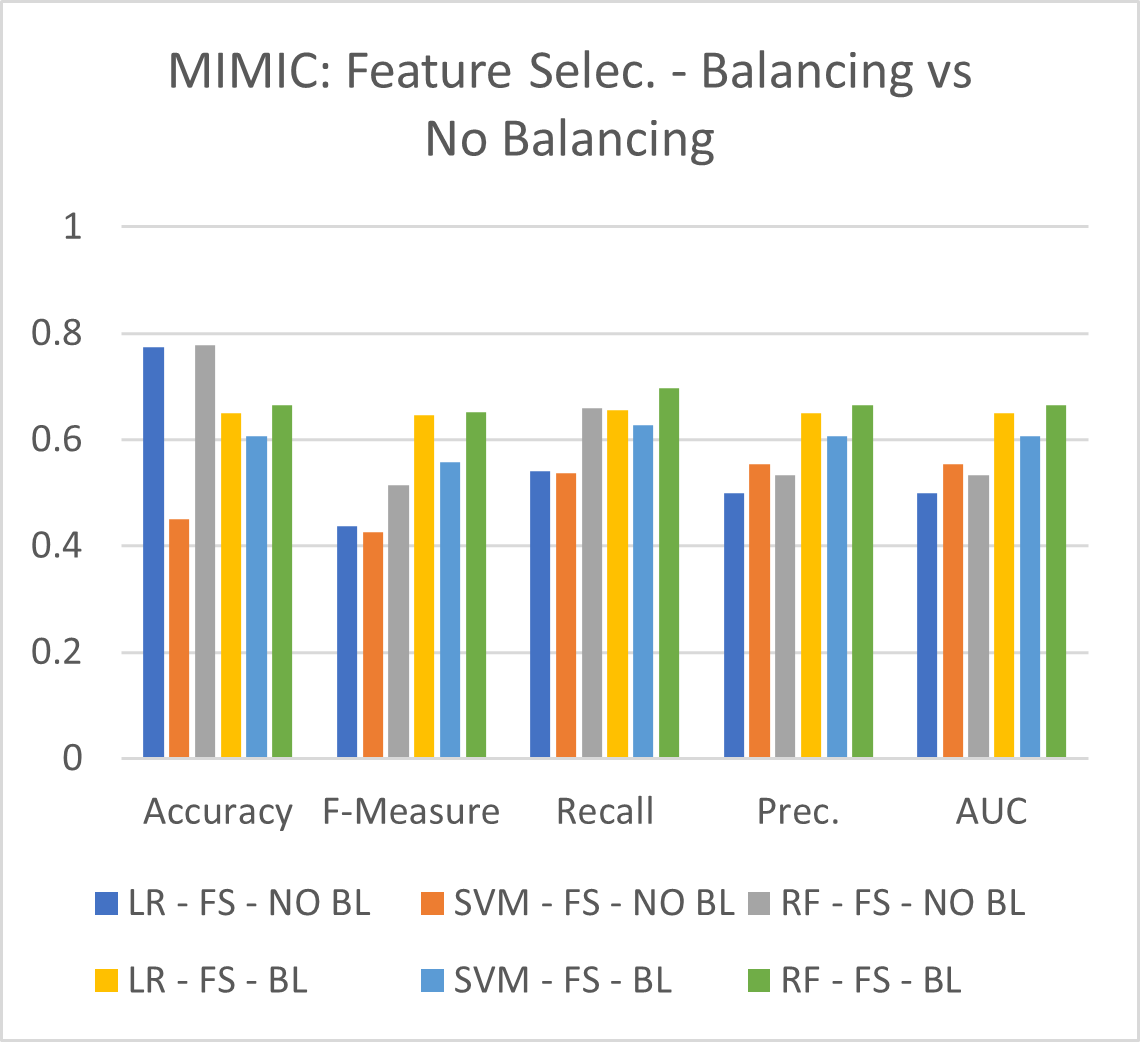}
    \caption{Feature Selection: Balancing versus No Balancing}
    \label{fig:MIMIC FS}
\end{subfigure}
\hfill
\begin{subfigure}{0.5\textwidth}
    \includegraphics[width=0.8\textwidth]{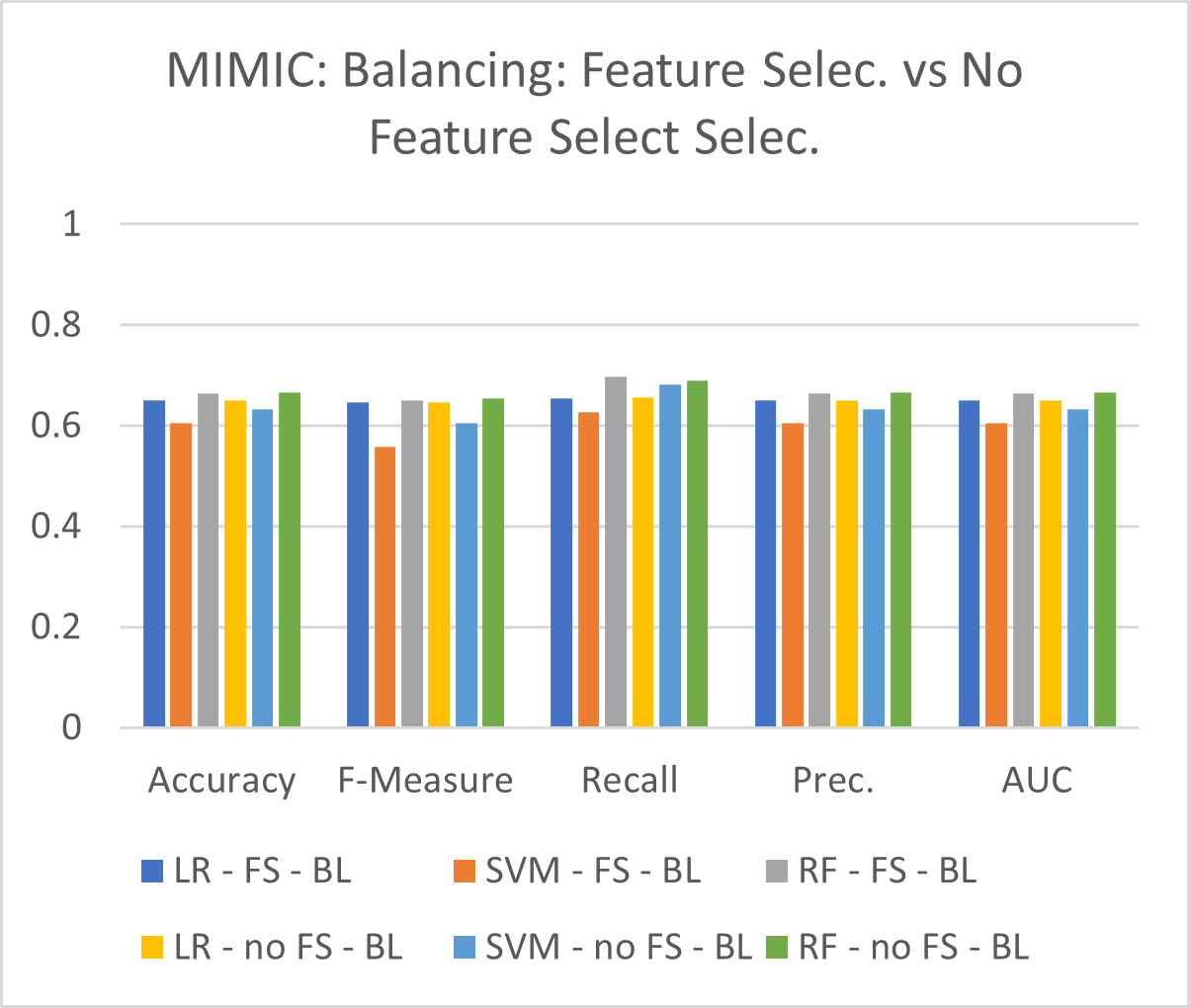}
    \caption{Balancing: Feature Selection versus No Feature Selection}
    \label{fig:MIMIC BAL}
\end{subfigure}

\caption{Comparison of Accuracy, F-measure, Recall, Precision, and AUC for the algorithms under study on MIMIC III dataset (FS: Feature Selection, BL: Data Balancing).}
\label{fig:performance MIMIC III}
\end{figure}

\begin{figure}
\centering
\begin{subfigure}{0.9\textwidth}
    \includegraphics[width=\textwidth]{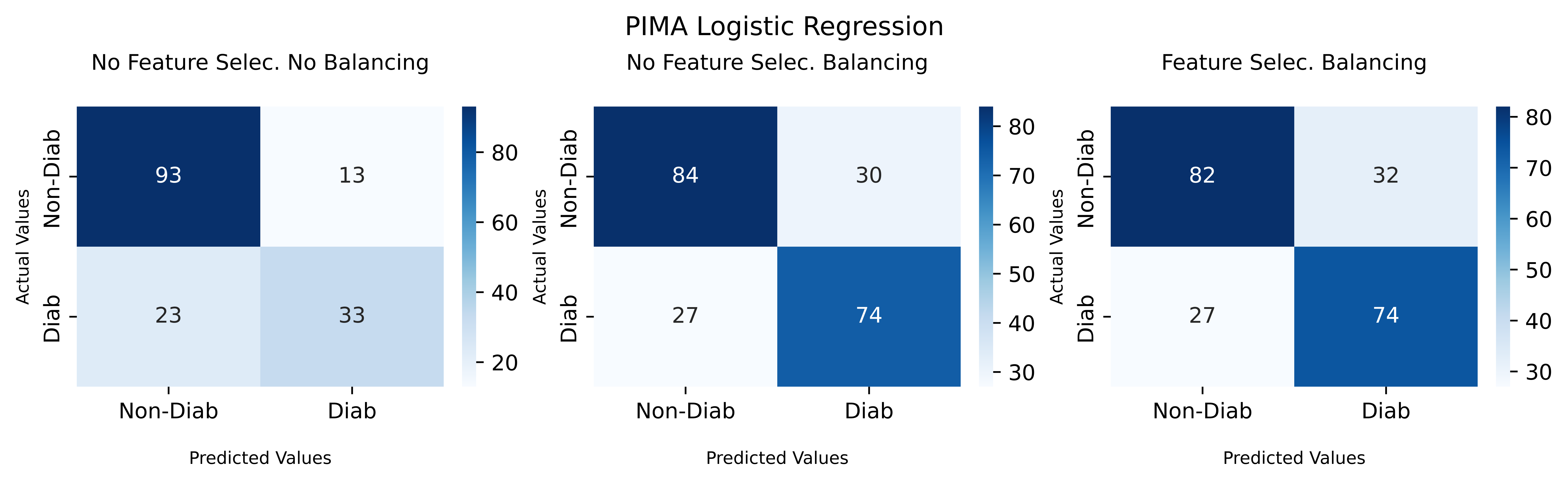}
    \caption{PIMA Indian: Logistic Regression}
    \label{fig:confusion matrix PIMA LR}
\end{subfigure}
\hfill
\begin{subfigure}{0.9\textwidth}
    \includegraphics[width=\textwidth]{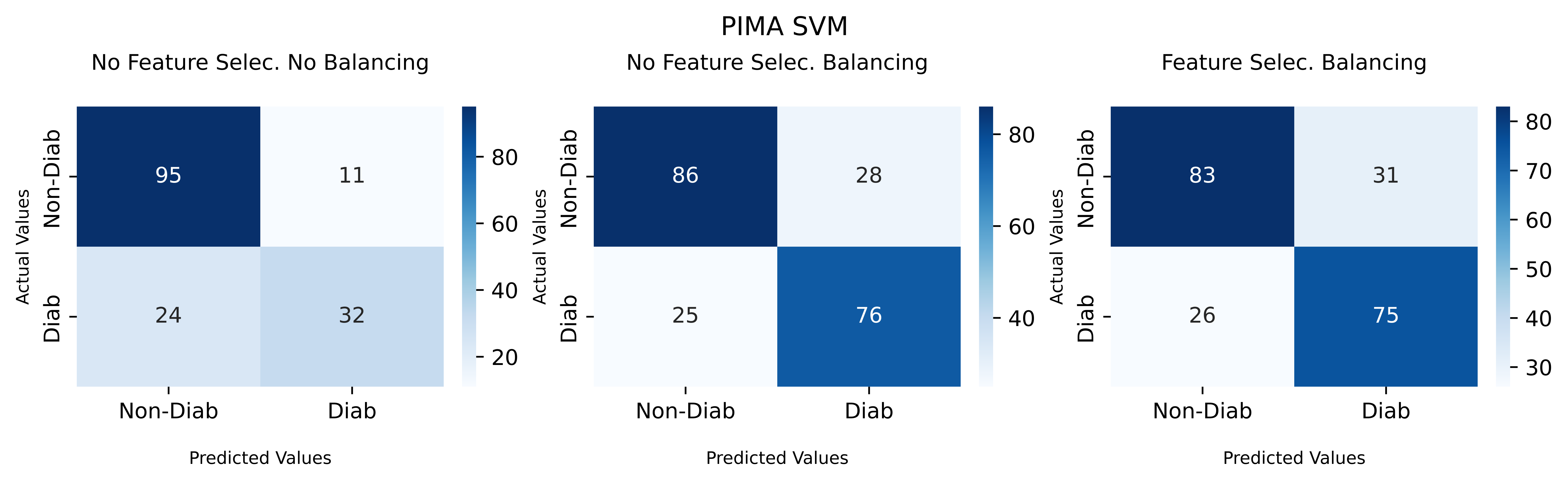}
    \caption{PIMA Indian: Support Vector Machine}
    \label{fig:confusion matrix PIMA SVM}
\end{subfigure}
\hfill
\begin{subfigure}{0.9\textwidth}
    \includegraphics[width=\textwidth]{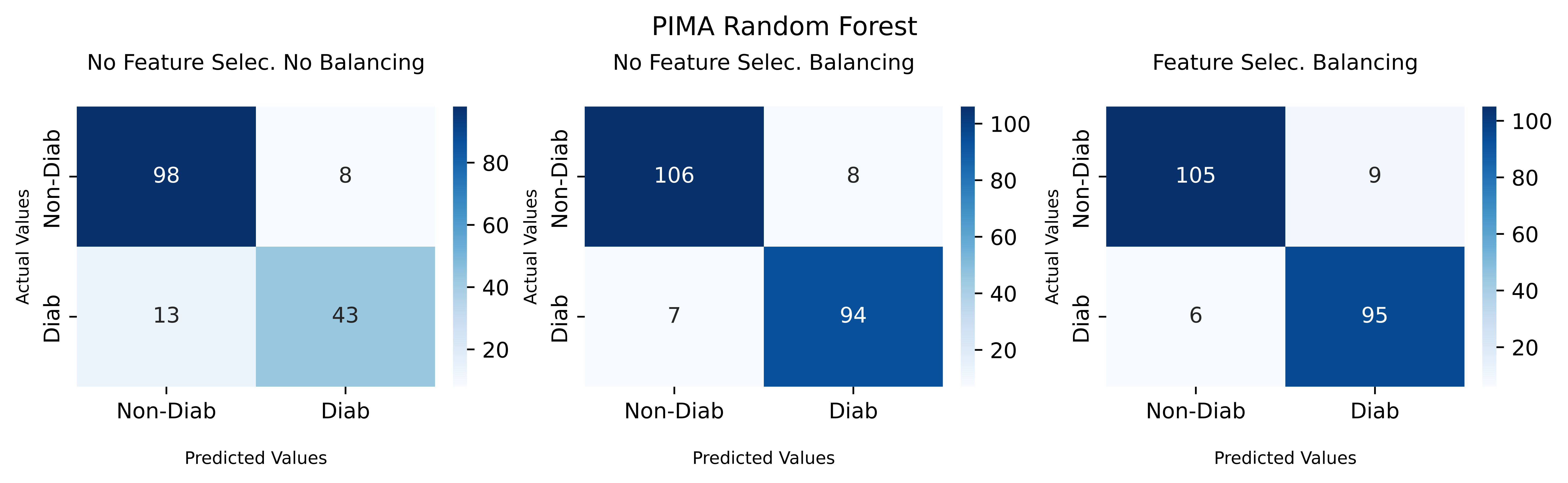}
    \caption{PIMA Indian: Random Forest}
    \label{fig:confusion matrix PIMA RF}
\end{subfigure}

\caption{Confusion Matrices for PIMA Indian dataset.}
\label{fig:confusion matrix PIMA}
\end{figure}

\begin{figure}
\centering
\begin{subfigure}{0.9\textwidth}
    \includegraphics[width=\textwidth]{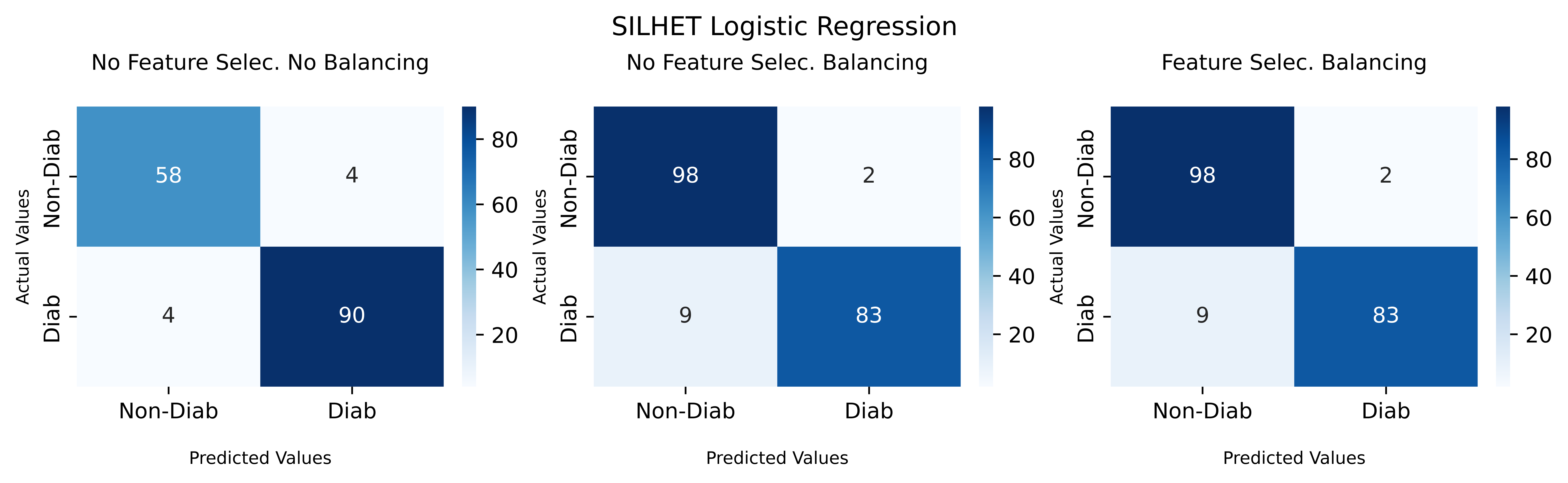}
    \caption{Sylhet: Logistic Regression}
    \label{fig:confusion matrix SYLHET LR}
\end{subfigure}
\hfill
\begin{subfigure}{0.9\textwidth}
    \includegraphics[width=\textwidth]{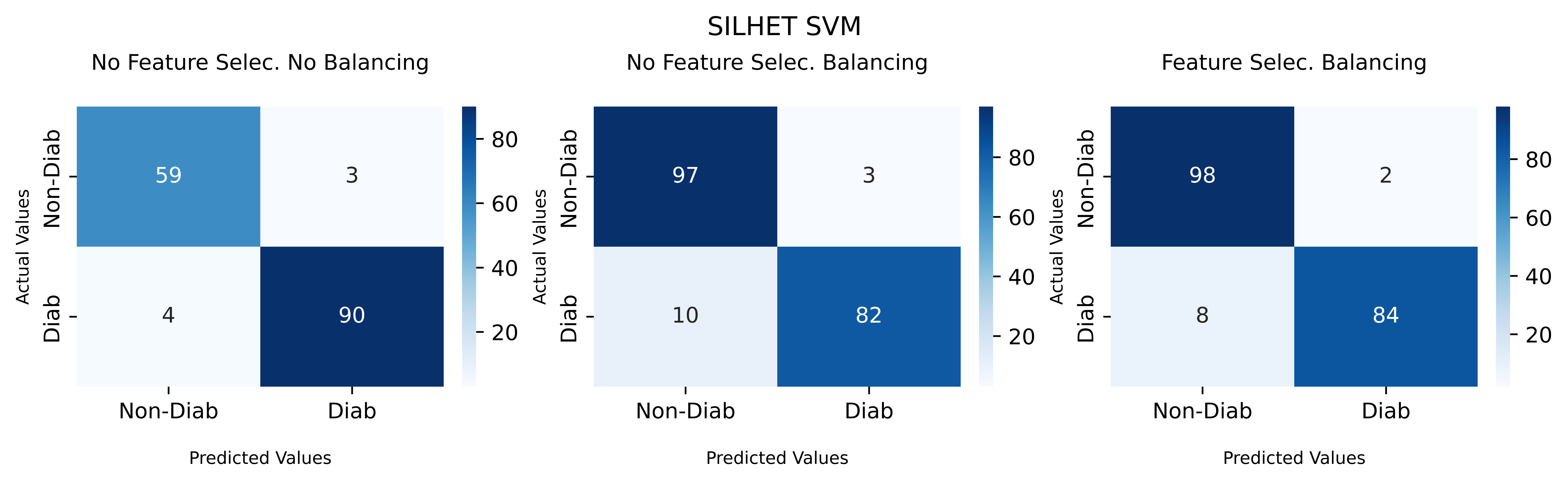}
    \caption{Sylhet: Support Vector Machine}
    \label{fig:confusion matrix SYLHET SVM}
\end{subfigure}
\hfill
\begin{subfigure}{0.9\textwidth}
    \includegraphics[width=\textwidth]{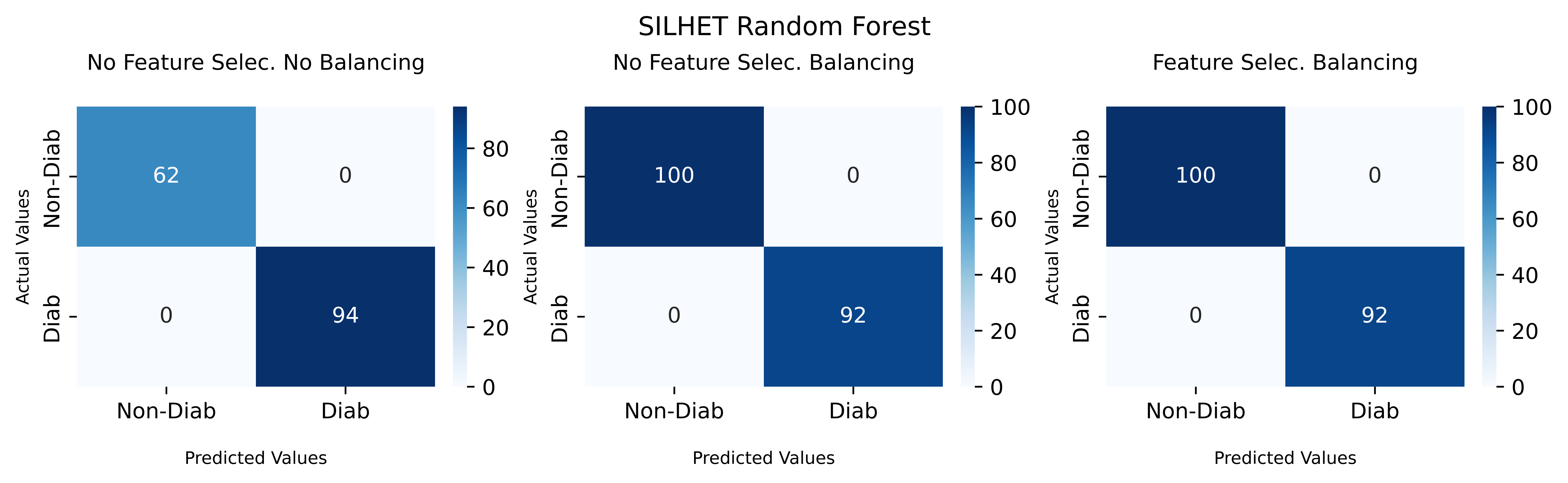}
    \caption{Sylhet: Random Forest}
    \label{fig:confusion matrix SYLHET RF}
\end{subfigure}
 
\caption{Confusion Matrices for Sylhet dataset.}
\label{fig:confusion matrix Sylhet} 
\end{figure}

\begin{figure}
\centering
\begin{subfigure}{0.9\textwidth}
    \includegraphics[width=\textwidth]{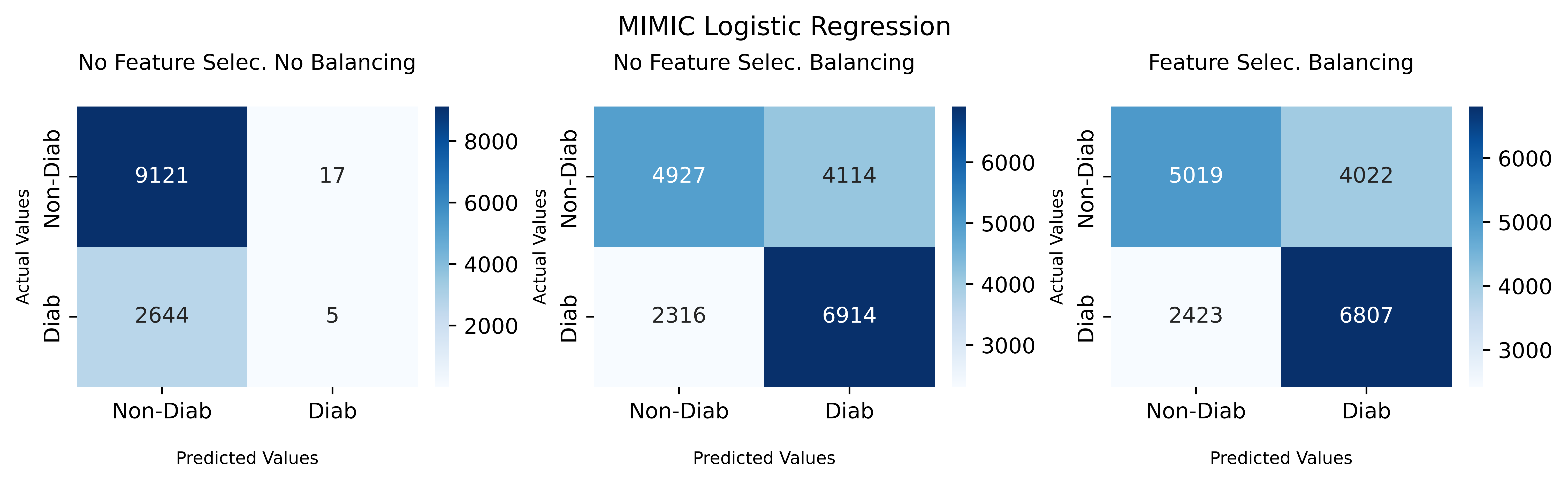}
    \caption{MIMIC III: Logistic Regression}
    \label{fig:confusion matrix MIMIC LR}
\end{subfigure}
\hfill
\begin{subfigure}{0.9\textwidth}
    \includegraphics[width=\textwidth]{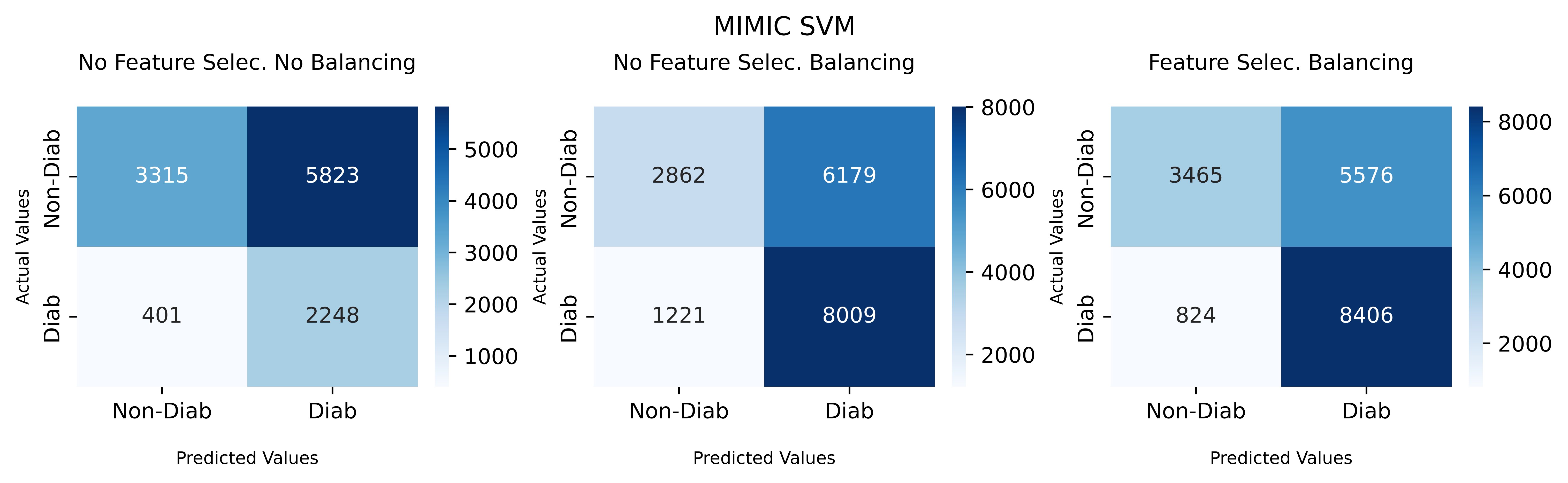}
    \caption{MIMIC III: Support Vector Machine}
    \label{fig:confusion matrix MIMIC SVM}
\end{subfigure}
\hfill
\begin{subfigure}{0.9\textwidth}
    \includegraphics[width=\textwidth]{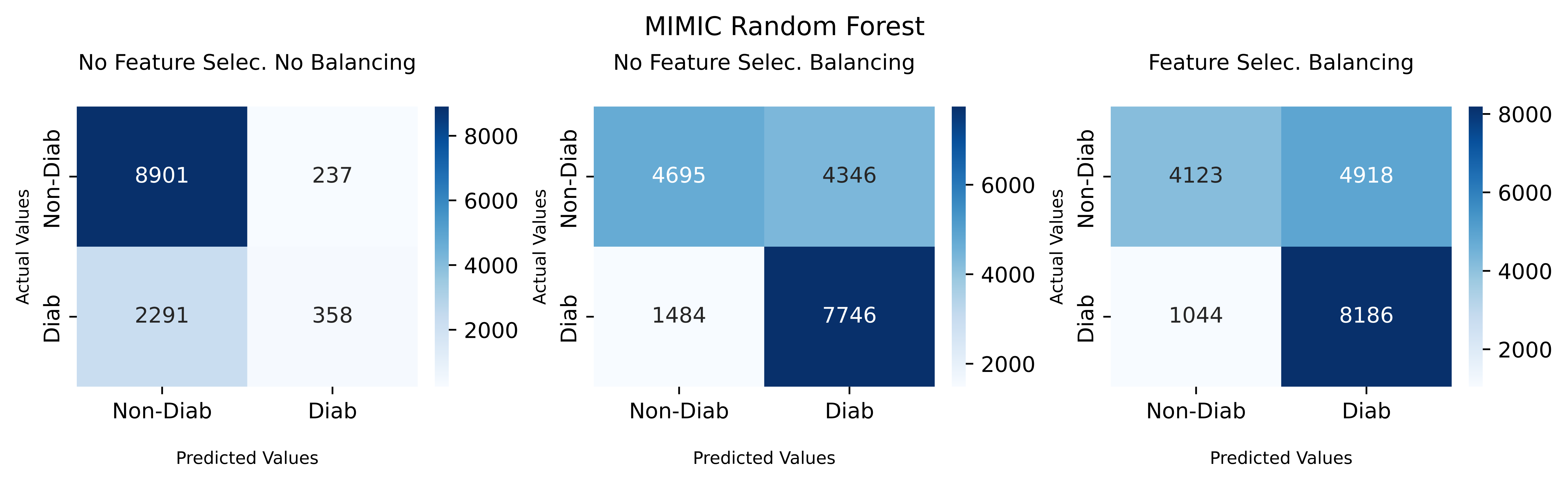}
    \caption{MIMIC III: Random Forest}
    \label{fig:confusion matrix MIMIC RF}
\end{subfigure}
 
\caption{Confusion Matrices for MIMIC III dataset.}
\label{fig:confusion matrix MIMIC}
\end{figure}

In this section, we analyze our experimental results and give insights into the reasons for the obtained performance. To compare the different metrics for the models under study Figures \ref{fig:performance PIMA}, \ref{fig:performance Sylhet}, and \ref{fig:performance MIMIC III} show the Accuracy, AUC, Recall, Precision, and F-measure values of the models for the different datasets, with and without feature selection before and after balancing. From the raw results of accuracy, we can see that feature selection improves or at least does not degrade accuracy. As for balancing, we observe mixed results on accuracy depending on how balanced were the data initially. Accuracy is improved with data balancing in the case of   RF algorithm where we can see an increase of accuracy from 0.77 to 0.81 for PIMA Indian, 0.97 to 0.98 for Sylhet.  For the MIMIC III dataset, accuracy decreases from 0.77 to 0.66 with data balancing, but the F-measure increases from 0.51 to 0.66. This is a general conclusion for all datasets and all algorithms, analysis of confusion matrices can give an insight on this. The confusion matrices (figures \ref{fig:confusion matrix PIMA}, \ref{fig:confusion matrix Sylhet}, and \ref{fig:confusion matrix MIMIC}) before and after feature selection and balancing, we can see that after balancing there is better detection of the minority class. For the RF algorithm, the increase in detection of the minority class is 70\% less false negative for PIMA Indian and 80\% less false negative for the MIMIC III dataset. For the Sylhet dataset, there is no significant improvement because it is already balanced and there was no false negative before balancing.

The execution times of the algorithms for the PIMA India and Sylhet are negligible while for the MIMIC III are the highest. This is because time is a function of the number of features and observations. The datasets we used are not heavily imbalanced, so we can notice that balancing does not show accrued accuracy (except for the Sylhet dataset where there is a slight improvement). This is particularly true for the MIMIC III dataset because it does not present enough risk factor features and it is also the most unbalanced. The balancing algorithm (SMOTE) is inefficient in producing better training in this case, this is because SMOTE oversamples uninformative samples. As a general result, we can observe that the feature selection that we perform in the system does not degrade the accuracy and can reduce the processing time. The Best ML algorithm for the PIMA Indian dataset is Random Forest when using Feature Selection with an accuracy score of 0.7827 The Best ML algorithm for Sylhet is Random Forest with an accuracy score of 0.9723, the feature selection brings a slight decrease in processing time and the accuracy does not suffer. The best ML algorithm for the MIMIC III dataset is Logistic Regression, but Random Forest is very near. The best accuracy for LR is 0.7734 and the best for RF is 0.7703 and difference of only 0.4\%. ROC curve analysis shows that for the Sylhet dataset, the Random Forest classifier is very good. The results for the Sylhet dataset are exceedingly better than those for PIMA Indian and even more than for MIMIC III. The main difference between these datasets is the number of available features that can be seen as risk factors for diabetes. In conclusion, we can say that we should strive to get data with as many risk factors as possible (i.e. Sylhet dataset. The datasets we used for analysis were not diverse enough to assess the need for balancing the data.  The use of Random Forest with feature selection is justified in the system, as it can reduce processing time.

\section{Conclusions}

In this paper, we propose an end-to-end integrated IoT-edge-AI-blockchain monitoring system for diabetes prediction.  In addition, we evaluate machine learning algorithms within the proposed system using three diabetes datasets in a unified setup and compare their performance in terms of accuracy, F-measure, and execution time. Our experimental results show that the RF is the most accurate. Additionally, we classify type 2 diabetes risk factors to analyze the most significant properties for diabetes prediction. When using a classification algorithm for the prediction of type 2 diabetes, the following requirements should be considered.

\begin{enumerate}
\item \textit{Accuracy vs F-measure:} Most of the algorithms give a high classification accuracy. However, evaluating the classifier performance using only the accuracy can be misleading. This is because, in the case of an imbalanced dataset, which is very frequent in the health domain, the algorithm might have high accuracy but will not be able to classify the minority class labels as revealed by the F-measure. In such a situation, the prediction results can lead to a life-threatening situation, as a diabetic patient can be classified as non-diabetic. Consequently, we recommend the data scientist include F-measure as one of the evaluation metrics.

\item \textit{Feature selection:}  Feature selection algorithms should be used on the dataset before training the classification model. This can avoid overfitting and reduces execution time. The experiments we conducted show that feature selection does not incur accuracy degradation.

\item \textit{Significant features:} As a recommendation we can propose to use age, ethnicity, glucose, family history of diabetes, and obesity for the prediction of type 2 diabetes based on our experimental results.  This is in alignment with the Finnish Diabetes Association's type 2 diabetes risk assessment form \cite{eRiskite84:online} and the American Diabetes Association's type 2 diabetes risk test \cite{DIABETES14:online}.
\end{enumerate}

\section*{Acknowledgments}
This research was funded by the National Water and Energy Center of the United Arab Emirates University (Grant 31R215).

\section*{Declarations of Interest}
Declarations of interest: none

\bibliographystyle{elsarticle-num}

\end{document}